\documentclass{article}

\PassOptionsToPackage{numbers,compress}{natbib}
\usepackage[preprint]{neurips_2026}

\usepackage[utf8]{inputenc}
\usepackage[T1]{fontenc}
\usepackage{hyperref}
\usepackage{url}
\usepackage{booktabs}
\usepackage{multirow}
\usepackage{amsfonts}
\usepackage{amsmath}
\usepackage{amssymb}
\usepackage{nicefrac}
\usepackage{microtype}
\usepackage{xcolor}
\usepackage{graphicx}
\usepackage{fancyvrb}
\usepackage{algorithm}
\usepackage[noend]{algpseudocode}

\algrenewcommand\algorithmicrequire{\textbf{Input:}}

\title{ChronoForest: Closed-Loop Multi-Tree Diffusion Planning for Efficient Bridge Search and Route Composition}

\author{%
  Jungmin Seo \\
  Seoul National University \\
  Seoul, South Korea \\
  \texttt{jmseo1204@snu.ac.kr} \\
  \And
  Jaesik Park \\
  Seoul National University \\
  Seoul, South Korea \\
  \texttt{jaesik.park@snu.ac.kr} \\
}

\begin{document}

\maketitle

\begin{abstract}
  How can we plan long-horizon routes that reach designated goals, visit required waypoints, and remain short when only short-horizon offline trajectories are available? This problem matters in offline navigation because collecting sufficiently rich long-horizon data is difficult, yet real agents must still solve long-range tasks with route-level efficiency rather than mere feasibility. The difficulty is twofold: at the microscopic level, composing many short-horizon segments creates a trade-off between search cost and path quality, while at the macroscopic level, waypoint ordering requires comparing pairwise travel costs among start, goal, and waypoint anchors that are unknown before planning and increasingly unreliable when estimated only from long-range temporal distance.

In this paper, we propose ChronoForest, a closed-loop planning system that couples local bridge search and online route re-solving through an anchor-chaining tree diffusion planner and an online multi-tree orchestrator. ChronoForest uses temporal distance for short-range guidance and node evaluation, while using search-time bridge evidence to validate long-range anchor connectivity and repeatedly re-solve the route. On OGBench AntMaze-Stitch, ChronoForest achieves 99.8\%, 99.3\%, and 99.5\% success on the medium, large, and giant splits and improves giant-stitch success by up to 34.5 points over prior reported diffusion-based results. On Hamiltonian route-composition benchmarks, online re-solving corrects poor temporal orderings and improves route quality while remaining substantially cheaper than exhaustive planning.

\end{abstract}

\section{Introduction}
The only available data are offline short-horizon navigation episodes. Each agent is given a start-goal pair, and the shared waypoint set must be covered by the resulting route family. Our objective is to construct a long-horizon route family that covers all waypoints while minimizing total route cost. This setting is difficult because long-horizon trajectory data is scarce, so the planner must extrapolate beyond the training horizon from only short-horizon offline data at inference time. In our problem, route efficiency has two coupled levels: each anchor pair formed by a start, goal, or waypoint should be connected by a short feasible bridge under a limited search budget, and the agents should cover the shared waypoints through a short overall route family. This high-level routing problem is related to the multiple Traveling Salesman Problem (mTSP) with distinct origins and destinations, except that anchor-pair costs are not known before planning and must instead be estimated online.

These two levels are difficult for different reasons and remain only partially addressed when treated in isolation. At the low level, extending a short-horizon diffusion model through composition creates a trade-off between \textbf{search efficiency} and \textbf{path efficiency}: searching more candidates can uncover a shorter bridge but increases inference cost, whereas committing early can return a feasible but unnecessarily long path. Existing diffusion-based and compositional planners mainly optimize plausible or goal-reaching generation rather than active path shortening \citep{janner2022diffuser,zhou2023replan,yoon2025cmctd}. At the high level, waypoint ordering requires comparing many anchor pairs even though the pairwise shortest-path matrix is unavailable a priori and long-range temporal-distance estimates become less reliable as the horizon grows. Classical navigation-coupled routing and mTSP methods therefore do not transfer directly because they assume an explicit map, roadmap, or precomputed pairwise costs.

In this paper, we propose \textbf{ChronoForest}, a closed-loop planning system that couples local bridge search with online route re-solving from short-horizon offline data. Its low-level \textbf{Anchor-chaining tree diffusion planner} grows bidirectional search trees and uses temporal distance to guide short-range bridge construction under a limited inference budget. Its high-level \textbf{Online multi-tree orchestrator} repeatedly re-solves the tentative route from accumulated bridge evidence and reallocates the next expansion budget to the ordered pairs that most affect the current route objective. This coupling is essential: low-level bridge evidence refines anchor-to-anchor cost estimates, and the updated route hypothesis steers subsequent search toward the ordered pairs on which the current tentative route depends. Figure~1 visualizes this closed-loop flow between bridge search and route re-solving.

Our contributions are as follows.

\begin{itemize}
  \item We formalize a navigation-coupled mTSP setting in which long-horizon waypoint planning and routing must be solved jointly from offline short-horizon data without a precomputed pairwise cost matrix.
  \item We propose \textbf{ChronoForest}, a closed-loop planning system that couples the \textbf{Anchor-chaining tree diffusion planner} with the \textbf{Online multi-tree orchestrator}: temporal distance drives local guidance and node evaluation to gather bridge evidence, and the orchestrator re-solves the tentative route from this evidence and reallocates search budget to the ordered pairs whose updated pairwise cost estimates can most directly affect the current route objective.
  \item On OGBench AntMaze-Stitch, ChronoForest reaches 99.8\%, 99.3\%, and 99.5\% success on the medium, large, and giant splits, improves giant-stitch success by up to 34.5 points over prior diffusion-based planners, and shows in Hamiltonian route-composition analyses that online route re-solving corrects poor temporal orderings and improves route quality without paying the cost of exhaustive planning.
\end{itemize}

The rest of the paper introduces the planning setting and local planning interface, presents the ChronoForest method, and evaluates its route quality and search efficiency on offline navigation benchmarks.

\section{Related Work}

\subsection{Temporal distance representations}

Prior work studies the ingredients of our setting from different angles; among them, temporal-distance and quasimetric representations are the closest line of work on reachability-aware state-goal estimation. Temporal-distance and quasimetric representations learn reachability-aware state-goal distance beyond Euclidean proximity, either through architecture-level structure or through objective and update design \citep{mrn2023,iqe2023,qrl2023,cmd2024,tmd2025,mqe2025,trl2025}. Application-oriented methods such as GAS and TLDR use these estimates for graph abstraction, subgoal search, or far-goal exploration, whereas ChronoForest reuses temporal-distance estimates inside diffusion guidance and node evaluation for bridge search \citep{gas2025,tldr2024}. Because long-range temporal-distance estimates degrade as horizon grows, ChronoForest does not treat them as final route costs; it uses them only for local progress estimation while distant anchor connectivity is validated by bidirectional multi-tree search.

\subsection{Diffusion planners for long-horizon composition}

Diffusion planners such as Diffuser, Replan, and milestone-based variants improve long-horizon control by generating or refining goal-conditioned trajectories, but they work best when training data already covers sufficiently long behavior horizons \citep{janner2022diffuser,zhou2023replan,milestone2023,hierdiffusion2024}. Related sequence-generation training paradigms such as Diffusion Forcing improve long-horizon sequence modeling, but their core contribution is a training paradigm rather than an explicit route-cost-aware planner \citep{chen2024diffusionforcing}. Compositional variants such as CompDiffuser, CDGS, and SCoTS reduce the dependence on training data that already covers sufficiently long behavior horizons by recombining short fragments or enlarging coverage of long-horizon behaviors, yet their main objective remains globally plausible long-horizon generation rather than explicit route-cost-aware ordering \citep{compdiffuser2025,cdgs2026,scots2025}. Tree-augmented planners such as MCTD, TDP, and C-MCTD further allocate inference-time search budget across candidate denoising branches, but they still primarily optimize sampling-time exploration rather than route-level re-solving from accumulated bridge evidence \citep{mctd2025,tdp2025,yoon2025cmctd}. ChronoForest combines these strands by using short-fragment diffusion inside a tree-based planning loop that allocates limited search budget to collecting bridge evidence and repeatedly re-solves the route from that evidence. This shifts the comparison focus from plausibility alone to path efficiency and to the effect of new bridge evidence on the current route objective.

\subsection{Navigation-coupled routing and mTSP}

Navigation-coupled routing and mTSP methods typically assume an explicit map, roadmap, or precomputed pairwise cost matrix before visit ordering is solved \citep{belieftsp2019,boundarynode2021,smug2023}. Expensive-edge-evaluation work such as LazySP studies how edge-evaluation effort should be allocated across unresolved edges when edge costs are not yet fully resolved \citep{lazysp2016}. Related adaptive routing work instead studies broader stochastic routing objectives rather than LazySP-style edge-evaluation allocation \citep{adaptiverouting2016}. ChronoForest differs because anchor-pair costs are not available a priori and must be estimated online from short-horizon offline bridge evidence, so routing is repeatedly re-solved as multi-tree search discovers new feasible connections.

\section{Preliminaries}
\subsection{Multiple Traveling Salesman Problem (mTSP) with Distinct Origins and Destinations}

Our high-level routing problem can be viewed as a \texttt{multiple Traveling Salesman Problem (mTSP)} with distinct origins and destinations defined over an anchor set. Let the agent index set be $\mathcal{M}:=\{1,\dots,M\}$, the shared waypoint set be $\mathcal{W}:=\{W_1,\dots,W_K\}$, and, for each agent $m\in\mathcal{M}$, let $S_m$ and $G_m$ denote its start and goal anchors. We write
\[
\mathcal{A}:=\mathcal{W}\cup\{S_m\}_{m=1}^M\cup\{G_m\}_{m=1}^M,
\qquad
\pi_m:=(a^{(m)}_0=S_m,\dots,a^{(m)}_{|\pi_m|}=G_m),
\quad a^{(m)}_r \in \mathcal{A}.
\]

A joint route is written as $\boldsymbol{\pi}:=(\pi_1,\dots,\pi_M)\in\Pi_M(\mathcal{A};\{S_m\},\{G_m\})$, where each route starts at $S_m$ and ends at $G_m$, each waypoint $W_k\in\mathcal{W}$ is assigned to exactly one route, and routes are otherwise disjoint across agents. Let $\mathrm{Adj}(\boldsymbol{\pi})=\bigcup_{m=1}^M \mathrm{Adj}(\pi_m)$ denote the set of ordered adjacent anchor pairs on the joint route.

If the cost of moving from anchor $i$ to $j$ is $c_{ij}$, the route objective minimizes total adjacent-pair cost under these pairwise costs. In our setting, however, the pairwise route costs are not fully given in advance and are instead refined online from low-level bridge evidence as planning proceeds.

\subsection{Two-Stage Planning}

\subsubsection{Diffusion Planner}
\label{sec:diffusion-planner}

The diffusion planner takes a boundary state $s_{\mathrm{in}}$ and a target state $g$ as input and predicts an $L$-step lookahead subplan $X_0=(x^{(1)},x^{(2)},\dots,x^{(L)})$. At diffusion step $t$, we denote the noisy sequence by $X_t$ and the denoised estimate by $\hat X_0=f_\theta(X_t,t)$. For later tree-state notation, we write a local trajectory segment as $\xi^{(s)}=(x^{(s,0)},x^{(s,1)},\dots,x^{(s,L)})\in\mathcal{S}^{L+1}$ with boundary state $x^{(s,0)}$ and generated suffix $X^{(s)}=(x^{(s,1)},\dots,x^{(s,L)})$. During planning, target-conditioned guidance steers denoising toward the assigned target using the temporal-distance score defined in Section~\ref{sec:temporal-distance}; Section~\ref{sec:junction-detector} gives the concrete guidance form used by ChronoForest. The resulting state sequence is executed by a separate tracking policy rather than by inverse dynamics.

\subsection{Temporal Distance}
\label{sec:temporal-distance}

Here, temporal distance denotes goal-reaching difficulty measured in expected step count. Let $d^*(s,g)$ denote the minimum expected number of steps required to reach goal state $g$ from state $s$. Following prior temporal-distance representation learning \citep{park2024hilp,park2023hiql}, we use a goal-conditioned value function $V_{\mathrm{TD}}(s,g)$, whose exact stitch and teleport value parameterizations are defined in Appendix~\ref{app:value-representations}. With discount factor $\gamma\in(0,1)$, the planner converts this value into the planning-time quantity
\[
d_{\mathrm{TD}}(s,g):=\frac{\log\!\left(1+(1-\gamma)V_{\mathrm{TD}}(s,g)\right)}{\log \gamma}.
\]
The planner uses $d_{\mathrm{TD}}(s,g)$ rather than the raw value because local planning requires a monotone estimate of remaining steps. Throughout the paper, $d_{\mathrm{TD}}$ is used for local planning operations introduced in Section~\ref{sec:method-overview} and the later method subsections rather than as an exact long-range route-cost oracle.

\section{Method}
\subsection{Overview}

\label{sec:method-overview}

ChronoForest solves two coupled problems: efficient low-level bridge construction from short-horizon offline data and high-level route refinement when anchor-pair costs are revealed only progressively during planning.

At the low level, the planner must find short feasible bridges within a limited inference budget, which creates a trade-off between search efficiency and path efficiency as the stitching horizon grows. At the high level, the route optimizer must compare many candidate anchor adjacencies even though the pairwise cost matrix is unavailable before planning and long-range temporal-distance estimates become less reliable.

ChronoForest addresses these challenges with two interacting modules. The Anchor-chaining tree diffusion planner expands bidirectional multi-tree frontiers, uses temporal distance for target-conditioned guidance and node evaluation, and accumulates candidate bridge evidence near each anchor pair. The Online multi-tree orchestrator repeatedly re-solves the tentative route from this evidence and allocates the next round of expansion budget to the ordered pairs whose updated pairwise cost estimates can most directly affect the current route objective.

These modules must be coupled rather than optimized in isolation. Low-level bridge construction reduces uncertainty in anchor-to-anchor cost estimates and therefore improves route recovery. Conversely, the current tentative route determines which ordered pairs should be expanded next so that planning budget is concentrated on the adjacencies whose evidence can most directly affect the route objective.

\begin{figure*}[t]
  \centering
  \includegraphics[width=\textwidth]{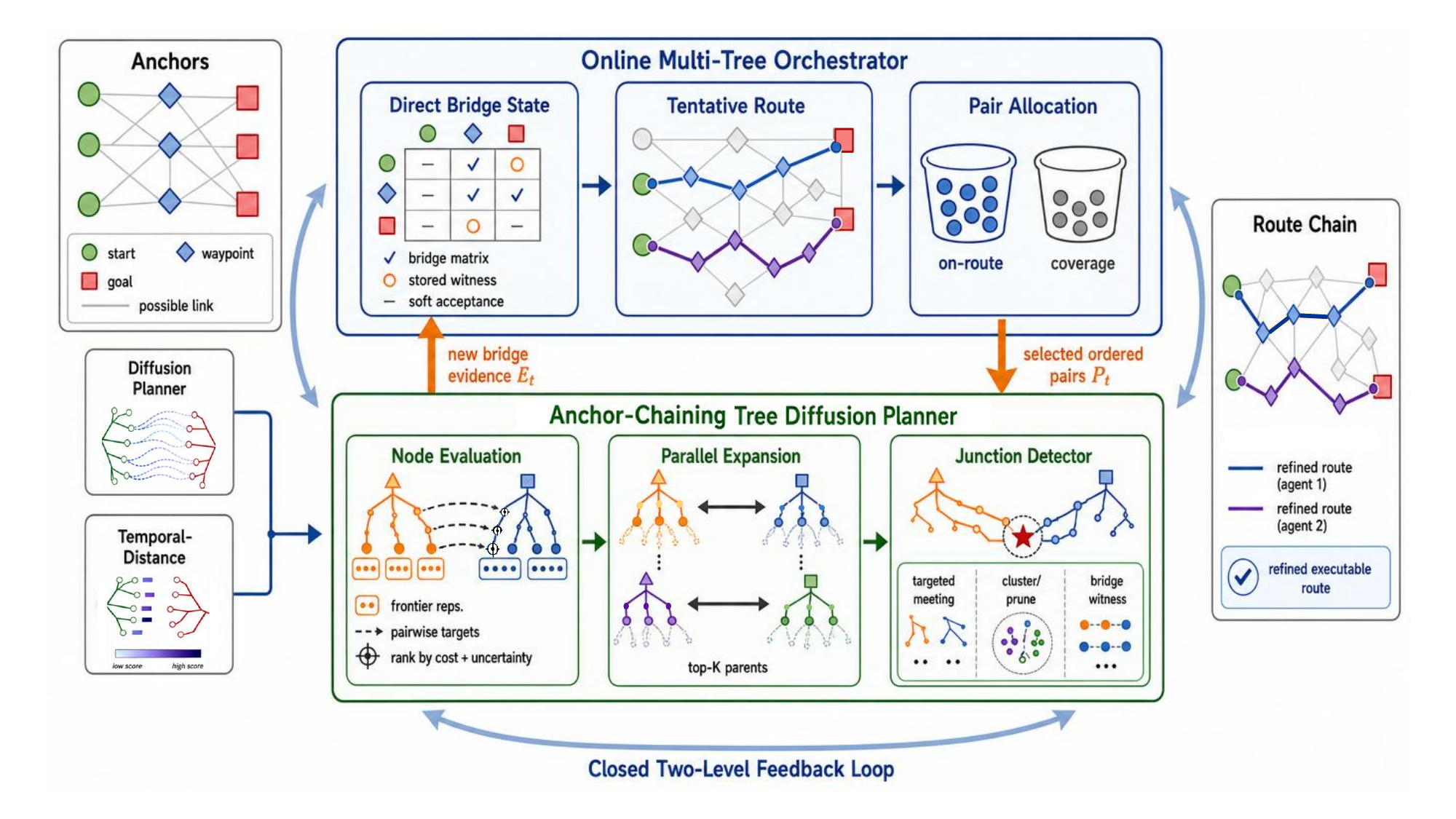}
  \caption{
    Closed-loop interaction between the Anchor-chaining tree diffusion planner and the Online multi-tree orchestrator. The planner uses temporal distance to gather bridge evidence, and the orchestrator re-solves the tentative route and reallocates expansion to the ordered pairs that most affect the current route objective.
  }
  \label{fig:chronoforest-main-overview}
\end{figure*}

Figure~\ref{fig:chronoforest-main-overview} summarizes this closed-loop interaction, and Appendix~\ref{app:full-algorithm} provides the full ChronoForest pseudocode before the later subsections specify each module in detail.

% Main overview figure for the method section.

\subsection{Anchor-chaining Tree Diffusion Planner}

\label{sec:anchor-chaining-tree-diffusion-planner}

At the low level, ChronoForest alternates among three operations. Node evaluation ranks expandable leaves using pair-conditioned route efficiency together with an uncertainty penalty; selective and parallel node expansion materializes the most promising children for the currently selected ordered pairs; and the junction detector generates, prunes, and clusters target-conditioned lookahead subplans so that the next round reuses a compact representative set. Together these steps search for short bridge evidence while controlling the branching factor of the tree.

\subsubsection{Architecture-Level Design}

\label{sec:architecture-level-design}

We use a tree-structured planner because multiple candidate futures often share long prefixes. For each anchor $a_i\in\mathcal{A}$, we therefore maintain a rooted tree $\mathcal{T}_i=(\mathcal{V}_i,\mathcal{E}_i,n_i^{\mathrm{root}})$ whose root starts at $a_i$; each node $n\in\mathcal{V}_i$ stores a materialized prefix $\mathrm{pref}(n)$, current node state $x(n)$, cost-to-come summary $g(n)$, target pointer $\mathrm{tgt}(n)$, and representative record $\mathcal{R}(n)$. Each representative caches a local segment together with its lookahead suffix, so common prefixes are materialized once while later node evaluation, meeting detection, and route assembly can reuse the same prefix state; Appendix~\ref{app:planner-state-details} gives the exact tuple, the recurrence for $g(n)$, and the root-target construction, while Section~\ref{sec:node-evaluation} later refreshes target pointers by the pair-conditioned rule.

Using multiple trees is beneficial because it broadens spatial coverage, enables pair-specific target assignment for heterogeneous bridge hypotheses, and reduces redundant denoising by reusing shared prefixes across related descendants.

\subsubsection{Node Evaluation}

\label{sec:node-evaluation}

\textbf{Pair-Conditioned Target Selection}

Because different ordered tree pairs can require different bridge hypotheses, target selection must be pair-conditioned rather than globally shared. Once the route-level orchestrator chooses an ordered source/counterpart pair $(i,j)$ to expand, each source-side representative $\rho$ in $\mathcal{T}_i$ selects a target node $m\in\mathcal{V}_j$ from the counterpart tree $\mathcal{T}_j$ by minimizing the pair-conditioned estimated total cost. Here $x(\rho)$ denotes the representative endpoint and $g(\rho)$ the cumulative cost of the latent child prefix obtained by appending $\rho$'s cached lookahead subplan to its parent prefix, so the source-side cost-to-come and target-side materialized prefix cost are $D_i^{\mathrm{src}}(\rho):=g(\rho)$ and $D_j^{\mathrm{tgt}}(m):=g(m)$,
\[
\mathcal{J}_{i\to j}(\rho,m)
:=
D_i^{\mathrm{src}}(\rho)
+ h\!\left(
 d_{\mathrm{TD}}\!\bigl(x(\rho),x(m)\bigr)
\right)
+ D_j^{\mathrm{tgt}}(m),
\qquad
h(x)=x+\alpha x^2
\]
\[
m_\rho^\star
\in
\arg\min_{m\in\mathcal{V}_j}
\mathcal{J}_{i\to j}(\rho,m).
\]
This pair-specific target is then used for same-round lookahead generation, and the same $\mathcal{J}_{i\to j}$ is reused by the route-level direct-bridge term in Section~\ref{sec:online-multi-tree-orchestrator}; Appendix~\ref{app:pair-conditioned-ranking-details} gives the exact candidate pool, latent-child construction, and remaining helper definitions.

\textbf{Source Node Evaluation}

Within a source tree, we rank candidate source nodes in the current pair-conditioned candidate set $\mathcal{S}_t^{(i,j)}\subseteq\mathcal{V}_i$, where $\mathcal{S}_t^{(i,j)}$ contains the source leaves with nonempty representative sets that are currently eligible for ordered pair $(i,j)$. The ranking jointly considers two factors: how close the induced path is to the shortest path and how uncertain the generated lookahead subplan is. The former encourages shortest-path formation, whereas the latter penalizes high endpoint dispersion in the stochastic setting. Accordingly, the source-node score is the ordinal combination
\[
\mathrm{score}(n)
=
-\Bigl(
\operatorname{rank}\bigl(\mathrm{cost}(n)\bigr)
\;+\;
\lambda_{\mathrm{unc}}\operatorname{rank}\bigl(U(n)\bigr)
\Bigr).
\]
Here, $\operatorname{rank}(\cdot)$ is the ascending rank computed over $\mathcal{S}_t^{(i,j)}$, $\mathrm{cost}(n)$ is the minimum pair-conditioned representative cost over $\mathcal{R}(n)$, and $U(n)$ is the cluster-mass-weighted endpoint-dispersion penalty over the same representatives; Appendix~\ref{app:pair-conditioned-ranking-details} gives the exact definitions of $\mathrm{Cost}(\rho)$, $p_\rho$, and the uncertainty functional.

\subsubsection{Selective and Parallel Node Expansion}

\label{sec:selective-parallel-node-expansion}

At each round, for each active ordered pair $(i,j)$, we keep the top-$K_{\mathrm{exp}}$ source nodes under the ranking above, where $K_{\mathrm{exp}}$ is the per-pair expansion quota,
\[
\mathcal{N}_t^{(i,j)}
=
\operatorname{TopK}\!\left(
\mathcal{S}_t^{(i,j)},
\mathrm{score}(\cdot),
K_{\mathrm{exp}}
\right),
\]
expand them by promoting promising lookahead descendants into the active frontier, and then test whether any newly exposed prefix can meet a target-conditioned counterpart from another tree. This merged step keeps only a small frontier in the main loop while still surfacing candidate junctions early enough for the route-level orchestrator. Appendix~\ref{app:expansion-meeting-details} records the exact top-$K_{\mathrm{exp}}$ selection rule, child-promotion formula, and meeting criterion used to instantiate this step.

\subsubsection{Junction Detector}

\label{sec:junction-detector}

\textbf{Target-Conditioned Diffusion Guidance} In the current round, guided denoising is executed over the mixed batch $\mathcal{B}_t$, where $\mathcal{P}_t$ is the active ordered-pair set, $\mathcal{N}_t^{(i,j)}$ is the selected source-node set for pair $(i,j)$, and $N_{\mathrm{samp}}$ is the number of guided lookahead samples per selected source node:
\[
\mathcal{B}_t
:=
\left\{
(i,j,n,q)
\;\middle|\;
(i,j)\in\mathcal{P}_t,\;
n\in\mathcal{N}_t^{(i,j)},\;
q\in\{1,\dots,N_{\mathrm{samp}}\}
\right\}.
\]
Thus a single denoising call generates target-conditioned future subplans from $x(n)$ toward $x(\mathrm{tgt}(n))$ in parallel across pairs, nodes, and samples. These lookaheads provide future-segment evidence for node evaluation, expose local multi-modality for the uncertainty term, and only those that survive pruning and clustering are retained as representatives for the next round. In this step, target-conditioned diffusion guidance injects the temporal-distance score into denoising so that each lookahead subplan is steered toward its assigned target node,
\[
g_{\mathrm{tgt}}:=x(\mathrm{tgt}(n)),
\qquad
\mathcal{U}(\hat X_0; g_{\mathrm{tgt}})
:=
-d_{\mathrm{TD}}\!\bigl(\hat x^{(L)},g_{\mathrm{tgt}}\bigr)
\qquad
\tilde X_0
=
\hat X_0 + \eta_t \nabla_{\hat X_0}\mathcal{U}(\hat X_0; g_{\mathrm{tgt}})
\]
The same mixed batch therefore preserves pair-, node-, and sample-level parallelism throughout guided denoising. Appendix~\ref{app:guidance-objective} defines the general sparse guidance score, and Appendix~\ref{app:guidance-denoised-update} gives the denoised-space injection rule instantiated here.

\textbf{Subplan Pruning and Representative Reuse} After guided lookahead evaluation, we discard infeasible or dominated candidates, cluster the surviving subplans, and keep one representative per retained local mode for reuse in later rounds. This preserves one reusable representative per surviving cluster while capping the branching factor of the online planner; the exact pruning tests, clustering threshold, and representative-selection rule are deferred to Appendix~\ref{app:pruning-representatives}.

\subsection{Online Multi-Tree Orchestrator}

\label{sec:online-multi-tree-orchestrator}

The low-level planner in Section~\ref{sec:node-evaluation} and Section~\ref{sec:junction-detector} produces pairwise bridge evidence in the form of pair-conditioned scores $\mathcal{J}_{i\to j}(\rho,m)$, but a separate route-level controller is needed to decide which ordered pairs should be expanded next. The online multi-tree orchestrator converts that evidence into route-level decisions: it updates the direct bridge state $C_t$ and its closure $\widetilde{C}_t$, re-solves the tentative route $\hat{\Pi}_t$, allocates the next pair budget, and refreshes a revocable accepted adjacency set $E_{\mathrm{soft}}^t$. Appendix~\ref{app:mtsp-route-comparison} analyzes the relaxed closure-based route comparison underlying this allocation rule, Appendix~\ref{app:mtsp-margin-threshold} interprets the route-certification role of $\tau$, and Appendix~\ref{app:mtsp-acceptance-coverage} discusses the accepted-edge and coverage side conditions.

\subsubsection{Direct Bridge Matrix and All-Hop Closure State}

We begin by defining the route-level state that aggregates the pair-conditioned bridge evidence returned by the low-level planner. Let $\widehat{c}_t^{\,ij}(\rho,m):=\mathcal{J}_{i\to j}(\rho,m)$ for a source-side representative or witness $\rho$ in tree $\mathcal{T}_i$ and a counterpart target node $m$ in tree $\mathcal{T}_j$, and let $\mathcal{N}_{ij}^{\mathrm{new}}(t)$ denote the set of newly evaluated representative-target witness pairs for ordered pair $(i,j)$ at round $t$.

The direct-bridge matrix stores the best current direct entry for each ordered anchor pair, with $\min_{\varnothing}:=+\infty$, while the route solver operates on the Floyd-Warshall all-hop closure:
\[
\begin{aligned}
C_t(i,j)
&=
\min\!\left(
C_{t-1}(i,j),
\min_{(\rho,m)\in\mathcal{N}^{\mathrm{new}}_{ij}(t)}
\widehat{c}_t^{\,ij}(\rho,m)
\right),
\;
\widetilde{C}_t(i,j)
\!&=
\min_{L\ge 1}
\min_{\substack{\zeta_0=i,\;\zeta_L=j,\\ \zeta_1,\dots,\zeta_{L-1}\in\mathcal{A}}}
\sum_{\ell=0}^{L-1} C_t(\zeta_\ell,\zeta_{\ell+1}).
\end{aligned}
\]

For each ordered pair, $B_t(i,j)$ stores the witness attaining the current direct entry, and $\gamma_t(i,j)$ denotes that witness's predicted gap; if no witness has been found, we set $\gamma_t(i,j)=+\infty$. Thus $C_t$ and $B_t(i,j)$ retain direct bridge evidence at the ordered-pair level, whereas $\widetilde{C}_t(i,j)$ is used only for route comparison and route re-solving.

\subsubsection{Tentative Route Re-Solving on the Closure}

The closure $\widetilde{C}_t$ supports route comparison, while actual bridge realization is handled later through meeting acceptance. Accordingly, each round re-solves a tentative joint route and its induced adjacency set

\[
\begin{aligned}
\hat{\Pi}_t
&\in
\arg\min_{\boldsymbol{\pi} \in \Pi_M\!\left(\mathcal{A};\{S_m\},\{G_m\}\right)}
\sum_{(i,j)\in \mathrm{Adj}(\boldsymbol{\pi})}
\widetilde{C}_t(i,j),
\quad
\mathrm{Adj}(\hat{\Pi}_t)
&:=
\bigcup_{m=1}^M \mathrm{Adj}(\hat{\pi}_{t,m}).
\end{aligned}
\]

This is a route hypothesis rather than a final executable trajectory. The route-level comparison object here is the same relaxed closure route defined in Appendix~\ref{app:mtsp-problem-setting} and analyzed in Appendix~\ref{app:mtsp-route-comparison}. The next subsection turns this tentative adjacency set into the two-budget decision about which route-adjacent and off-route anchor pairs should be executed in the current decision round.

\subsubsection{Route-Prioritized Budget Allocation}

Given $\hat{\Pi}_t$, the next question is which anchor-pair adjacencies should receive the limited round budget. ChronoForest splits the pair budget into current-route exploitation and off-route coverage: the current-route quota prioritizes unresolved adjacencies on the tentative route, whereas the coverage quota retains a small off-route floor for pairs whose bridge evidence is still uncertified or missing. Appendix~\ref{app:mtsp-margin-threshold} explains the route-certification role of $\tau$, and Appendix~\ref{app:mtsp-acceptance-coverage} records the exact set definitions and scheduling rules. Each selected anchor-pair candidate is then executed in both directions, and Section~\ref{sec:anchor-chaining-tree-diffusion-planner} applies target selection, node evaluation, child expansion, and meeting detection to every ordered pair in that batch.

\subsubsection{Route-Consistent Soft Acceptance}
\label{sec:route-consistent-soft-acceptance}

Budget allocation determines which ordered pairs are expanded next, whereas executability is tracked through a separate revocable accepted adjacency set $E_{\mathrm{soft}}^t$. A route-adjacent meeting is soft-accepted only when the current round produces a meeting witness and adding that adjacency preserves the accepted-edge compatibility screen. Once an adjacency is supported by $E_{\mathrm{soft}}^t$, it leaves the active current-route pool and is reactivated automatically if later route re-solving removes it from the tentative route. Appendix~\ref{app:mtsp-degree-constraints} gives the exact accepted-edge formulas, Appendix~\ref{app:mtsp-margin-threshold} explains the route-certification role of $\tau$, and Appendix~\ref{app:mtsp-scope} discusses the separation between route-level certification and the junction-detector meeting threshold $\delta_{\mathrm{meet}}$.

\section{Experiments}
We focus the main body on three questions. First, we ask whether ChronoForest solves long-horizon goal-reaching on the official OGBench AntMaze-Stitch benchmark. Second, we ask whether those success gains come from route-efficient planning rather than from detours hidden by a binary success metric. Third, we ask whether ChronoForest obtains such route quality without paying the cost of exhaustive planning. Appendix Table~\ref{tab:antmaze_teleport_benchmark} then tests the same design in a broader stochastic and quasimetric teleport regime.

Unless otherwise noted, all stitch and teleport tasks come from OGBench \citep{park2024ogbench}. For the main stitch comparison, we evaluate \texttt{antmaze-\{medium,large,giant\}-stitch-v0} on five held-out tasks with 20 evaluation episodes per task and four repeated benchmark runs. In Table~\ref{tab:antmaze_stitch_benchmark}, our rows report mean $\pm$ standard deviation over those repeated evaluations of a fixed trained model, whereas prior baselines are copied from the original papers, so we use the table as a within-method evaluation plus a cross-paper trend comparison rather than a unified leaderboard.

\subsection{Question 1: Goal-Reaching Performance on AntMaze-Stitch}
\label{subsec:q1_stitch_success}
Question~1 tests whether ChronoForest is already a strong long-horizon planner on the AntMaze-Stitch benchmark.

Table~\ref{tab:antmaze_stitch_benchmark} answers this question on AntMaze-Stitch. First, ChronoForest is best on all three stitch splits, reaching $99.8\pm0.4$, $99.3\pm1.3$, and $99.5\pm0.5$ on medium, large, and giant. Second, the giant split is the decisive benchmark because it is farthest from ceiling: ChronoForest exceeds CompDiffuser by 34.5 points, CDGS by 15.5 points, and C-MCTD-Preplan by 24.5 points. Third, \emph{ChronoForest-single} drops to $94\pm1$, $72\pm4$, and $18\pm3$, showing that sampling multiple lookahead subplans for node expansion matters most exactly when route composition becomes hardest.

\begin{table*}[t]
  \centering
  \scriptsize
  \setlength{\tabcolsep}{4pt}
  \resizebox{\textwidth}{!}{%
  \begin{tabular}{lcccccccc}
    \toprule
    Benchmark & Diffuser-Replan & Diffusion Forcing & GSC & CompDiffuser & C-MCTD-Preplan & CDGS\textsuperscript{\S} & ChronoForest-single & ChronoForest \\
    \midrule
    Medium & 46$\pm$18 & 60$\pm$13 & 97$\pm$2 & 96$\pm$2 & 94$\pm$9 & -- & 94$\pm$1 & \textbf{99.8$\pm$0.4} \\
    Large & 12$\pm$10 & 26$\pm$9 & 66$\pm$2 & 86$\pm$2 & 94$\pm$9 & -- & 72$\pm$4 & \textbf{99.3$\pm$1.3} \\
    Giant & 0$\pm$0 & 0$\pm$0 & 20$\pm$1 & 65$\pm$3 & 75$\pm$18 & 84$\pm$3 & 18$\pm$3 & \textbf{99.5$\pm$0.5} \\
    \bottomrule
  \end{tabular}%
  }
  \caption{Success rate (\%) on OGBench AntMaze-Stitch for diffusion-based planners. Methods are ordered by ascending giant-stitch success. ChronoForest-single disables sampling multiple lookahead subplans for node expansion, and CDGS\textsuperscript{\S} reports only giant-stitch results with a smaller training set.}
  \label{tab:antmaze_stitch_benchmark}
\end{table*}

\subsection{Question 2: Are the successful plans route-efficient?}
\label{subsec:q2_route_efficiency}
Question~2 asks whether the stitch-success gains come from genuinely short routes.

Table~\ref{tab:hamiltonian_online_fixed} provides the main route-quality evidence. It isolates mismatch waypoint groups on \texttt{antmaze-giant-stitch-v0} where the temporal-distance order disagrees with the graph-optimal order under the OGBench maze-grid BFS backend, keeps the official OGBench start-goal constraints for both $M=1$ and $M=2$, and retains only the upper three quartiles of the graph-only margin $\mathrm{GapRel}$ while excluding the near-tie quartile. Under this controlled setup, the online planner reduces total route length from 181.8 to 172.2 for $M=1$ and from 229.3 to 216.3 for $M=2$, while approaching the privileged graph-fixed reference at 170.2 and 210.1. These gains show that ChronoForest starts from an imperfect temporal-distance prior and corrects poor route orderings, yielding routes closer to the graph-fixed reference.

Appendix Tables~\ref{tab:plan_length_bfs},~\ref{tab:hamiltonian_online_fixed_pairs}, and Appendix Figure~\ref{fig:hamiltonian_online_fixed_pair_examples} show the same route-efficiency trend on official stitch rollouts, across all ten two-agent task pairs, and in representative pair examples.

\begin{table*}[t]
  \centering
  \scriptsize
  \setlength{\tabcolsep}{3pt}
  \resizebox{\textwidth}{!}{%
  \begin{tabular}{l ccc ccc}
    \toprule
    & \multicolumn{3}{c}{$M = 1$ (single agent)} & \multicolumn{3}{c}{$M = 2$ (double agents)} \\
    \cmidrule(lr){2-4}\cmidrule(lr){5-7}
    Margin Bin & Online $\downarrow$ & Temporal Fixed $\downarrow$ & Graph Fixed $\downarrow$ & Online $\downarrow$ & Temporal Fixed $\downarrow$ & Graph Fixed $\downarrow$ \\
    \midrule
    High   & 174.0 & 193.7 & 172.8 & 216.0 & 236.6 & 212.5 \\
    Medium & 167.6 & 175.6 & 164.9 & 213.3 & 226.0 & 206.4 \\
    Low    & 175.0 & 176.2 & 172.9 & 219.6 & 225.4 & 211.5 \\
    Total  & 172.2 & 181.8 & 170.2 & 216.3 & 229.3 & 210.1 \\
    \bottomrule
  \end{tabular}%
  }
  \caption{Route-length comparison for online, temporal-fixed, and graph-fixed Hamiltonian planning on \texttt{antmaze-giant-stitch-v0}. The two column blocks correspond to $M\in\{1,2\}$ active start-goal sets with $T=5$ tasks and $N=3$ waypoint anchors, and entries report mean total postprocessed route length $\sum_{a=1}^{M} L_a$ over retained mismatch waypoint groups stratified by $\mathrm{GapRel}$. Graph-fixed uses the benchmark graph shortest-path order as a privileged reference.}
  \label{tab:hamiltonian_online_fixed}
\end{table*}

\subsection{Question 3: Does ChronoForest obtain such route quality efficiently?}
\label{subsec:q3_search_efficiency}
Question~3 asks whether these route-quality gains are achieved without paying the cost of exhaustive planning.

Table~\ref{tab:planning_latency} gives the cleanest matched evidence for planning efficiency. The online planner stays close to the privileged graph-fixed reference in overall cost, with 91.9 seconds and 29.0 expanded nodes versus 90.1 seconds and 26.5 nodes for graph-fixed, while remaining far cheaper than exhaustive search at 126.7 seconds and 44.6 nodes. This shows that the route-quality gain from online re-solving is not a brute-force effect of maintaining a much larger candidate set, while the slight low-margin edge of online over graph-fixed is only suggestive evidence that flexible re-solving can sometimes settle quickly among several near-optimal routes.

\begin{table*}[t]
  \centering
  \footnotesize
  \setlength{\tabcolsep}{5pt}
  \caption{Planning cost of temporal-fixed, graph-fixed, online, and exhaustive Hamiltonian planners on AntMaze Giant with $M=1$ and $N=3$. Entries report mean $\pm$ std of planner-side latency and expanded nodes over the same stratified $\mathrm{GapRel}$ bins used in Table~\ref{tab:hamiltonian_online_fixed}.}
  \label{tab:planning_latency}
  \resizebox{\textwidth}{!}{%
  \begin{tabular}{l cc cc cc cc}
    \toprule
    & \multicolumn{2}{c}{Temporal Fixed} & \multicolumn{2}{c}{Graph Fixed} & \multicolumn{2}{c}{Online} & \multicolumn{2}{c}{Exhaustive} \\
    \cmidrule(lr){2-3}\cmidrule(lr){4-5}\cmidrule(lr){6-7}\cmidrule(lr){8-9}
    Margin Bin & Planning Time (s) $\downarrow$ & Nodes $\downarrow$ & Planning Time (s) $\downarrow$ & Nodes $\downarrow$ & Planning Time (s) $\downarrow$ & Nodes $\downarrow$ & Planning Time (s) $\downarrow$ & Nodes $\downarrow$ \\
    \midrule
    High   & $84.4 \pm 19.1$ & $28.7 \pm 8.4$ & $\mathbf{76.3 \pm 14.8}$ & $\mathbf{23.2 \pm 6.9}$ & $82.5 \pm 32.7$ & $26.1 \pm 11.7$ & $119.1 \pm 32.3$ & $44.5 \pm 11.2$ \\
    Medium & $108.7 \pm 46.3$ & $37.1 \pm 16.7$ & $\mathbf{99.0 \pm 45.8}$ & $\mathbf{29.8 \pm 14.7}$ & $109.8 \pm 59.0$ & $34.4 \pm 15.3$ & $132.1 \pm 42.6$ & $45.4 \pm 13.9$ \\
    Low    & $89.4 \pm 40.7$ & $29.4 \pm 14.2$ & $95.1 \pm 58.0$ & $26.6 \pm 13.9$ & $\mathbf{83.5 \pm 25.8}$ & $\mathbf{26.4 \pm 10.5}$ & $129.0 \pm 53.8$ & $44.1 \pm 14.0$ \\
    Total  & $94.2 \pm 38.7$ & $31.7 \pm 14.1$ & $\mathbf{90.1 \pm 44.6}$ & $\mathbf{26.5 \pm 12.7}$ & $91.9 \pm 43.6$ & $29.0 \pm 13.2$ & $126.7 \pm 44.1$ & $44.6 \pm 13.1$ \\
    \bottomrule
  \end{tabular}
  }
\end{table*}

Appendix Tables~\ref{tab:planning_cost_ablation} and~\ref{tab:antmaze_teleport_benchmark} show the same efficiency trend in the full planner and in the broader stochastic teleport regime.

\section{Limitations}
\label{sec:limitations}
ChronoForest still depends on the quality of the learned temporal-distance prior: when long-range temporal-distance estimates are miscalibrated, the planner can spend expansion budget on poor bridges and preserve a suboptimal tentative route until enough bridge evidence accumulates. Algorithmically, the online route module is evaluated only with small anchor sets and fixed-endpoint Hamiltonian updates, so scaling to larger waypoint sets will likely require stronger pruning and more approximate combinatorial solvers. The theoretical analysis in Appendix~\ref{app:mtsp-theory} provides only a static sufficient justification for the route-certification threshold and the current-route/off-route budget split, not an exact exploration-optimal rule.

\section{Conclusion}
We studied long-horizon offline goal-reaching in the regime where short dataset fragments must be composed into efficient waypoint routes without a precomputed anchor-cost matrix. ChronoForest addresses this setting by coupling an Anchor-chaining tree diffusion planner, which uses temporal distance for local guidance and node evaluation, with an Online multi-tree orchestrator that re-solves the route from accumulated bridge evidence and reallocates search to route-critical anchor pairs. On OGBench AntMaze-Stitch, ChronoForest reaches 99.8\%, 99.3\%, and 99.5\% success on the medium, large, and giant splits, achieves its clearest gain on giant-stitch, and the Hamiltonian analyses show that online route re-solving corrects poor temporal orderings and improves route quality without paying the cost of exhaustive planning. These results support the paper's central claim that long-horizon offline route composition benefits most from coupling temporal-distance-guided local bridge search with explicit online route re-solving.

{\small
\bibliographystyle{plainnat}
\bibliography{references}
}

\appendix

\section{Denoised-Trajectory Guidance Formulation and Injection}
\label{app:guidance-injection}

\subsection{Scope and Setup}

Our planner's guidance does not train an additional conditional diffusion model,
but instead makes fine-grained adjustments to the sampling trajectory of a pretrained diffusion planner at test time.
This is \emph{training-free guidance}. The key point of this section is not the mere existence of the guidance objective,
but rather which representation the objective is injected into.

We assume a diffusion planner that predicts a denoised trajectory estimate from
a noisy latent, following Diffusion Forcing \citep{chen2024diffusionforcing}.
Given a noisy latent $X_t$, the denoiser outputs

\[
\hat X_0 = f_\theta(X_t, t)
\]

The sampler then advances using this denoised estimate together with the
implied noise. Our guidance therefore acts on clean-trajectory geometry
represented by $\hat X_0$, not on the noisy latent $X_t$ itself.

Let the planned trajectory of length $T$ be $X_0 = (s_1,\dots,s_T)$, and divide it into $K$ segments.
Denote the start and end indices of segment $k$ by $h_k$ and $\tau_k$, respectively.
Also, $P(s) \in \mathbb{R}^{d_p}$ is the projection that extracts only the coordinates to which guidance is actually applied from the state,
and, in the maze planning setting considered in this appendix, it extracts the 2-D spatial coordinate $(x,y)$ with $d_p = 2$.
The important point is that guidance does not act densely on every frame of the trajectory.
Only the boundary state $s_{h_k}, s_{\tau_k}$ and the projection $P(s)$ are targets of guidance.

This design is not a convenience for reducing test-time compute but rather an assumption of the method.
Because the constraints we seek to impose are sparse, have clear semantic meaning, and take the form of boundary-centered correction,
it is natural to update directly in clean trajectory space while preserving the intended support.

\subsection{Sparse Composite Guidance on the Denoised Trajectory}
\label{app:guidance-objective}

We define the following composite guidance score on the denoised estimate $\hat X_0$:

\[
\mathcal{U}(\hat X_0)
=
\lambda_{\text{goal}} \mathcal{U}_{\text{goal}}(\hat X_0)
+ \lambda_{\text{anchor}} \mathcal{U}_{\text{anchor}}(\hat X_0)
+ \lambda_{\text{rdf}} \mathcal{U}_{\text{rdf}}(\hat X_0).
\]

Each term depends only on sparse boundary coordinates.
For notational brevity, $\operatorname{sg}(\cdot)$ denotes the stop-gradient
copy of the current boundary quantity, making explicit which endpoint is treated
as fixed and which endpoint is updated.

\subsubsection{Goal-Reaching Term}

We assign a directional score so that each segment tail $s_{\tau_k}$ moves toward the current target anchor $g_k \in \mathbb{R}^{d_p}$
via the following directional score:

\[
\mathcal{U}_{\text{goal}}(\hat X_0)
=
\sum_{k=1}^{K}
\left\langle
P(\hat s_{\tau_k}),
\operatorname{sg}(d_k)
\right\rangle.
\]

Here, $V(s,g)$ provides the direction of progress from the current state $s$ toward the target anchor $g$ through
a fixed learned temporal-distance/value field, and
$d_k \in \mathbb{R}^{d_p}$ is the unit direction indicating which way to push the corresponding tail.
In the Near-target regime, we use the projected gradient of $V$,
whereas in the far-target regime, where that field is not yet reliable, we use a geometric fallback direction:

\[
d_k
=
\begin{cases}
\operatorname{norm}\!\bigl(P(\nabla_s V(\hat s_{\tau_k}, g_k))\bigr),
& \text{if } \rho_k = 1, \\[4pt]
\operatorname{norm}\!\bigl(g_k - P(\hat s_{\tau_k})\bigr),
& \text{if } \rho_k = 0.
\end{cases}
\]

Here, $\rho_k \in \{0,1\}$ is a reliability gate, and
$\operatorname{norm}(v)=v/(\|v\|+\varepsilon)$. Thus the goal term selects,
at each guidance step, either the learned directional field or a geometric
fallback, yielding a hybrid directional correction.

\subsubsection{Anchor / Boundary-Consistency Term}

To enforce connectivity across adjacent blocks, we align the head of the current segment with the tail of the previous segment:

\[
\mathcal{U}_{\text{anchor}}(\hat X_0)
=
- \sum_{k=2}^{K}
\left\|
P(\hat s_{h_k}) - \operatorname{sg}\!\bigl(P(\hat s_{\tau_{k-1}})\bigr)
\right\|_2^2.
\]

This term is not a regularizer that makes the entire trajectory globally smooth,
but rather a \emph{stitching regularizer} that reduces only the boundary mismatch at an adjacent segment interface.
It is also not a symmetric penalty but an asymmetric correction.
That is, the previous tail is fixed as the reference, and only the head of the newly generated segment is corrected.
This asymmetry allows the planner to avoid strongly perturbing the already formed prefix while
adjusting only the connectivity of the current frontier locally.

\subsubsection{RDF-Based Intra-Segment Expansion Term}

Within the same segment the tail is prevented from excessive collapse toward the head by a short-range repulsive potential corresponding to the RDF guidance term introduced in the main text:

\[
\mathcal{U}_{\text{rdf}}(\hat X_0)
=
\sum_{k=1}^{K}
\psi\!\left(
\left\|
P(\hat s_{\tau_k}) - \operatorname{sg}\!\bigl(P(\hat s_{h_k})\bigr)
\right\|_2
\right),
\]

where

\[
\psi(d)
=
-\frac{\exp\!\left(-d^2/(2\sigma^2)\right)}{d+\varepsilon}.
\]

The role of this term is not to increase diversity itself,
but to encourage each segment to embody actual spatial progress or local expansion.
Here again, the head is kept as the reference and only the tail receives correction, so that
the choice promotes outward growth of the current segment rather than translating the whole trajectory arbitrarily.

\subsection{Guidance Injection on the Denoised Estimate}
\label{app:guidance-denoised-update}

Under the denoised-trajectory parameterization above, the noisy latent is given by

\[
X_t = \sqrt{\bar\alpha_t} X_0 + \sqrt{1-\bar\alpha_t}\,\epsilon
\]

Because the score defined above depends only on projected boundary coordinates,
$\nabla_{\hat X_0}\mathcal{U}(\hat X_0)$ also has sparse support.
We therefore update the clean estimate as

\[
\tilde X_0
=
\hat X_0 + \eta_t \nabla_{\hat X_0}\mathcal{U}(\hat X_0)
\]

with a noise-dependent step size
\[
\eta_t
=
\min\!\left(
\eta_{\max},
\kappa \sqrt{\frac{1-\bar\alpha_t}{\bar\alpha_t}}
\right)
\]
This schedule limits over-correction in the high-noise regime.
Early denoising stages do not yet yield a stable semantic plan, whereas later
stages admit stronger structurally meaningful corrections.

The implied noise corresponding to the modified clean estimate is
\[
\tilde\epsilon_t
=
\frac{X_t - \sqrt{\bar\alpha_t}\tilde X_0}{\sqrt{1-\bar\alpha_t}}
\]
The sampler then performs the next denoising step using
$(\tilde X_0,\tilde\epsilon_t)$. Thus the correction target is the clean
trajectory hypothesis represented by $\hat X_0$, not the noisy latent itself.

\begin{algorithm}[t]
\caption{Denoised-Trajectory Guidance Injection}
\label{alg:denoised-guidance-injection}
\begin{algorithmic}[1]
\Require noisy sample $X_t$, denoiser $f_\theta$, guidance score $\mathcal{U}$, step size $\eta_t$
\State $\hat X_0 \gets f_\theta(X_t, t)$
\State $g_t \gets \nabla_{\hat X_0}\mathcal{U}(\hat X_0)$
\State $\tilde X_0 \gets \hat X_0 + \eta_t g_t$
\State $\tilde\epsilon_t \gets \bigl(X_t - \sqrt{\bar\alpha_t}\tilde X_0\bigr)/\sqrt{1-\bar\alpha_t}$
\State Advance the sampler using $(\tilde X_0,\tilde\epsilon_t)$
\end{algorithmic}
\end{algorithm}

\section{Why Inject Guidance into \texorpdfstring{$\hat X_0$}{X-hat-0} Instead of \texorpdfstring{$X_t$}{X_t}?}

The claim of this section is not that "direct $\hat X_0$ injection is always superior."
More narrowly, when \emph{guidance is defined on sparse boundary coordinates of the clean trajectory},
$X_t$-space injection may fail to preserve that sparsity and locality.

Let $g = \nabla_{\hat X_0}\mathcal{U}(\hat X_0)$, and let the denoiser Jacobian be

\[
J_\theta(X_t,t)
=
\frac{\partial \hat X_0(X_t,t)}{\partial X_t}
\]

Suppose instead that guidance is injected through $X_t$. Then, by the chain
rule,

\[
\nabla_{X_t}\mathcal{U}(\hat X_0(X_t,t))
=
J_\theta(X_t,t)^\top g
\]

If a perturbation $\delta X_t = \xi_t J_\theta^\top g$ is applied to $X_t$,
then the induced change in the clean estimate is, to first order,

\[
\delta \hat X_0
\approx
J_\theta(X_t,t)\,\delta X_t
=
\xi_t\, J_\theta(X_t,t) J_\theta(X_t,t)^\top g
\]

Thus, the originally intended sparse correction $g$ in clean space can be mixed
across other timesteps and other state dimensions through
$J_\theta J_\theta^\top$. By contrast, direct $\hat X_0$ injection uses
\[
\delta \hat X_0 = \eta_t g
\]
and therefore preserves the support of the boundary coordinates on which the
guidance is defined.

This distinction matters specifically for sparse planning guidance; it is not a
universal claim about all diffusion-guidance settings. Diffusion posterior
sampling \citep{chung2023dps}, for example, combines a full observation
likelihood with the diffusion prior, so dense coupling can align with its
objective. Our score instead evaluates sparse boundary structure on the clean
trajectory, making Jacobian-mediated mixing less faithful to the intended
correction.

\subsection{Why Denoised-Space Injection Matches Sparse Planning Guidance}

\subsubsection{Parameterization and Semantic Alignment}

If the main object predicted directly by the model is $\hat X_0$, then
auxiliary guidance is also most naturally interpreted in that representation.
Goal attraction, boundary stitching, and intra-segment expansion are all
constraints on the geometric structure that the denoised plan should have, not
on observability of the noisy latent itself. Accordingly, $\hat X_0$-space
injection aligns the semantics of the guidance signal with the correction
space. This rationale is consistent with prior denoised-domain guidance work
\citep{bansal2023universalguidance,nair2024dreamguider}.

\subsubsection{Locality Matters More Than Global Coupling}

Our guidance does not act densely over the full trajectory range.
It operates mainly only on segment head/tail states and their spatial coordinates.
What matters in this case is whether the correction remains \emph{only where it is needed}.
Direct $\hat X_0$ update preserves that locality by design,
whereas $X_t$-space update can induce non-local mixing because of Jacobian coupling effects.
In particular, because segment stitching is a local interface problem, non-local perturbations
risk worsening boundary consistency independently of goal progress.

\subsubsection{Maze Geometry as an Application-Specific Amplifier}

In maze-like planning, the locally useful direction changes rapidly near walls
or bottlenecks. In such high-curvature regions, the fidelity of which boundary
coordinate should move and in which direction becomes especially important.
Accordingly, it can be advantageous to deliver the intended correction directly
to the boundary in clean space. This is an application-specific amplifier of
the sparse-locality argument above, not a separate claim.

\section{Temporal-Distance and Directional Value Representations}
\label{app:value-representations}

The main text uses temporal distance only as a planning-time cost proxy; we do not claim that the learned value equals the ideal step-count distance exactly. For a goal state $g$, let
\[
T_g := \inf\{t \ge 0 \mid s_t = g\},
\qquad
d^*(s,g):=\min_{\pi}\mathbb{E}_{\pi}[T_g \mid s_0=s].
\]
The stitch experiments use a symmetric goal-conditioned value representor following prior offline goal-conditioned value learning \citep{park2024hilp,park2023hiql}. We parameterize a learned goal-conditioned value estimator with an encoder $\phi:\mathcal{S}\to\mathcal{Z}$ as
\[
V_{\mathrm{TD}}(s,g)=-\|\phi(s)-\phi(g)\|_2.
\]
The encoder is optimized with the goal-conditioned temporal-difference objective
\[
\mathcal{L}_{\mathrm{TD}}
=
\mathbb{E}_{s,s',g}
\left[
\ell_\tau^2
\left(
-\mathbf{1}(s \neq g)
+ \gamma \bar V_{\mathrm{TD}}(s',g)
- V_{\mathrm{TD}}(s,g)
\right)
\right],
\qquad
\ell_\tau^2(x)=|\tau-\mathbf{1}(x<0)|x^2.
\]
Under this discounted model, a state that is approximately $d^*(s,g)$ steps from the goal has value
\[
V_{\mathrm{TD}}(s,g)\approx -\frac{1-\gamma^{d^*(s,g)}}{1-\gamma},
\]
and we therefore use
\[
d_{\mathrm{TD}}(s,g):=\frac{\log\!\left(1+(1-\gamma)V_{\mathrm{TD}}(s,g)\right)}{\log \gamma}
\]
as the planning-time cost for target-conditioned guidance, node evaluation, and route-cost estimation.

For the teleport benchmark in Question~3, we replace the symmetric form above with a directional quasimetric value model so that asymmetric reachability is preserved at the representor level. Following prior quasimetric representation work \citep{mrn2023,iqe2023}, we use
\[
V_{\mathrm{dir}}(s,g)=-\|\operatorname{ReLU}(\psi(s)-\phi(g))\|_2.
\]
Here $\psi$ and $\phi$ are learned encoders of the current state and goal, respectively. In the teleport experiments, the planner uses $-V_{\mathrm{dir}}(s,g)$ everywhere the stitch pipeline would otherwise use $d_{\mathrm{TD}}(s,g)$. 

\section{Low-Level Planner State, Scoring, and Meeting Details}
\label{app:low-level-planner-details}

\subsection{Tree State and Target-Pointer Details}
\label{app:planner-state-details}

This subsection records the exact state tuple, cost-to-come decomposition, and root-target construction that are summarized in Section~\ref{sec:anchor-chaining-tree-diffusion-planner}.

For each anchor $a_i\in\mathcal{A}$, we maintain a rooted directed tree $\mathcal{T}_i=(\mathcal{V}_i,\mathcal{E}_i,n_i^{\mathrm{root}})$ whose root starts at $a_i$. This tree serves simultaneously as a search structure over shared prefixes and as the data structure used to compress multiple candidate continuations. Each node $n\in\mathcal{V}_i$ is represented by the tuple

    \[
n
=
\bigl(
\mathrm{pref}(n),\,
x(n),\,
g(n),\,
\mathrm{tgt}(n),\,
\mathcal{R}(n)
\bigr).
    \]

Here, $\mathrm{pref}(n)=\xi_n^{(1)}\oplus\cdots\oplus\xi_n^{(q_n)}$ is the materialized prefix, where $q_n$ is the number of stored local segments in that prefix.
Following the local-segment interface in Section~\ref{sec:diffusion-planner}, each subplan segment is written as $\xi_n^{(s)}=(x_n^{(s,0)},\dots,x_n^{(s,L)})\in\mathcal{S}^{L+1}$, where $x_n^{(s,0)}$ is the query boundary state and $(x_n^{(s,1)},\dots,x_n^{(s,L)})$ is the generated lookahead subplan.
We write $\operatorname{tail}(\mathrm{pref}(n))$ for the final state of the materialized prefix and abbreviate $x(n):=\operatorname{tail}(\mathrm{pref}(n))$.

    \[
g(n_i^{\mathrm{root}})=0,
\qquad
g(n)=g(\operatorname{par}(n))+d_{\mathrm{TD}}(x(\operatorname{par}(n)),x(n))
    \]

The cumulative prefix cost is given by the recurrence above, so $g(n)$ records the planning-time cost-to-come from the root to node $n$.
For every non-root node, $\operatorname{par}(n)$ denotes the unique parent of $n$ in $\mathcal{T}_i$.
Each node also maintains a target pointer into a counterpart tree and a representative set $\mathcal{R}(n)=\{\rho_{n,1},\dots,\rho_{n,M_n}\}$.
A representative $\rho_{n,c}$ stores the cached local segment $\operatorname{seg}(\rho_{n,c})=(x_{\rho_{n,c}}^{(0)},\dots,x_{\rho_{n,c}}^{(L)})$, the lookahead subplan $\operatorname{sub}(\rho_{n,c}):=(x_{\rho_{n,c}}^{(1)},\dots,x_{\rho_{n,c}}^{(L)})$, and the target-node slot $\mathrm{tgt}(\rho_{n,c})$ assigned by pair-conditioned target selection.
Because the boundary state $x_{\rho_{n,c}}^{(0)}$ is shared with the parent prefix tail, only $\operatorname{sub}(\rho_{n,c})$ is appended when the representative is later materialized as a child.
For the root-target rule below, let $\hat{\Pi}_t$ denote the current tentative joint route and let $\operatorname{nbr}_t(a_i)$ denote the anchor adjacent to $a_i$ on that route.

    \[
\mathrm{tgt}(n_i^{\mathrm{root}})
=
n_{\operatorname{nbr}_t(a_i)}^{\mathrm{root}}
    \]

At round $t$, the root target assignment uses the route neighbor $\operatorname{nbr}_t(a_i)$ on the current tentative joint route $\hat{\Pi}_t$, as shown above. Thereafter, the target pointer of each non-root node is refreshed separately for each ordered pair by the pair-conditioned target-selection rule in Section~\ref{sec:node-evaluation}.

\subsection{Pair-Conditioned Target Selection and Ranking Details}
\label{app:pair-conditioned-ranking-details}

This subsection records the full pair-conditioned target-selection objective together with the exact route-cost and uncertainty terms used by source-node ranking.

The representatives supplied to node evaluation are not newly generated in the current round; rather, they are the cached cluster representatives retained at each expandable leaf by the previous junction detector after pruning and clustering. Here, $\mathcal{P}_t$ denotes the set of ordered pairs selected by the high-level orchestrator for execution in the current round, and, for strong bidirectional search, it includes both $(i,j)$ and $(j,i)$ for the same anchor-pair candidate. Accordingly, within one round, $\mathcal{T}_i$ and $\mathcal{T}_j$ each serves once as the source tree, and pair-conditioned target selection and node evaluation are carried out separately for each direction. Concretely, when $(i,j)\in\mathcal{P}_t$, each representative $\rho\in\mathcal{R}(n)$ held by an expandable source node $n$ in $\mathcal{T}_i$ selects, within the counterpart tree $\mathcal{T}_j$, the target node that minimizes its estimated total cost. The resulting pair-specific target is then used directly for same-round lookahead subplan generation, as described in Section~\ref{sec:junction-detector}.

\[
\mathcal{P}_t \subseteq \mathcal{A}\times\mathcal{A},
\qquad
\mathcal{S}_t^{(i,j)}
:=
\left\{
 n\in\mathcal{V}_i
\;\middle|\;
n \text{ is expandable at round } t
\right\}
\quad\text{for each } (i,j)\in\mathcal{P}_t.
\]

Here, $\mathcal{S}_t^{(i,j)}$ is the \texttt{pair-conditioned candidate source set} that can actually be evaluated for pair $(i,j)$ at round $t$, and an expandable node means a leaf node whose representative set is nonempty. According to the definition in Section~\ref{sec:architecture-level-design}, each representative $\rho$ has cached segment $\operatorname{seg}(\rho)=\bigl(x_\rho^{(0)},\dots,x_\rho^{(L)}\bigr)$, lookahead subplan $\operatorname{sub}(\rho)=(x_\rho^{(1)},\dots,x_\rho^{(L)})$, and target-node slot $\mathrm{tgt}(\rho)$. For notational convenience, in this subsection we treat representative $\rho$ as a latent child node, define $\mathrm{pref}(\rho):=\mathrm{pref}(n)\oplus\operatorname{sub}(\rho)$, $x(\rho):=\operatorname{tail}(\operatorname{seg}(\rho))=\operatorname{tail}(\operatorname{sub}(\rho))$, and $g(\rho):=g(n)+d_{\mathrm{TD}}(x(n),x(\rho))$, so $g(\rho)$ is the cumulative cost of the latent child prefix $\mathrm{pref}(\rho)$.

The pair-conditioned estimated total cost of representative $\rho$ is defined as the sum of source-side cost-to-come, cross-tree contact cost, and target-side materialized prefix cost.

\[
D_i^{\mathrm{src}}(\rho)
:=
g(\rho),
\qquad
D_j^{\mathrm{tgt}}(m):=g(m).
\]

\[
\mathcal{J}_{i\to j}(\rho,m)
:=
D_i^{\mathrm{src}}(\rho)
+ h\!\left(
d_{\mathrm{TD}}\!\bigl(x(\rho),x(m)\bigr)
\right)
+ D_j^{\mathrm{tgt}}(m),
\qquad
h(x)=x+\alpha x^2
\]

\[
m_\rho^\star
\in
\arg\min_{m\in\mathcal{V}_j}
\mathcal{J}_{i\to j}(\rho,m).
\]

Here, $m_\rho^\star$ indicates which target node in counterpart tree $\mathcal{T}_j$ yields the smallest pair-conditioned estimated total cost when connected to representative endpoint $x(\rho)$. In Section~\ref{sec:online-multi-tree-orchestrator}, this $\mathcal{J}_{i\to j}$ is reused as the canonical pair-conditioned score when constructing the route-level direct-bridge entry $C_t(i,j)$ for ordered pair $(i,j)$.

\textbf{Estimated Total Cost}

For a given ordered pair $(i,j)$, we define the estimated total cost from a source node to a target node as follows.

A source node $n\in\mathcal{S}_t^{(i,j)}$ carries representatives $\mathcal{R}(n)=\{\rho_{n,1},\dots,\rho_{n,M_n}\}$. For each representative $\rho\in\mathcal{R}(n)$, we define $\mathrm{Cost}(\rho):=\mathcal{J}_{i\to j}(\rho,m_\rho^\star)$, and the node-level cost term is the minimum representative-wise cost

\[
\mathrm{Cost}(\rho)
=
\mathcal{J}_{i\to j}\!\bigl(\rho,m_\rho^\star\bigr),
\qquad
\mathrm{cost}(n)
:=
\min_{\rho\in\mathcal{R}(n)} \mathrm{Cost}(\rho).
\]

The normalized mass of the cluster summarized by representative $\rho$ is denoted by $p_\rho$.

\textbf{Uncertainty Penalty}

The uncertainty penalty $U(n)$ is a penalty for postponing the expansion of a node whose representative endpoints are widely dispersed. Let the subplan cluster summarized by Representative $\rho$ be $\mathcal{M}(\rho)$, and

\[
p_\rho
:=
\frac{|\mathcal{M}(\rho)|}
{\sum_{\rho'\in\mathcal{R}(n)} |\mathcal{M}(\rho')|}.
\]

Then $U(n)$ is defined as the following weighted endpoint-dispersion objective:

\[
U(n)
:=
\min_{c\in\mathcal{S}}
\sum_{\rho\in\mathcal{R}(n)} p_\rho\, d_{\mathrm{TD}}\!\bigl(x(\rho),c\bigr)^2.
\]

\subsection{Expansion and Meeting Details}
\label{app:expansion-meeting-details}

This subsection records the exact top-$K_{\mathrm{exp}}$ frontier update, child-promotion rule, and meeting criterion used by the low-level planner.

At round $t$, node expansion selects the top-$K_{\mathrm{exp}}$ parent nodes from each candidate source set $\mathcal{S}_t^{(i,j)}$ according to the score.

\[
\mathcal{N}_t^{(i,j)}
=
\operatorname{TopK}\!\left(
\mathcal{S}_t^{(i,j)},
\mathrm{score}(\cdot),
K_{\mathrm{exp}}
\right).
\]

That is, $\mathcal{N}_t^{(i,j)}\subseteq\mathcal{S}_t^{(i,j)}$ is the set of source parents for which ordered pair $(i,j)$ actually performs materialization and lookahead expansion in the current round. The ranking is computed independently within each pair-conditioned pool $\mathcal{S}_t^{(i,j)}$, rather than globally across all trees, so the expand set reflects the local route evidence relevant to the selected ordered pair.

\textbf{Node Expansion}

For each selected parent node $n$ and representative $\rho_c\in\mathcal{R}(n)$, let $n_c^+$ denote the promoted child leaf. Because $\rho_c$ stores both the cached local segment $\operatorname{seg}(\rho_c)$ and the lookahead subplan $\operatorname{sub}(\rho_c)$, promotion appends only the lookahead subplan to the parent prefix; the boundary state $x_{\rho_c}^{(0)}$ is already shared with the parent tail $x(n)$.

\[
\mathrm{pref}(n_c^+)
=
\mathrm{pref}(n)\oplus \operatorname{sub}(\rho_c),
\qquad
\mathrm{tgt}(n_c^+)=\mathrm{tgt}(\rho_c),
\qquad
x(n_c^+)=\operatorname{tail}\!\left(\operatorname{sub}(\rho_c)\right).
\]

Accordingly, only $\operatorname{sub}(\rho_c)$ is newly materialized, the target node is inherited unchanged, and the cached representative segment is reused without separate resampling, which reduces inference-time cost.

\textbf{Meeting Detection}

Subsequently, in meeting detection, we determine whether each promoted child leaf $n_c^+$ has achieved meeting with the assigned target node. Here, $m:=\mathrm{tgt}(n_c^+)$, and $d_{\mathrm{exec}}$ is the task-specific state distance in execution space.

\[
(\hat u,\hat v)
\in
\arg\min_{u \in \mathrm{pref}(n_c^+),\; v \in \mathrm{pref}(m)}
d_{\mathrm{exec}}(u,v),
\qquad
\Delta_{\mathrm{meet}}(n_c^+,m)
:=
d_{\mathrm{exec}}(\hat u,\hat v).
\]

If $\Delta_{\mathrm{meet}}(n_c^+,m)\le\delta_{\mathrm{meet}}$, it is treated as meeting. In Section~\ref{sec:online-multi-tree-orchestrator}, the round-$t$ materialization indicator $\chi_t(i,j)$ and the accepted set $A_t^{\mathrm{meet}}$ for ordered pair $(i,j)$ are derived from this meeting event.

\section{Pruning and Representative Selection Details}
\label{app:pruning-representatives}

For each selected parent node $n\in\mathcal{N}_t^{(i,j)}$, the junction detector samples candidate local segments
\[
\xi=(x^{(0)},x^{(1)},\dots,x^{(L)}),
\qquad
x^{(0)}=x(n).
\]
Using the same conservative bidirectional distance as in Section~\ref{sec:junction-detector},
\[
\bar d(x,y):=\max\{d_{\mathrm{TD}}(x,y),\,d_{\mathrm{TD}}(y,x)\},
\]
a sampled segment is retained only if it satisfies the three pruning tests
\[
\max_{\ell\in\{1,\dots,L\}} \bar d\!\bigl(x^{(\ell-1)},x^{(\ell)}\bigr) \le \tau_{\mathrm{cont}}
\qquad\text{(local continuity)},
\]
\[
d_{\mathrm{TD}}\!\bigl(x^{(0)},x^{(L)}\bigr) \ge \tau_{\mathrm{prog}}
\qquad\text{(progress)},
\]
\[
\min_{u\in \mathrm{pref}(n)} d_{\mathrm{exec}}\!\bigl(u,x^{(L)}\bigr) \ge \tau_{\mathrm{ov}}
\qquad\text{(overlap)}.
\]
Only the samples satisfying all three inequalities are passed to complete-linkage clustering and partitioned into clusters $\mathcal{M}_c$.

To select one stored representative from each cluster, we reuse the world-coordinate projection $P(s)$ introduced above and subsample training states from the corresponding dataset split to build a KDE support estimate. Let $\{z_r\}_{r=1}^{N_{\mathrm{KDE}}}$ denote the subsampled training states and let $h>0$ be the fixed KDE bandwidth. The support estimate is
\[
\hat p_{\mathrm{data}}(x)
:=
\frac{1}{N_{\mathrm{KDE}}(2\pi h^2)^{d_p/2}}
\sum_{r=1}^{N_{\mathrm{KDE}}}
\exp\!\left(
-\frac{\|P(x)-P(z_r)\|_2^2}{2h^2}
\right).
\]
For a feasible segment $\xi=(x^{(0)},\dots,x^{(L)})$, the newly expanded local frames are indexed by
\[
\mathcal{L}_{\mathrm{new}}(\xi):=\{1,\dots,L\}.
\]
The stored representative is then chosen by the same maximin support rule used in the implementation,
\[
\xi_c^\star
\in
\arg\max_{\xi\in\mathcal{M}_c}
\min_{\ell\in\mathcal{L}_{\mathrm{new}}(\xi)}
\log \hat p_{\mathrm{data}}\!\bigl(\xi^{(\ell)}\bigr),
\qquad
\xi_c^\star=(x_c^{(0)},\dots,x_c^{(L)}).
\]
The selected segment is materialized as the stored representative record through $\operatorname{seg}(\rho_c):=\xi_c^\star$ and $\operatorname{sub}(\rho_c):=(x_c^{(1)},\dots,x_c^{(L)})$. On the teleport benchmark, the same pruning and representative-selection logic is used after replacing $d_{\mathrm{TD}}$ by $-V_{\mathrm{dir}}$ in the relevant directional comparisons.

\section{Theoretical Justification of the mTSP Algorithm}
\label{app:mtsp-theory}

In this section, we analyze a multi-agent setting in which \textbf{$M \ge 1$ agents} partition-cover $K$ shared waypoint
in a multi-agent setting, and study the route computation defined on the Floyd-Warshall closure of the direct bridge matrix
through the all-hop route computation abstraction.
When $M=1$, this reduces to the classical single-agent Hamiltonian path problem.
We first organize the relaxed all-hop core,
and then conditionally discuss the effects of the accepted-edge constraint and the non-starvation augmentation in later sections.

The direct results of this section are the following four claims:

\begin{enumerate}
  \item We show that the direct bridge cost estimate $C_t(i,j)$ is an upper envelope of the true direct cost $c^\star_{ij}$.
  \item We show that the error of the Floyd-Warshall closure $\widetilde{C}_t$ is controlled by the sum of the direct edge errors that compose the true closure path.
  \item As a result, we present the error of the all-hop Hamiltonian route solved by this method and its exact recovery condition. In particular, we show that the suboptimality bound for the joint route of $M$ agents scales with the factor $(K+M)$.
  \item We explain how the accepted-edge constraint and the small exploration floor connect to this relaxed core.
\end{enumerate}

Importantly, the guarantee in this section concerns \textbf{all-hop route computation},
not the entire final assembly trajectory as a whole.

\subsection{Problem Setting}
\label{app:mtsp-problem-setting}

Suppose that each of the $M\ge 1$ agents has start $S_m$ and goal $G_m$.
When $M=1$, this reduces to the classical single-agent Hamiltonian path problem.

Define the Anchor set as

\[
\mathcal{A}
=
\{S_1,\dots,S_M\}
\cup
\{W_1,\dots,W_K\}
\cup
\{G_1,\dots,G_M\}
\]

which contains a total of $K+2M$ anchors.

\textbf{Joint route.} In the joint route $\boldsymbol{\pi}=(\pi_1,\dots,\pi_M)$ of $M$ agents, each $\pi_m$ is
the sub-route of agent $m$, with $\pi_m[0]=S_m$ and $\pi_m[-1]=G_m$.
The $K$ waypoint $\{W_1,\dots,W_K\}$ are \textbf{partition-covered (partition cover)} by the $M$ agents.
That is, each $W_k$ is included in exactly one $\pi_m$.
If agent $m$ is assigned $K_m$ waypoint slots, then $\sum_{m=1}^M K_m=K$.
Joint route space is

\[
\Pi_M\!\left(\mathcal{A};\{S_m\},\{G_m\}\right)
=
\left\{
\boldsymbol{\pi}=(\pi_1,\dots,\pi_M)
\;\middle|\;
\begin{aligned}
&\pi_m[0]=S_m,\;\pi_m[-1]=G_m\;\forall m,\\
&\{W_k:\exists m,\,W_k\in\pi_m\}=\{W_1,\dots,W_K\},\\
&\pi_m\cap\pi_{m'}=\emptyset\;\forall m\neq m'
\end{aligned}
\right\}
\]

This is the feasible set.

\textbf{Adjacent pairs.} Let $\mathrm{Adj}(\boldsymbol{\pi})=\bigcup_m\mathrm{Adj}(\pi_m)$ denote the set of adjacent anchor pairs in the joint route.

\[
|\mathrm{Adj}(\boldsymbol{\pi})|
=
\sum_{m=1}^M (K_m+1)
=
K+M
\tag{0}
\]

This holds true for any $\boldsymbol{\pi}\in\Pi_M$.

Let the set of executable segments that connect $a_i$ directly to $a_j$ without passing through any Hidden intermediate anchor

\[
\mathcal{B}_{ij}
:=
\left\{
\xi \;\middle|\;
\xi \text{ connects } a_i \text{ to } a_j
\text{ without passing through another anchor}
\right\}
\]

be denoted as follows. If each segment $\xi$ has objective value $J(\xi)$, the true direct cost is

\[
c^\star_{ij}
:=
\inf_{\xi\in\mathcal{B}_{ij}} J(\xi)
\]

This in turn defines the true direct-cost matrix as

\[
C^\star(i,j):=c^\star_{ij}
\]

We use this notation.

The current method evaluates a route not with the direct matrix itself but with its all-hop closure.
Accordingly, we define the true all-hop shortest-path matrix as

\[
\widetilde{C}^\star(i,j)
:=
\min_{
\substack{
L\ge 1,\\
\zeta_0=i,\;\zeta_L=j,\\
\zeta_1,\dots,\zeta_{L-1}\in\mathcal{A}
}
}
\sum_{\ell=0}^{L-1} C^\star(\zeta_\ell,\zeta_{\ell+1})
\tag{1}
\]

Here, $L$ is the hop count of the path, and the intermediate anchor is selected from the entirety of $\mathcal{A}$, which contains $K+2M$ anchors.
This is equivalent to the Floyd-Warshall closure of the true direct-cost matrix $C^\star$.

We now consider the objective

\[
\widetilde{J}^\star(\boldsymbol{\pi})
:=
\sum_{(i,j)\in\mathrm{Adj}(\boldsymbol{\pi})}\widetilde{C}^\star(i,j)
\]

and define

\[
\widetilde{\boldsymbol{\pi}}^\star
\in
\arg\min_{\boldsymbol{\pi}\in\Pi_M(\mathcal{A};\{S_m\},\{G_m\})} \widetilde{J}^\star(\boldsymbol{\pi})
\tag{2}
\]

which we call the true all-hop optimal joint route.

At Round $t$, we define the newly evaluated direct bridge candidate cost for anchor pair $(i,j)$ as

\[
\widehat{c}^{\,ij}_t(n,m)
=
D_i^{\mathrm{src}}(n)
+
D_j^{\mathrm{tgt}}(m)
+
h\!\left(\widehat{g}_t(n,m)\right),
\qquad
h(x)=x+\alpha x^2,\;\alpha>0
\tag{3}
\]

where

\begin{itemize}
  \item $D_i^{\mathrm{src}}(n)$ is the realized prefix cost from anchor $a_i$ to source-side node $n$;
  \item $D_j^{\mathrm{tgt}}(m)$ is the realized prefix cost from target-side node $m$ to anchor $a_j$;
  \item $\widehat{g}_t(n,m)$ is the predictor defined between the two node endpoint values, namely the raw residual-gap predictor
\end{itemize}

These are the components. In this appendix, $\widehat{c}_t^{\,ij}(n,m)$ serves as the abstract direct-bridge score corresponding to the pair-conditioned score $\mathcal{J}_{i\to j}$ used in Section~\ref{sec:anchor-chaining-tree-diffusion-planner}.

Now let $\mathcal{E}^{\mathrm{new}}_i(t)$ denote the set of newly expanded source-side nodes in tree $i$ at round $t$.
We write it as $\mathcal{E}^{\mathrm{new}}_i(t)$.
For each $n\in\mathcal{E}^{\mathrm{new}}_i(t)$, considering all $j\neq i$,
we evaluate a candidate for each ordered pair $(i,j)$ and $(j,i)$.
Accordingly, let $\mathcal{N}^{\mathrm{new}}_{ij}(t)$
denote the set of candidate node-pairs generated for the ordered pair $(i,j)$ during round $t$.
At initialization, we fill the direct matrix with root-to-root bridge scores,
and thereafter update an entry only when a lower direct bridge cost is found.
Therefore, for $t\ge 1$, the direct bridge cost matrix is

\[
C_t(i,j)
=
\min\Biggl\{
C_{t-1}(i,j),\;
\min_{(n,m)\in\mathcal{N}^{\mathrm{new}}_{ij}(t)}
\widehat{c}^{\,ij}_t(n,m)
\Biggr\}
\tag{4}
\]

This is defined as follows. If no new candidate is available, the inner minimum is interpreted as $+\infty$.
In other words, $C_t(i,j)$ stores the cumulative minimum over the direct bridge candidate costs observed so far.
The route solver always uses the all-hop closure of this matrix.

Accordingly, the all-hop shortest-path matrix at round $t$ is

\[
\widetilde{C}_t(i,j)
:=
\min_{
\substack{
L\ge 1,\\
\zeta_0=i,\;\zeta_L=j,\\
\zeta_1,\dots,\zeta_{L-1}\in\mathcal{A}
}
}
\sum_{\ell=0}^{L-1} C_t(\zeta_\ell,\zeta_{\ell+1})
\tag{5}
\]

This is defined as follows. The intermediate anchor is chosen freely from the whole of $\mathcal{A}$, that is, from the sets of all agents.
This is equivalent to the Floyd-Warshall closure of the direct bridge matrix $C_t$.

\textbf{Relaxed closure interpretation.} In equation (5), the intermediate
anchor is selected while ignoring agent boundaries. Therefore,
$\widetilde{C}_t(i,j)$ is a relaxed all-hop objective that is not restricted
to the feasible joint route space and may not coincide with the cost of
actually executable intra-agent segments. A later subsection explains how this
relaxed objective relates to executable routes through the accepted-edge
constraint.

The tentative joint route at Round $t$ is the same route object as $\hat{\Pi}_t$ in the main method, and in this appendix we denote it by $\boldsymbol{\pi}_t$.

\[
\boldsymbol{\pi}_t
\in
\arg\min_{\boldsymbol{\pi}\in\Pi_M(\mathcal{A};\{S_m\},\{G_m\})}
\sum_{(i,j)\in\mathrm{Adj}(\boldsymbol{\pi})}\widetilde{C}_t(i,j)
\tag{6}
\]

This optimization simultaneously determines the waypoint assignment across
agents and the permutation of each agent sub-route.

In addition to the relaxed solver we may also consider a constrained solver that satisfies the commitment set $E_{\mathrm{acc}}^t$.
For this purpose, let the commitment set be $E_{\mathrm{acc}}^t$.

\[
\Pi_t^{\mathrm{acc}}
:=
\left\{
\boldsymbol{\pi}\in\Pi_M(\mathcal{A};\{S_m\},\{G_m\})
\;\middle|\;
E_{\mathrm{acc}}^t\subseteq \mathrm{Adj}(\boldsymbol{\pi})
\right\}
\tag{6a}
\]

We also define the event that the true all-hop optimal joint route has not yet been excluded by these commitments.

\[
\mathcal{E}_t^\star
:=
\left\{
E_{\mathrm{acc}}^t\subseteq \mathrm{Adj}(\widetilde{\boldsymbol{\pi}}^\star)
\right\}
\tag{6b}
\]

The core theorems in Sections 4 and 5 analyze the relaxed solver in
equation~(6). A later subsection explains how this comparison transfers to the
constrained solver under the event $\mathcal{E}_t^\star$.

Finally, for each pair $(i,j)$, let $P^\star_{ij}$ be a simple shortest-path
witness attaining the true closure value $\widetilde{C}^\star(i,j)$.
\[
P^\star_{ij}
=
(v^{ij}_0=i,v^{ij}_1,\dots,v^{ij}_{H^\star_{ij}}=j)
\]
Because edge costs are nonnegative, we may choose $P^\star_{ij}$ to be simple
after removing cycles. Its intermediate anchors may lie anywhere in
$\mathcal{A}$, and since it is a simple path over $K+2M$ anchors,
\[
1\le H^\star_{ij}\le K+2M-1
\tag{7}
\]
equation~(7) follows.

\subsection{Direct-Edge Assumptions and Envelope}

\textbf{Assumption 0 (Informative Candidate Support).}
For every route-relevant direct pair $(i,j)$, the exploration process
eventually generates at least one candidate pair $(n,m)$ whose realized
prefixes $D_i^{\mathrm{src}}(n)$ and $D_j^{\mathrm{tgt}}(m)$ lie on, or
sufficiently near, a direct segment attaining $c^\star_{ij}$.

This is a structural coverage assumption on candidate generation, not a theorem
derived from the diffusion model or the search dynamics. In the multi-agent
setting, it applies both to intra-agent and inter-agent anchor pairs; feasibility
of the executed route is handled separately through the accepted-edge
constraint introduced later in this appendix.

\textbf{Assumption 1 (Candidate-level prefix consistency).}
For every direct pair $(i,j)$, round $t$, and candidate
$(n,m)\in\mathcal{N}^{\mathrm{new}}_{ij}(t)$, the realized prefixes can be
interpreted as the source-side and target-side portions of some executable
direct segment. Therefore
\[
g_t^{\star,ij}(n,m)
:=
c^\star_{ij}
- D_i^{\mathrm{src}}(n)
- D_j^{\mathrm{tgt}}(m)
\ge 0
\tag{8}
\]
the quantity in equation~(8) is well defined.

This is a structural prefix-realizability assumption, not a calibration
assumption on the predictor. Under this assumption, we analyze the envelope
induced by the direct-bridge score.

\textbf{Assumption 2 (Conditional finite-horizon upper-envelope event).}
Fix a planning horizon $T$ and condition on the finite set of candidate pairs generated up to round $T$ in the planning instances under consideration. With probability at least $1-\delta_{\mathrm{cal}}$, for every direct pair $(i,j)$, round $t\le T$, and candidate $(n,m)\in\mathcal{N}^{\mathrm{new}}_{ij}(t)$ generated within this horizon,
\[
h\!\left(\widehat{g}_t(n,m)\right)
\ge
g_t^{\star,ij}(n,m)
\tag{9}
\]
the corrected residual score upper-bounds the remaining direct cost on this finite event.

Importantly, this is a conditioned route-analysis assumption, not a uniform asymptotic guarantee over all future rounds or all possible candidates. It concerns the relationship between $\widehat{g}_t(n,m)$ itself and $g_t^{\star,ij}(n,m)$, and does not require any pointwise ordering of the raw predictor. In other words, $\widehat{g}_t(n,m)$ may overestimate or underestimate $g_t^{\star,ij}(n,m)$; we assume only that the corrected quantity $h(\widehat{g}_t(n,m))$ serves as an upper envelope for the remaining direct cost within the finite-horizon event above.

We define the candidate-level error as

\[
\eta_t^{ij}(n,m)
:=
h\!\left(\widehat{g}_t(n,m)\right)-g_t^{\star,ij}(n,m)
\ge 0
\tag{10}
\]

The following proposition characterizes the resulting candidate-level
direct-bridge envelope.

\textbf{Proposition 1 (Candidate-level direct-edge envelope).}
If Assumption 1 and 2 hold, then for every direct pair $(i,j)$, every round $t$,
and for every candidate $(n,m)\in\mathcal{N}^{\mathrm{new}}_{ij}(t)$,

\[
\widehat{c}^{\,ij}_t(n,m)
=
c^\star_{ij}+\eta_t^{ij}(n,m),
\qquad
c^\star_{ij}\le \widehat{c}^{\,ij}_t(n,m).
\tag{11}
\]

\emph{Proof.}
Combining equation (3) and equation (8) gives

\[
\widehat{c}^{\,ij}_t(n,m)
=
D_i^{\mathrm{src}}(n)+D_j^{\mathrm{tgt}}(m)+h\!\left(\widehat{g}_t(n,m)\right)
=
c^\star_{ij}+\eta_t^{ij}(n,m).
\]

The conclusion follows from equation (10): $\square$

We define the matrix-level direct-edge error as

\[
\varepsilon_{ij}(t):=C_t(i,j)-c^\star_{ij}
\tag{12}
\]

The matrix-entry error $\varepsilon_{ij}(t)$ differs from the candidate-level slack $\eta_t^{ij}(n,m)$.

The former refers to the current matrix entry $C_t(i,j)$, whereas the latter refers to a specific candidate $(n,m)$. If $w_t^{ij}$ denotes the witness candidate that currently attains $C_t(i,j)$, then Proposition 1 gives

\[
\varepsilon_{ij}(t)=\eta_t^{ij}(w_t^{ij})
\]

Thus, the matrix-level error is exactly the slack of the current witness candidate. Equivalently, $\varepsilon_{ij}(t)$ is the running minimum candidate slack among the candidates observed so far for pair $(i,j)$.

\textbf{Proposition 2 (Matrix-entry direct-edge envelope).}
If Assumptions 1 and 2 hold for the initial root-to-root candidate and every subsequent update candidate, then for every pair $(i,j)$ and round $t$,

\[
\varepsilon_{ij}(t)\ge 0,
\qquad
c^\star_{ij}\le C_t(i,j).
\tag{13}
\]

\emph{Proof.}
At initialization, the stored entry equals the direct-bridge candidate cost and therefore upper-bounds $c^\star_{ij}$ by Proposition 1. Each update in equation (4) replaces the current entry by the minimum of the old entry and a new candidate cost, both of which upper-bound $c^\star_{ij}$. The upper-bound property is therefore preserved at every round, which implies $\varepsilon_{ij}(t)\ge 0$ and $C_t(i,j)\ge c^\star_{ij}$ for all $t$. $\square$

\subsubsection{Practical Tighter Bound on Direct-Edge Entries}

The upper-envelope guarantee above is safe but can be loose. In the small-gap regime, a sharper bound is obtained by combining the calibration envelope for $\widehat{g}_t$ with a local absolute-error control on the true residual gap $g_t^{\star,ij}$.

\textbf{Assumption 3 (Local calibration near small predicted gaps).}
With probability at least $1-\delta_{\mathrm{loc}}$, there exists a nondecreasing function $\varepsilon_{\mathrm{loc}}:[0,\infty)\to[0,\infty)$ such that $\lim_{\delta\downarrow 0}\varepsilon_{\mathrm{loc}}(\delta)=0$, and for every direct pair $(i,j)$, every round $t$, and every candidate $(n,m)\in\mathcal{N}^{\mathrm{new}}_{ij}(t)$,

\[
\widehat{g}_t(n,m)\le \delta
\;\Longrightarrow\;
\left|
\widehat{g}_t(n,m)-g_t^{\star,ij}(n,m)
\right|
\le
\varepsilon_{\mathrm{loc}}(\delta)
\tag{14}
\]

Assumption 3 does not require $\widehat{g}_t$ to be consistently above or below $g_t^{\star,ij}$. It only requires that, once the predicted gap is sufficiently small, the absolute calibration error is also small.

In the remainder of this subsection, we work on the joint event

\[
\mathcal{E}_{\mathrm{joint}}
:=
\mathcal{E}_{\mathrm{cal}}\cap \mathcal{E}_{\mathrm{loc}}
\]

where $\mathcal{E}_{\mathrm{cal}}$ is the event in Assumption 2 and $\mathcal{E}_{\mathrm{loc}}$ is the event in Assumption 3. By the union bound, $\Pr(\mathcal{E}_{\mathrm{joint}})\ge 1-(\delta_{\mathrm{cal}}+\delta_{\mathrm{loc}})$.

Let $w_t^{ij}=(n_t^{ij},m_t^{ij})$ denote the witness candidate that attains $C_t(i,j)$ at round $t$. Suppose this witness also satisfies

\[
\widehat{g}_t(n_t^{ij},m_t^{ij})\le \tau
\tag{15}
\]

For brevity, below we evaluate candidate-level quantities at $w_t^{ij}$. Because the raw residual-gap predictor is a temporal-distance estimate, it is nonnegative; together with equation (15), this gives $\widehat{g}_t(w_t^{ij})^2\le\tau^2$.

On $\mathcal{E}_{\mathrm{joint}}$, Assumptions 2 and 3 therefore imply

\[
\eta_t^{ij}(w_t^{ij})
\le
\alpha \widehat{g}_t(w_t^{ij})^2
+
\left|
\widehat{g}_t(w_t^{ij})-g_t^{\star,ij}(w_t^{ij})
\right|
\le
\varepsilon_{\mathrm{loc}}(\tau)+\alpha\tau^2
\tag{16}
\]

Because $C_t(i,j)$ is realized by the witness cost, equation (12) yields

\[
\varepsilon_{ij}(t)\le \varepsilon_{\mathrm{loc}}(\tau)+\alpha\tau^2
\tag{17}
\]

for every pair whose current witness satisfies equation (15).

In practice, $\varepsilon_{\mathrm{loc}}(\tau)$ can be estimated on a calibration set by the upper envelope of the absolute residual error over samples with $\widehat{g}\le\tau$. Denoting this estimate by $\widehat{\varepsilon}_{\mathrm{loc}}(\tau)$ yields the engineering approximation

\[
\varepsilon_{ij}(t)\;\lesssim\;
\widehat{\varepsilon}_{\mathrm{loc}}(\tau)+\alpha\tau^2
\]

This surrogate does not assume idealized systematic underestimation; it only uses the empirically measured small-gap absolute-error envelope to track the direct-edge error more tightly in the local regime.

\subsection{Floyd-Warshall Closure Error}

We define the closure-level error as

\[
\widetilde{\varepsilon}_{ij}(t)
:=
\widetilde{C}_t(i,j)-\widetilde{C}^\star(i,j)
\tag{18}
\]

We now upper-bound the closure-level error in terms of the direct-edge errors along the true closure witness path.

\textbf{Proposition 3 (Closure-pair error bound).}
If Assumptions 1 and 2 hold, then for every pair $(i,j)$,

\[
0\le \widetilde{\varepsilon}_{ij}(t)
\le
\sum_{m=0}^{H^\star_{ij}-1}
\varepsilon_{v^{ij}_m,v^{ij}_{m+1}}(t)
\tag{19}
\]

In particular, the closure error is bounded by the sum of the direct-edge errors along the true closure witness path.

\emph{Proof.}
Proposition 2 gives $C_t(u,v)\ge C^\star(u,v)$ for every direct edge $(u,v)$, so taking the minimum over the same path family implies

\[
\widetilde{C}_t(i,j)\ge \widetilde{C}^\star(i,j)
\]

For the upper bound, evaluate $\widetilde{C}_t(i,j)$ on the true closure witness path $P^\star_{ij}$ to obtain

Equation (4) then gives

\[
\widetilde{C}_t(i,j)
\le
\sum_{m=0}^{H^\star_{ij}-1} C_t(v^{ij}_m,v^{ij}_{m+1}).
\]

Substituting $\varepsilon_{uv}(t)=C_t(u,v)-c^\star_{uv}$ into the right-hand side yields

\[
\widetilde{C}_t(i,j)
\le
\sum_{m=0}^{H^\star_{ij}-1}
\Bigl(
c^\star_{v^{ij}_m,v^{ij}_{m+1}}
+
\varepsilon_{v^{ij}_m,v^{ij}_{m+1}}(t)
\Bigr)
=
\widetilde{C}^\star(i,j)
+
\sum_{m=0}^{H^\star_{ij}-1}
\varepsilon_{v^{ij}_m,v^{ij}_{m+1}}(t).
\]

Combining the lower and upper bounds gives equation (19). $\square$

\subsubsection{Threshold Bounds on Closure Entries}

If every direct edge on the true closure witness path $P^\star_{ij}$ satisfies the premises of Assumption 3 and the witness-threshold condition in equation (15), then combining equations (17) and (19) yields

\[
\widetilde{\varepsilon}_{ij}(t)
\le
H^\star_{ij}\,\bigl(\varepsilon_{\mathrm{loc}}(\tau)+\alpha\tau^2\bigr)
\tag{20}
\]

Equation (20) lifts the direct-edge small-gap bound to the closure entry by summing the per-edge witness errors along the true closure witness path.

This bound is pair-specific through the hop count $H^\star_{ij}$ and therefore depends on how many direct edges the true closure witness path uses.

\textbf{Scaling for $M$.} The factor $H^\star_{ij}$ in equation (20) is the hop count of the true closure witness path for pair $(i,j)$, so it is pair-specific rather than directly controlled by $M$. However, using the worst-case bound $H^\star_{\max}\le K+2M-1$ from equation (29) shows that the worst-case closure error can scale linearly with the total number of anchor states. As $M$ grows, maintaining the same route-certification accuracy therefore requires more exploration.

\subsection{Route-Level Comparison Theorem}
\label{app:mtsp-route-comparison}

We define the worst-case closure error over the adjacency set of the true all-hop optimal joint route as

\[
\widetilde{\bar{\varepsilon}}_t
:=
\max_{(i,j)\in\mathrm{Adj}(\widetilde{\boldsymbol{\pi}}^\star)}
\widetilde{\varepsilon}_{ij}(t)
\tag{21}
\]

We now convert the pairwise closure error into a route-level suboptimality bound.

\textbf{Theorem 1 (All-hop route suboptimality).}
If Assumptions 1 and 2 hold, then with high probability $1-\delta_{\mathrm{cal}}$,

\[
\widetilde{J}^\star(\boldsymbol{\pi}_t)
\le
\widetilde{J}^\star(\widetilde{\boldsymbol{\pi}}^\star)
+
(K+M)\widetilde{\bar{\varepsilon}}_t
\tag{22}
\]

where $\widetilde{\bar{\varepsilon}}_t$ is the worst closure error over the adjacencies of the true all-hop optimal joint route.

\emph{Proof.}
Because $\boldsymbol{\pi}_t$ minimizes the route objective built from $\widetilde{C}_t$ in equation (6),

\[
\sum_{(i,j)\in\mathrm{Adj}(\boldsymbol{\pi}_t)}\widetilde{C}_t(i,j)
\le
\sum_{(i,j)\in\mathrm{Adj}(\widetilde{\boldsymbol{\pi}}^\star)}\widetilde{C}_t(i,j).
\tag{23}
\]

Proposition 3 also gives $\widetilde{C}_t(i,j)\ge \widetilde{C}^\star(i,j)$ for every pair, so

\[
\widetilde{J}^\star(\boldsymbol{\pi}_t)
\le
\sum_{(i,j)\in\mathrm{Adj}(\boldsymbol{\pi}_t)}\widetilde{C}_t(i,j).
\tag{24}
\]

On the other hand, $\widetilde{C}_t(i,j)\le \widetilde{C}^\star(i,j)+\widetilde{\varepsilon}_{ij}(t)$ for every pair. Since $|\mathrm{Adj}(\widetilde{\boldsymbol{\pi}}^\star)|=K+M$ by equation (0),

\[
\sum_{(i,j)\in\mathrm{Adj}(\widetilde{\boldsymbol{\pi}}^\star)}\widetilde{C}_t(i,j)
\le
\widetilde{J}^\star(\widetilde{\boldsymbol{\pi}}^\star)
+
(K+M)\widetilde{\bar{\varepsilon}}_t.
\tag{25}
\]

Combining equations (23)--(25) yields equation (22). $\square$

This theorem is stated on the Floyd-Warshall closure of the direct bridge matrix $C_t$ and on the joint Hamiltonian route induced by that closure. When $M=1$, it reduces to the single-agent form with $(K+M)=(K+1)$.

\subsection{Margin-Based Exact Recovery}

We define the all-hop route margin as

\[
\widetilde{\Delta}_{\mathrm{route}}
:=
\min_{\boldsymbol{\pi}\neq\widetilde{\boldsymbol{\pi}}^\star}
\Bigl(
\widetilde{J}^\star(\boldsymbol{\pi})-\widetilde{J}^\star(\widetilde{\boldsymbol{\pi}}^\star)
\Bigr)
>
0
\tag{26}
\]

Corollary 1 turns the route margin into a sufficient exact-recovery condition.

\textbf{Corollary 1 (Exact recovery of the all-hop joint route).}
Suppose that $\widetilde{\boldsymbol{\pi}}^\star$ is the unique minimizer of $\widetilde{J}^\star$. If

\[
(K+M)\widetilde{\bar{\varepsilon}}_t
<
\widetilde{\Delta}_{\mathrm{route}}
\tag{27}
\]

then, with high probability $1-\delta_{\mathrm{cal}}$,

\[
\boldsymbol{\pi}_t=\widetilde{\boldsymbol{\pi}}^\star
\tag{28}
\]

so the tentative route exactly matches the true all-hop optimal joint route.

\emph{Proof.}
Under equation (27), Theorem 1 gives

\[
\widetilde{J}^\star(\boldsymbol{\pi}_t)
<
\widetilde{J}^\star(\widetilde{\boldsymbol{\pi}}^\star)
+
\widetilde{\Delta}_{\mathrm{route}}
=
\min_{\boldsymbol{\pi}\neq\widetilde{\boldsymbol{\pi}}^\star}\widetilde{J}^\star(\boldsymbol{\pi}).
\]

The right-hand side is the best cost among all routes other than $\widetilde{\boldsymbol{\pi}}^\star$. Since $\widetilde{\boldsymbol{\pi}}^\star$ is the unique optimizer of $\widetilde{J}^\star$, the only possibility is $\boldsymbol{\pi}_t=\widetilde{\boldsymbol{\pi}}^\star$. $\square$

\subsubsection{Threshold Interpretation with Hop Counts}
\label{app:mtsp-margin-threshold}

For adjacencies on the true all-hop optimal joint route, define the maximum closure-witness hop count by

\[
H^\star_{\max}
:=
\max_{(i,j)\in\mathrm{Adj}(\widetilde{\boldsymbol{\pi}}^\star)} H^\star_{ij}
\le K+2M-1
\tag{29}
\]

The bound $H^\star_{\max}\le K+2M-1$ follows because a simple path on the anchor-state graph contains at most $K+2M$ nodes and therefore at most $K+2M-1$ hops; see equation (7).

If every direct edge on the true closure witness path of each $(i,j)\in\mathrm{Adj}(\widetilde{\boldsymbol{\pi}}^\star)$ satisfies the premises of Assumption 3 and the witness-threshold condition in equation (15), then equation (20) gives

\[
\widetilde{\bar{\varepsilon}}_t
\le
H^\star_{\max}\bigl(\varepsilon_{\mathrm{loc}}(\tau)+\alpha\tau^2\bigr).
\tag{30}
\]

Consequently, a sufficient exact-recovery condition is

\[
(K+M)H^\star_{\max}\bigl(\varepsilon_{\mathrm{loc}}(\tau)+\alpha\tau^2\bigr)
<
\widetilde{\Delta}_{\mathrm{route}}
\tag{31}
\]

Whenever equation (31) holds, Corollary 1 yields $\boldsymbol{\pi}_t=\widetilde{\boldsymbol{\pi}}^\star$.

\textbf{Remark (Conditional interpretation of the route-certification threshold).}
Equation (31) should be read as a conditional interpretation of threshold $\tau$ under the relaxed closure analysis, not as a theorem that the implemented scheduler is optimal or sufficient for the full algorithm. Through this lens, $\tau$ marks when an anchor pair has entered the local small-gap regime relevant to Corollary 1, not a separate on-route acceptance quota.

Under this interpretation, the 2-budget split is a design heuristic rather than a proved consequence. Prioritizing current-route pairs with larger witness gaps is aligned with reducing the route-comparison uncertainty in Theorem 1, whereas off-route pairs whose witness gaps are already below $\tau$ become lower-priority unless the tentative route changes.

Thus the role of $\tau$ in the 2-budget design is to provide a conservative screening threshold for the relaxed route-comparison analysis. It does not certify that off-route probing is globally unnecessary, and it does not by itself prove that the current-route proxy set $\mathrm{Adj}(\boldsymbol{\pi}_t)$ matches the unknown target set $\mathrm{Adj}(\widetilde{\boldsymbol{\pi}}^\star)$.

Accordingly, the inverted sufficient condition

\[
(K+M)H^\star_{\max}\bigl(\varepsilon_{\mathrm{loc}}(\tau)+\alpha\tau^2\bigr)
<
\widetilde{\Delta}_{\mathrm{route}}
\]

and the approximation

\[
\tau
\approx
\sqrt{
\frac{\widetilde{\Delta}_{\mathrm{route}}}
{(K+M)H^\star_{\max}\alpha}
}
\]

should be interpreted only as heuristic design rules under the relaxed model. When $M=1$, the same reading reduces to the single-agent form with $(K+M)=(K+1)$.

\subsection{Compatibility with Accepted-Edge Constraints and Optional Coverage Notes}
\label{app:mtsp-acceptance-coverage}

\subsubsection{Accepted-Edge Compatibility}

The theorem chain in Section 4 and 5 directly analyzes the relaxed solver (equation (6)).
Now consider a constrained solver that satisfies the commitment set $E_{\mathrm{acc}}^t$.
In this case, the additional event required to transfer the relaxed result is $\mathcal{E}_t^\star$ in equation (6b).

\paragraph{Local Degree Screen for Accepted-Edge Compatibility}
\label{app:mtsp-degree-constraints}
Budget allocation determines which ordered pairs are expanded next, whereas executability is tracked through a separate accepted adjacency set. We first define a local degree screen for that set through the admissible degree counts

\[
\deg_E^+(a)
:=
\sum_{b \in \mathcal{A}\setminus\{a\}}
\mathbf{1}\!\bigl[(a,b)\in E\bigr],
\qquad
\deg_E^-(a)
:=
\sum_{b \in \mathcal{A}\setminus\{a\}}
\mathbf{1}\!\bigl[(b,a)\in E\bigr],
\]

with constraints

\[
\deg_E^-(S_m)=0,\qquad \deg_E^+(S_m)\le 1,
\]

\[
\deg_E^+(G_m)=0,\qquad \deg_E^-(G_m)\le 1,
\]

\[
\deg_E^-(W_k)\le 1,\qquad \deg_E^+(W_k)\le 1.
\]

\textbf{Remark (Feasible-competitor transfer under accepted-edge compatibility).}
The degree inequalities above are only local admissibility conditions; by themselves they do not exclude subtours, disconnected components, or other globally incompatible partial commitments. The transfer statement below therefore relies on the explicit compatibility event $\mathcal{E}_t^\star$, not on the degree screen alone.

Suppose the constrained solver
\[
\boldsymbol{\pi}_t^{\mathrm{acc}}
\in
\arg\min_{\boldsymbol{\pi}\in\Pi_t^{\mathrm{acc}}}
\sum_{(i,j)\in\mathrm{Adj}(\boldsymbol{\pi})}\widetilde{C}_t(i,j)
\tag{31a}
\]
and the event $\mathcal{E}_t^\star$ holds. Then
$\widetilde{\boldsymbol{\pi}}^\star\in\Pi_t^{\mathrm{acc}}$, so the proof of
Theorem~1 applies verbatim after replacing $\boldsymbol{\pi}_t$ with
$\boldsymbol{\pi}_t^{\mathrm{acc}}$. Hence
\[
\widetilde{J}^\star(\boldsymbol{\pi}_t^{\mathrm{acc}})
\le
\widetilde{J}^\star(\widetilde{\boldsymbol{\pi}}^\star)
+
(K+M)\widetilde{\bar{\varepsilon}}_t
\tag{31b}
\]
the same comparison bound holds for the constrained solver. That is, if the
true all-hop optimal joint route remains feasible under the accepted-edge
constraint, the relaxed comparison theorem transfers directly.
This remark does not guarantee $\mathcal{E}_t^\star$ itself; it only states
how the relaxed theorem is reused once that compatibility event holds.

\textbf{Remark (Floyd-Warshall Closure Propagation as Indirect Coverage).}
When we concentrate exploration on pairs corresponding to the adjacency structure of the current tentative route,
pairs in $\mathrm{Adj}(\widetilde{\boldsymbol{\pi}}^\star) \setminus \mathrm{Adj}(\boldsymbol{\pi}_t)$ can also improve indirectly.

Specifically, for a pair $(i,j) \in \mathrm{Adj}(\widetilde{\boldsymbol{\pi}}^\star) \setminus \mathrm{Adj}(\boldsymbol{\pi}_t)$,
suppose there exists an intermediate sequence $i = \zeta_0, \zeta_1, \dots, \zeta_L = j$
such that every consecutive pair $(\zeta_\ell, \zeta_{\ell+1}) \in \mathrm{Adj}(\boldsymbol{\pi}_t)$.
Then reducing the edge values $C_t(\zeta_\ell, \zeta_{\ell+1})$
through the multi-hop minimum in the definition of closure (equation 5)

\[
\widetilde{C}_t(i,j)
\le
\sum_{\ell=0}^{L-1} C_t(\zeta_\ell,\zeta_{\ell+1})
\]

improves this upper bound and creates an incentive for the tentative joint route $\boldsymbol{\pi}_t$ to flip toward $\widetilde{\boldsymbol{\pi}}^\star$.

We call this \textbf{closure-reachability}:
if the pair $(i,j)$ is closure-reachable through a multi-hop path composed only of edges in $\mathrm{Adj}(\boldsymbol{\pi}_t)$,
then exploring $\mathrm{Adj}(\boldsymbol{\pi}_t)$ indirectly improves $\widetilde{C}_t(i,j)$ for pair $(i,j)$ when it is closure-reachable.
Conversely, a pair that is not closure-reachable does not receive this indirect benefit,
and this is precisely why the coverage augmentation below is needed.

\textbf{Caution in the $M>1$ setting.} Closure-reachability is defined while
ignoring agent boundaries. The intermediate anchor sequence
$\zeta_0,\dots,\zeta_L$ may therefore pass through anchors associated with
multiple agents. The resulting closure improvement is mathematically valid, but
an executable joint route still uses only directly realized intra-agent
segments. Closure-reachability should thus be interpreted as an indirect search
mechanism, while feasibility is handled separately by the accepted-edge
constraint in equation~(6a).

\subsubsection{Optional Coverage Augmentation}

The theorem chain above assumes that route-relevant direct edges are eventually
explored sufficiently often. The following optional augmentation makes this
coverage requirement explicit.
Let $U_t$ denote the unresolved ordered-pair set at round $t$, and allocate a
fraction $\rho\in(0,1)$ of the probing budget to uniform random exploration
over $U_t$. The remaining $1-\rho$ is then assigned to the original ranking
rule. For a random unresolved ordered pair $(i,j)\in U_t$,
\[
\Pr\!\bigl((i,j)\text{ is probed at round }t \mid (i,j)\in U_t\bigr)
\ge
\frac{\rho}{|U_t|}
\ge
\frac{\rho}{(K+2M)(K+2M-1)}
\tag{31c}
\]
this yields the pairwise coverage bound.
Equation~(31c) provides a weak but explicit pairwise coverage guarantee from a
small exploration floor. This augmentation is not required by the core theorem
chain, but it can be added when an explicit non-starvation property is desired.

\textbf{Remark (Direct-matrix screening heuristic for coverage-floor tightening).}
A conservative direct-matrix screen can be used to reduce the target set $U_t$ of the coverage floor.
Let $\bar{\varepsilon}_t := \max_{(i,j)\in U_t}\varepsilon_{ij}(t)$ be the worst-case direct-edge error over the current unresolved pairs. For a joint route $\boldsymbol{\pi}$, consider

\[
\sum_{(i,j)\in\mathrm{Adj}(\boldsymbol{\pi})} C_t(i,j)
>
\sum_{(i,j)\in\mathrm{Adj}(\boldsymbol{\pi}_t)} C_t(i,j)
+
2(K+M)\bar{\varepsilon}_t
\tag{31d}
\]

Equation (31d) compares route sums built from the direct matrix $C_t$, not from the closure objective $\widetilde{C}_t$ analyzed in Theorem~1 and Corollary~1. It therefore should not be read as a theorem-level certificate for the closure-based solver. Instead, it is a conservative screening heuristic: if a candidate route remains worse than the current tentative route even after allowing a direct-edge error budget $2(K+M)\bar{\varepsilon}_t$, that route is less plausible as a source of future route flips under the current direct matrix.

Accordingly, one may define $\widetilde{U}_t \subseteq U_t$ by removing pairs that appear only in routes screened out by equation (31d) and then apply the coverage floor on $\widetilde{U}_t$. This can improve the probability lower bound from equation (31c) to $\rho/|\widetilde{U}_t|$, but the reduction should be interpreted as a conservative heuristic rather than a theorem-certified elimination rule.

\subsubsection{Conceptual Interpretation via Combinatorial Pure Exploration}

The theorem chain in Section 4--5 provides a static route-comparison guarantee once sufficient exploration has occurred. The following discussion uses \textbf{Combinatorial Pure Exploration (CPE)} only as a conceptual lens for thinking about exploration pressure; it does not claim a regret bound, sample-complexity theorem, or stopping-time theorem for the implemented planner.

\begin{tabular}{|l|p{8cm}|}
\hline
CPE concept & corresponding concept in this problem \\
\hline
arm & anchor pair $(i,j)$ \\
arm pull & expansion 1 round toward the corresponding pair \\
arm value & $c^\star_{ij}$ \\
upper confidence bound & $C_t(i,j)\ge c^\star_{ij}$ (Proposition 2) \\
confidence width & $\varepsilon_{ij}(t) = C_t(i,j) - c^\star_{ij}$ \\
super-arm & joint Hamiltonian route $\boldsymbol{\pi}$ of $M$ agents \\
exploration goal & identification of $\widetilde{\boldsymbol{\pi}}^\star$ with the minimum number of explorations \\
\hline
\end{tabular}

This correspondence is only partial. The running minimum in equation (4) makes the stored direct costs monotone, the current tentative route $\boldsymbol{\pi}_t$ is only an indirect proxy for the unknown $\widetilde{\boldsymbol{\pi}}^\star$, and witness-gap quantities such as $\widehat{g}_t(w_t^{ij})$ are not the theorem object $\widetilde{\bar{\varepsilon}}_t$. We therefore use the CPE analogy only to explain why current-route exploitation together with a small off-route coverage floor is qualitatively reasonable, not to claim an additional theorem beyond the relaxed closure analysis.

\subsection{Empirical Support and Measurement Notes}

The auxiliary calibration analysis evaluates the quadratic correction
\[
\widehat{C}_{\alpha}(\widehat{g})
:=
 s_{\mathrm{local}}\bigl(\widehat{g}+\alpha \widehat{g}^{2}\bigr)
\]
on unordered feasible-point pairs drawn from the feasible-point file \texttt{antmaze\_giant\_feasible\_points.yaml} under \texttt{configurations/task\_overrides/}. Concretely, the calibration set contains 86 feasible points and 3{,}655 unordered pairs, where $T$ denotes the raw temporal-distance estimate for a pair and $G$ denotes the corresponding reference direct-pair graph cost. We first fit a shared local scale $s_{\mathrm{local}}=0.15131$ on the local window $T\le 50$ by minimizing the RMS log-error in $\log G-\log s-\log T$, and then keep this scale fixed while sweeping the quadratic coefficient $\alpha$ over all pairs. Under this fixed-$s_{\mathrm{local}}$ protocol, the best-fit coefficient is $\alpha^\star \approx 0.00164$ according to global log-RMSE, while the conservative envelope coefficient $\alpha_{95}\approx 0.00609$ is the smallest value whose corrected prediction overestimates the reference direct-pair cost on about $95\%$ of calibration pairs. In the implementation used for the reported experiments, we set $\alpha=0.005$, which lies between these two values and empirically overestimates the reference direct-pair cost on about $93\%$ of calibration pairs. Accordingly, the experimental setting should be interpreted as a near-conservative practical approximation to the upper-envelope condition $h(\widehat{g})\ge g^\star$, while $\alpha_{95}$ serves as a stricter calibration-side reference.

\begin{figure}[t]
  \centering
  \begin{minipage}{0.48\linewidth}
    \centering
    \includegraphics[width=\linewidth]{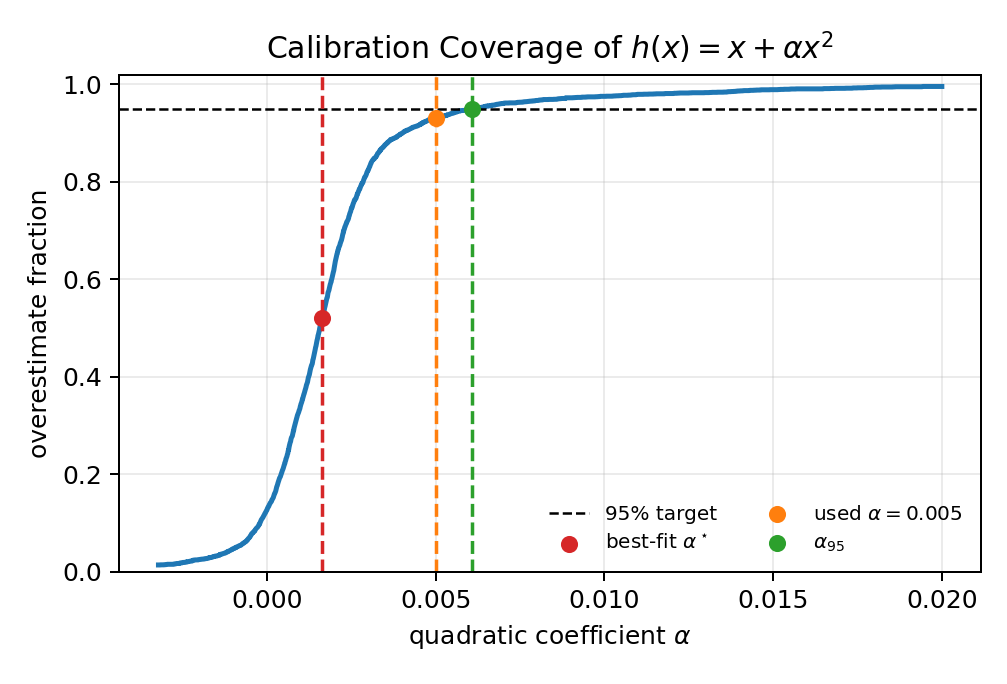}
  \end{minipage}\hfill
  \begin{minipage}{0.48\linewidth}
    \centering
    \includegraphics[width=\linewidth]{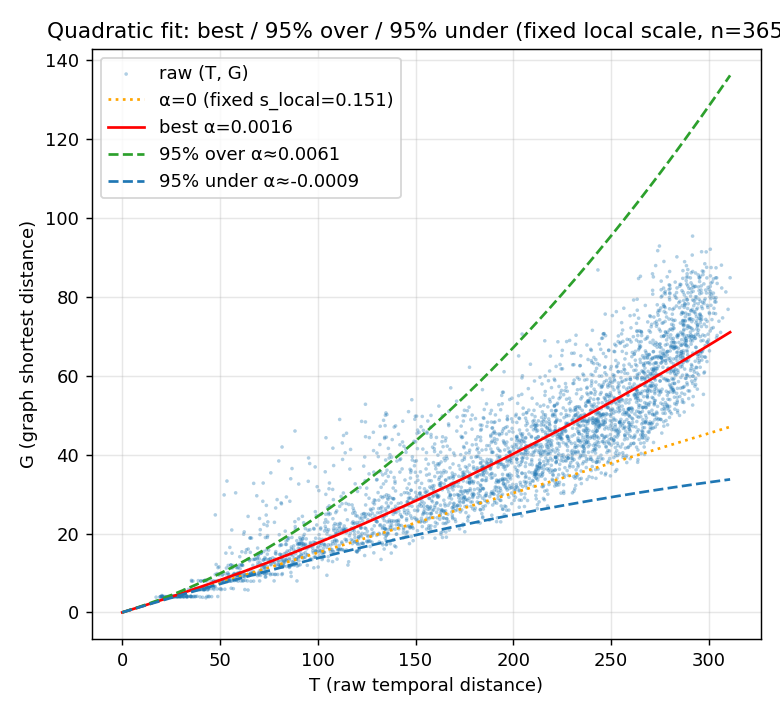}
  \end{minipage}
  \caption{Quadratic temporal-distance calibration used in Appendix~\ref{app:mtsp-theory}. Calibration pairs come from the same feasible-point file under \texttt{configurations/task\_overrides/} (86 points, 3{,}655 unordered pairs), where $T$ is the raw temporal prediction and $G$ is the reference direct-pair graph cost. A shared local scale $s_{\mathrm{local}}=0.15131$ is first fit on the local window $T\le 50$ by minimizing RMS log-error and then held fixed during the global $\alpha$ sweep. Left: empirical overestimate fraction of $h(x)=x+\alpha x^2$ as $\alpha$ varies, highlighting the best-fit coefficient $\alpha^\star\approx 0.00164$, the implemented coefficient $\alpha=0.005$, and the conservative envelope coefficient $\alpha_{95}\approx 0.00609$. Right: corrected temporal-vs.-graph scatter under the same fixed-$s_{\mathrm{local}}$ protocol, showing the quadratic envelopes used to motivate the upper-bound interpretation.}
  \label{fig:appendix-alpha-calibration}
\end{figure}

\begin{itemize}
  \item Figure~\ref{fig:appendix-alpha-calibration} reports both the $\alpha$-coverage sweep and the corrected temporal-vs.-graph scatter under the shared local-fit protocol.
  \item The left panel makes the selection rules explicit: $\alpha^\star$ minimizes global log-RMSE after fixing $s_{\mathrm{local}}$, whereas $\alpha_{95}$ is selected by the $95\%$ overestimate criterion.
  \item The same calibration pairs also provide low-gap absolute-error envelopes for estimating $\widehat{\varepsilon}_{\mathrm{loc}}(\tau)$.
\end{itemize}

Thus, the coefficient used in the reported experiments should be interpreted as a practically conservative quadratic correction, not as either a pure best-fit regressor or an exact empirical realization of the $95\%$ envelope.

This observation does not verify Assumption 2 uniformly over all rounds, candidates, or task distributions. It only provides qualitative support that, on the calibration distribution and especially in the low-gap regime used by the threshold analysis, the corrected predictor with $\alpha=0.005$ is often close to a conservative envelope, while $\alpha_{95}$ serves as a stricter calibration-side reference.

There are four quantities that can be measured for additional empirical support:

\begin{enumerate}
  \item \textbf{Upper-calibration coverage curve.}
If $\Pr[h(\widehat{g}_t(n,m))\ge g_t^{\star,ij}(n,m)]$ is directly measured as a function of probe depth or residual scale, empirical support of Assumption 2 is strengthened.
  \item \textbf{Calibration slack quantiles.}
If one measures absolute residual-error quantiles for each low-gap bin, one can connect $\widehat{\varepsilon}_{\mathrm{loc}}(\tau)$ to a concrete threshold-design rule.
  \item \textbf{Route margin statistics.}
By measuring the empirical distribution of $\widetilde{\Delta}_{\mathrm{route}}$, it is possible to determine how often the regime in equation (31) holds in the actual task family.
  \item \textbf{Accepted-edge precision.}
Measuring the false-positive rate in accepted adjacency can indirectly support the plausibility of the event $\mathcal{E}_t^\star$.
\end{enumerate}

\subsection{Scope of the Route-Level Guarantee}
\label{app:mtsp-scope}

The core theorem chain in this section is directly aligned with the Floyd-Warshall closure of the direct bridge matrix
and with the relaxed all-hop route objective.
In other words, the route analyzed here is always computed on the closure of the direct bridge matrix.

The theorem chain in this section is paired with a separate route-execution mechanism, namely meeting acceptance, the revocable accepted-set update $E_{\mathrm{soft}}^t$, and the completion event $\mathcal{C}_t$. Accordingly, the purpose of the present section is to justify the route-level allocation rule and the associated completion objective through these route-level objects.

Therefore, the route-level theory supports the following narrow interpretation.

\begin{itemize}
  \item In the setting where $M\ge 1$ agents partition-cover $K$ waypoints, the core theorem chain analyzes a relaxed mTSP-style Hamiltonian routing problem on the Floyd--Warshall closure of the direct bridge matrix.
  \item Under the conditioned upper-envelope event on direct-edge residual cost, the all-hop joint-route error is upper-bounded by $(K+M)\widetilde{\bar{\varepsilon}}_t$.
  \item Under the additional local small-gap and hop-count conditions, equation (31) gives a sufficient exact-recovery condition for the relaxed solver.
  \item These results motivate, but do not prove, prioritizing unresolved current-route adjacencies and using a thresholded off-route coverage floor.
  \item Meeting acceptance and $E_{\mathrm{soft}}^t$ are separate execution-side mechanisms and are not covered by the core theorem chain.
\end{itemize}

\subsection{Optional Planning-Augmented Route Loss Discussion}

The remainder of this section is an optional interpretive extension, not part of the core theorem chain. It introduces an additional loss that charges planning time in order to explain why exhaustive exploration need not be preferred once computation is part of the objective. The conclusions below therefore depend on this extra modeling choice and should not be read as consequences of Theorem~1 or Corollary~1 alone.

This motivates a stopping-time formulation: the planner must decide when to stop probing and execute the current tentative route, so the analysis should be expressed through a loss that balances completion time against final route quality.

\subsubsection{Planning-Augmented Stop Loss}

Let the set of direct sub-edges materialized up to round $t$ be denoted by $E_{\mathrm{soft}}^t$. This is the appendix notation corresponding to the revocable route-consistent accepted set defined in Section~\ref{sec:route-consistent-soft-acceptance}. At round $t$, let the event that the tentative joint route $\boldsymbol{\pi}_t$ is already executable be

\[
\mathcal{C}_t
:=
\Bigl\{
\forall (i,j)\in \mathrm{Adj}(\boldsymbol{\pi}_t),\;
\exists
\bigl(i=\zeta_0,\zeta_1,\dots,\zeta_H=j\bigr)
\text{ such that }
\sum_{\ell=0}^{H-1} C_t(\zeta_\ell,\zeta_{\ell+1})
=
\widetilde{C}_t(i,j),
\;
(\zeta_\ell,\zeta_{\ell+1})\in E_{\mathrm{soft}}^t
\;\forall \ell
\Bigr\}
\tag{32}
\]

That is, $\mathcal{C}_t$ means that every adjacency in the tentative joint route already has a materialized witness chain that attains the corresponding closure cost. This definition does not fix a specific witness path in advance; it only requires the existence of such a chain.

Define the route-completion time by

\[
\tau_{\mathrm{comp}}
:=
\inf\{t\ge 0 \mid \mathcal{C}_t \text{ holds}\}
\tag{33}
\]

We then define the planning-augmented loss by

\[
\mathcal{L}
:=
\lambda_{\mathrm{time}}\,\tau_{\mathrm{comp}}
+
\widetilde{J}^\star(\boldsymbol{\pi}_{\tau_{\mathrm{comp}}})
\tag{34}
\]

The first term penalizes the time at which the tentative joint route becomes fully executable, while the second term measures the true all-hop length of the joint route executed at that time.

Equivalently, since $\widetilde{J}^\star(\widetilde{\boldsymbol{\pi}}^\star)$ is constant for a fixed instance,

\[
\mathcal{L}
=
\lambda_{\mathrm{time}}\,\tau_{\mathrm{comp}}
+
\bigl(
\widetilde{J}^\star(\boldsymbol{\pi}_{\tau_{\mathrm{comp}}})
-
\widetilde{J}^\star(\widetilde{\boldsymbol{\pi}}^\star)
\bigr)
+
\widetilde{J}^\star(\widetilde{\boldsymbol{\pi}}^\star)
\tag{35}
\]

Thus, up to the additive constant $\widetilde{J}^\star(\widetilde{\boldsymbol{\pi}}^\star)$, minimizing completion time plus final route length is equivalent to minimizing completion time plus route-quality error.

Suppose that, at each round $t$, the pair budget $B_{\mathrm{pair}}$ is split as

\[
\bigl(
Q_t^{\mathrm{on}},
Q_t^{\mathrm{cov}}
\bigr)
\sim
\mathrm{Multinomial}
\!\left(
B_{\mathrm{pair}};
q_{\mathrm{on}},
q_{\mathrm{cov}}
\right),
\qquad
q_{\mathrm{on}}+q_{\mathrm{cov}}=1
\tag{36}
\]

The on-route bucket is assigned only to unresolved pairs on the tentative route and is spent using a single descending witness-gap priority over those route-adjacent pairs.

\begin{itemize}
  \item current-route unresolved pairs are ranked by witness gap in descending order inside the on-route bucket,
  \item off-route unresolved pairs receive only the fixed coverage floor that preserves route-flip possibility and non-starvation.
\end{itemize}

Thus the 2-budget allocation rule does not create separate on-route certification and materialization quotas. The role of $\tau$ is instead to determine which off-route pairs remain in the coverage pool as route-certification candidates.

\subsubsection{Mechanism-to-Loss Mapping}

In the 2-term loss above, each component plays the following role.

\begin{itemize}
\item \textbf{Components that primarily reduce the second term $\widetilde{J}^\star(\boldsymbol{\pi}_{\tau_{\mathrm{comp}}})$}
  \begin{itemize}
    \item the direct bridge matrix mechanism, namely the all-hop route solver
    \item the mechanism that preferentially exploits current-route pairs with larger witness gaps, namely the on-route descending-gap scheduling rule
    \item the mechanism justified by the temporal overestimate envelope and Theorem 1 / Corollary 1, namely the witness-gap reduction criterion
  \end{itemize}
\end{itemize}

\begin{itemize}
\item \textbf{Components that primarily reduce the first term $\tau_{\mathrm{comp}}$}
  \begin{itemize}
    \item meeting acceptance and the update of $E_{\mathrm{soft}}$
    \item the mechanism that reuses the current tentative route to maintain a route-consistent accepted set of executable adjacencies
    \item the mechanism that stops at the first time the completion event $\mathcal{C}_t$ is established, namely the stopping rule
  \end{itemize}
\end{itemize}

\begin{itemize}
\item \textbf{Component that acts indirectly on both terms}
  \begin{itemize}
\item off-route coverage floor: Ensures future route flip possibility and non-starvation rather than immediate loss reduction.
\item symmetric-direction deduplication: Allows more route-relevant probes with the same budget.
  \end{itemize}
\end{itemize}

In other words, the explore/exploit tradeoff in this problem is not between separate certification and materialization quotas.
It is formalized as the tradeoff between spending additional probes on current-route adjacencies to reduce route uncertainty and reserving a small coverage floor on off-route pairs to preserve future route correction and non-starvation.

\subsubsection{Observable Width Proxy}

The second term in equation (34) is latent during planning, so it cannot be evaluated directly online. We therefore define the width proxy on the corrected predictor rather than on the raw temporal prediction:

\[
\widehat{C}_{\alpha}(\widehat{g})
:=
s_{\mathrm{local}}\bigl(\widehat{g}+\alpha \widehat{g}^{2}\bigr).
\]

Using the same calibration pairs, define the corrected multiplicative residual ratio by

\[
\rho_{\alpha}
:=
\frac{g^\star}{\widehat{C}_{\alpha}(\widehat{g})}.
\tag{37}
\]

Here, the experimental setting uses $\alpha=0.005$, while the stricter calibration-side reference is $\alpha_{95}\approx 0.00609$. In the local-fit window, let $\kappa_{q}^{\alpha,\mathrm{loc}}:=Q_q(\rho_{\alpha})$ denote the empirical $q$-quantile of the corrected ratio. For the implementation value $\alpha=0.005$, the slot form is

\[
\begin{aligned}
\kappa_{0.05}^{0.005,\mathrm{loc}} &\approx \text{[insert q05 ratio]},\\
\kappa_{0.50}^{0.005,\mathrm{loc}} &\approx \text{[insert median ratio]},\\
\kappa_{0.95}^{0.005,\mathrm{loc}} &\approx \text{[insert q95 ratio]}.
\end{aligned}
\tag{38}
\]

Accordingly, we define the local uncertainty band by

\[
W_{0.005,\mathrm{loc}}(\widehat{g})
:=
\bigl(
\kappa_{0.95}^{0.005,\mathrm{loc}}
-
\kappa_{0.05}^{0.005,\mathrm{loc}}
\bigr)
\widehat{C}_{0.005}(\widehat{g})
\approx
\text{[insert local width coeff]}\,\widehat{C}_{0.005}(\widehat{g}).
\tag{39}
\]

An analogous all-pair width can be reported as
\[
W_{0.005,\mathrm{all}}(\widehat{g})
\approx
\text{[insert all-pair width coeff]}\,\widehat{C}_{0.005}(\widehat{g}),
\]
using the all-pair quantiles of $\rho_{0.005}$ instead of the local-fit quantiles. When a corrected-ratio quantile plot is attached, these slots can be filled directly from the $5\%/50\%/95\%$ binned quantiles of $\rho_{0.005}$ against either $\widehat{g}$ or $\widehat{C}_{0.005}(\widehat{g})$.

This motivates the surrogate planning loss

\[
\widehat{\mathcal{L}}_t
:=
\lambda_{\mathrm{time}}\,t
+
\lambda_{\mathrm{err}}
\sum_{(i,j)\in\mathrm{Adj}(\boldsymbol{\pi}_t)}
W_{0.005}\!\bigl(\widehat{g}_t(w_t^{ij})\bigr)
\tag{40}
\]

This quantity is not a theorem-level objective. Rather, it combines the route-error bound with an empirical width law on the corrected predictor to provide an interpretable stopping indicator during planning.

\subsubsection{Comparison with an Exhaustive-Search Baseline}

Now consider an idealized exhaustive baseline $\mathsf{Full}$. It explores every ordered pair sufficiently before execution, stops at time $\tau_{\mathrm{full}}$, and produces the final joint route $\boldsymbol{\pi}_{\tau_{\mathrm{full}}}^{\mathrm{full}}$.

Once equation (34) is adopted, exhaustive search is no longer automatically optimal. Although it can reduce the route-quality term, it also incurs the potentially large completion-time cost $\lambda_{\mathrm{time}}\tau_{\mathrm{full}}$. The relevant comparison is therefore over the full planning-augmented loss.

By Theorem 1, for any stopping time $\tau$,

\[
\widetilde{J}^\star(\boldsymbol{\pi}_\tau)-\widetilde{J}^\star(\widetilde{\boldsymbol{\pi}}^\star)
\le
(K+M)\widetilde{\bar{\varepsilon}}_\tau
\tag{41}
\]

If the exhaustive baseline is idealized as recovering the exact all-hop joint route, then the loss difference when this method stops at completion time $\tau_{\mathrm{comp}}$ satisfies

\[
\mathcal{L}-\mathcal{L}_{\mathrm{full}}
\le
\lambda_{\mathrm{time}}
\bigl(
\tau_{\mathrm{comp}}-\tau_{\mathrm{full}}
\bigr)
+
(K+M)\widetilde{\bar{\varepsilon}}_{\tau_{\mathrm{comp}}}
\tag{42}
\]

Therefore, a sufficient condition for this method to outperform exhaustive search is

\[
\lambda_{\mathrm{time}}
\bigl(
\tau_{\mathrm{full}}-\tau_{\mathrm{comp}}
\bigr)
>
(K+M)\widetilde{\bar{\varepsilon}}_{\tau_{\mathrm{comp}}}
\tag{43}
\]

Under equation (43),

\[
\mathcal{L}
<
\mathcal{L}_{\mathrm{full}}
\tag{44}
\]

Equation (43) identifies the regime in which this method can outperform exhaustive search under the planning-augmented loss: if the route-quality benefit of exhaustive search is smaller than the additional completion delay it incurs, then this method attains the smaller loss.

In particular, suppose there exists a stopping time $\tau_{\mathrm{stop}}$ such that

\begin{enumerate}
  \item $\mathcal{C}_{\tau_{\mathrm{stop}}}$ holds, so the tentative joint route is executable,
  \item the condition of Corollary 1 gives $\boldsymbol{\pi}_{\tau_{\mathrm{stop}}}=\widetilde{\boldsymbol{\pi}}^\star$,
  \item exhaustive search is still exploring off-route pairs, so $\tau_{\mathrm{stop}}<\tau_{\mathrm{full}}$.
\end{enumerate}

then the loss definition directly gives

\[
\mathcal{L}
-
\mathcal{L}_{\mathrm{full}}
\le
\lambda_{\mathrm{time}}
\bigl(
\tau_{\mathrm{stop}}-\tau_{\mathrm{full}}
\bigr)
<
0
\tag{45}
\]

which means that this method has the smaller loss. Under this optional planning-augmented loss, this gives an idealized motivation for emphasizing route-relevant current-route adjacencies while retaining a small off-route coverage floor.

\subsubsection{Resulting Design Principle}

The appendix therefore suggests the following heuristic allocation design under the relaxed route analysis and the optional planning-augmented loss.
Use the closure-based route solver to update the tentative route, allocate the main exploitation budget to unresolved current-route adjacencies in descending witness-gap order, and reserve a small coverage floor for off-route pairs whose witness gaps still exceed the route-certification threshold $\tau$.

\paragraph{Selected Pair Interface}
The two budgets satisfy

\[
Q_t^{\mathrm{on}} + Q_t^{\mathrm{cov}} = B_{\mathrm{pair}},
\qquad
Q_t^{\mathrm{on}},Q_t^{\mathrm{cov}}\in\{0,\dots,B_{\mathrm{pair}}\},
\]

and the selected candidate sets obey

\[
\mathcal{E}_t^{\mathrm{on}} \subseteq \mathcal{O}_t,
\qquad
\mathcal{E}_t^{\mathrm{cov}} \subseteq \mathcal{U}_t,
\qquad
|\mathcal{E}_t^{\mathrm{on}}|\le Q_t^{\mathrm{on}},
\qquad
|\mathcal{E}_t^{\mathrm{cov}}|\le Q_t^{\mathrm{cov}}.
\]

Each selected anchor-pair candidate is converted into a same-round bidirectional execution bundle,

\[
\mathcal{P}_t^{\mathrm{on}}
:=
\bigcup_{(i,j)\in\mathcal{E}_t^{\mathrm{on}}}
\{(i,j),(j,i)\},
\qquad
\mathcal{P}_t^{\mathrm{cov}}
:=
\bigcup_{(i,j)\in\mathcal{E}_t^{\mathrm{cov}}}
\{(i,j),(j,i)\},
\]

and the final ordered-pair batch passed to Section~\ref{sec:node-evaluation} is

\[
\mathcal{P}_t
:=
\mathcal{P}_t^{\mathrm{on}}
\cup
\mathcal{P}_t^{\mathrm{cov}}.
\]

Every ordered pair in $\mathcal{P}_t$ is then processed by the Section~\ref{sec:anchor-chaining-tree-diffusion-planner} pipeline: pair-conditioned target selection, source-node evaluation, child expansion, and meeting detection are executed for each direction in the same round.

\begin{enumerate}
  \item re-solve the tentative route on the Floyd-Warshall closure of the direct bridge matrix,
  \item exploit current-route adjacencies first by descending witness gap,
  \item use meeting acceptance and $E_{\mathrm{soft}}$ to convert route-level evidence into the completion event $\mathcal{C}_t$ while maintaining a small off-route coverage floor.
\end{enumerate}

\section{Full ChronoForest Algorithm}
\label{app:full-algorithm}

% Full ChronoForest algorithm for appendix.
% Expands Algorithm 1's four abstractions (ORCHESTRATE / EXPAND / UPDATE / COMPLETE)
% using the notation established in the main text and Appendix~\ref{app:mtsp-theory}.
%
% Notation carried over from the paper:
%   A          -- anchor set
%   f_theta    -- diffusion planner (= f_plan in Alg. 1)
%   d_TD       -- temporal-distance estimator
%   T_i        -- tree rooted at anchor a_i
%   g(n)       -- cost-to-come at node n
%   x(n)       -- state at node n
%   tgt(n)     -- assigned target node for n
%   V_j        -- representative nodes of T_j used as target candidates
%   J_{i->j}   -- pair-conditioned estimated total cost (Section 3.2)
%   h(x)       -- corrected overestimate h(x)=x+alpha x^2 (Appendix B)
%   C_t, C_tilde_t -- direct bridge matrix and its Floyd-Warshall closure
%   B_t, gamma_t   -- best witness record and its predicted gap
%   E_soft^t   -- revocable route-consistent accepted adjacency set
%   chi_t(i,j) -- indicator: round t produced a meeting witness for (i,j)

\begin{algorithm}[t]
\caption{ChronoForest (Full)}
\label{alg:chronoforest-full}
\begin{algorithmic}[1]
\Require anchor set $\mathcal{A}$,
         diffusion planner $f_\theta$,
         temporal-distance estimator $d_{\mathrm{TD}}$,
         top-$K$ expansion width,
         pair budget $B_{\mathrm{pair}}$ split as $Q^{\mathrm{on}}+Q^{\mathrm{cov}}=B_{\mathrm{pair}}$,
         $\lambda_{\mathrm{unc}}$,\;
         certification threshold $\tau$,\;
         meeting threshold $\delta_{\mathrm{meet}}$

\smallskip
\State \textit{// Initialization}
\State For each $a\in\mathcal{A}$: initialize tree $\mathcal{T}_a$ with a single root at $a$, \; $g(\mathrm{root}_a)\gets 0$
\State $C_0(i,j)\gets$ root-to-root bridge score for all $(i,j)$;\quad
       $B_0\gets\varnothing$;\quad
       $\gamma_0(i,j)\gets+\infty$;\quad
       $E_{\mathrm{soft}}^0\gets\varnothing$
\State $\hat{\Pi}_0\gets$ any feasible $\boldsymbol{\pi}\in\Pi_M(\mathcal{A};\{S_m\},\{G_m\})$

\smallskip
\For{$t=0,1,2,\dots$}

  \smallskip
  \State \textit{// --- ORCHESTRATE: Online Multi-Tree Orchestrator ---}

  \State \textit{// Floyd-Warshall all-hop closure}
  \State $\displaystyle\widetilde{C}_t(i,j)\gets
         \min_{\substack{L\ge 1,\;\zeta_0=i,\;\zeta_L=j\\ \zeta_1,\dots,\zeta_{L-1}\in\mathcal{A}}}
         \sum_{\ell=0}^{L-1}C_t(\zeta_\ell,\zeta_{\ell+1})$

  \smallskip
  \State \textit{// Tentative route re-solving}
  \State $\displaystyle\hat{\Pi}_t\in\arg\min_{\boldsymbol{\pi}\in\Pi_M(\mathcal{A};\{S_m\},\{G_m\})}
         \sum_{(i,j)\in\mathrm{Adj}(\boldsymbol{\pi})}\widetilde{C}_t(i,j)$

  \smallskip
  \State \textit{// Route-prioritized budget allocation}
  \State $\mathcal{O}_t\gets\{(i,j)\in\mathrm{Adj}(\hat{\Pi}_t)\mid(i,j)\notin E_{\mathrm{soft}}^{t-1}\}$
         \hfill\textit{// on-route uncertified pairs}
  \State $\mathcal{U}_t\gets\{(i,j)\notin\mathrm{Adj}(\hat{\Pi}_t)\mid
         \gamma_t(i,j)>\tau\;\text{or}\;B_t(i,j)=\varnothing\}$
         \hfill\textit{// off-route coverage floor}
  \State $\mathcal{E}_t^{\mathrm{on}}\gets\operatorname{Top}_{Q^{\mathrm{on}}}
         \!\bigl(\operatorname{SortDesc}_{\gamma_t}(\mathcal{O}_t)\bigr)$;\quad
         $\mathcal{E}_t^{\mathrm{cov}}\subseteq\mathcal{U}_t$,\;
         $|\mathcal{E}_t^{\mathrm{cov}}|\le Q^{\mathrm{cov}}$ \hfill\textit{// cyclic round-robin}
  \State $\mathcal{P}_t\gets\{(i,j),(j,i)\mid(i,j)\in\mathcal{E}_t^{\mathrm{on}}\cup\mathcal{E}_t^{\mathrm{cov}}\}$
         \hfill\textit{// both-direction ordered-pair batch}

  \smallskip
  \State \textit{// --- EXPAND: Anchor-chaining Tree Diffusion Planner ---}

  \For{each ordered pair $(i,j)\in\mathcal{P}_t$}

    \State \textit{// Pair-conditioned target selection}
    \For{each source-side representative $\rho\in\mathcal{T}_i$}
      \State $\displaystyle m_\rho^\star\in\arg\min_{m\in\mathcal{V}_j}\mathcal{J}_{i\to j}(\rho,m)$,\quad
             $\mathcal{J}_{i\to j}(\rho,m):=g(\rho)+h\!\bigl(d_{\mathrm{TD}}(x(\rho),x(m))\bigr)+g(m)$
    \EndFor

    \smallskip
    \State \textit{// Source node evaluation and selection}
    \State $\mathrm{score}(n)\gets
           -\bigl(\operatorname{rank}(\mathrm{cost}(n))+\lambda_{\mathrm{unc}}\operatorname{rank}(U(n))\bigr)$
           \;for each candidate source $n\in\mathcal{T}_i$
    \State Select top-$K$ source nodes by $\mathrm{score}$;\;
           promote promising lookahead descendants into active frontier

    \smallskip
    \State \textit{// Target-conditioned guided diffusion sampling (Junction Detector)}
    \For{each selected leaf $n$ (with assigned target $\mathrm{tgt}(n)$, target state $g_{\mathrm{tgt}}:=x(\mathrm{tgt}(n))$)}
      \State Denoise multiple lookahead subplans $\hat{X}_0=f_\theta(X_t,t)$ in parallel
      \State $\displaystyle\tilde{X}_0\gets\hat{X}_0
             +\eta_t\,\nabla_{\!\hat{X}_0}
             \mathcal{U}(\hat{X}_0;\,g_{\mathrm{tgt}})$,\quad
             $\mathcal{U}(\hat{X}_0;\,g_{\mathrm{tgt}}):=-d_{\mathrm{TD}}\!\bigl(\hat{x}^{(L)},g_{\mathrm{tgt}}\bigr)$
    \EndFor

    \smallskip
    \State \textit{// Subplan pruning and representative clustering}
    \State Discard infeasible or dominated subplans;\;
           cluster survivors by temporal distance;\;
           retain one representative per cluster
    \algstore{chronoforestfull}
\end{algorithmic}
\end{algorithm}

\begin{algorithm}[t]
\caption[]{ChronoForest (Full, continued)}
\begin{algorithmic}[1]
\algrestore{chronoforestfull}

    \smallskip
    \State \textit{// Meeting detection}
    \For{each newly exposed node pair $(\rho,m)$ with $\rho\in\mathcal{T}_i$,\;$m\in\mathcal{T}_j$}
      \If{$d_{\mathrm{TD}}(x(\rho),x(m))<\delta_{\mathrm{meet}}$}
        \State Record $(\rho,m)$ as meeting witness;\; $\chi_t(i,j)\gets 1$
      \EndIf
    \EndFor

  \EndFor

  \smallskip
  \State \textit{// --- UPDATE ---}

  \For{each ordered pair $(i,j)$ with new evidence $\mathcal{N}_{ij}^{\mathrm{new}}(t)$}
    \State $\displaystyle C_t(i,j)\gets
           \min\!\Bigl(C_{t-1}(i,j),\;
           \min_{(\rho,m)\in\mathcal{N}_{ij}^{\mathrm{new}}(t)}\mathcal{J}_{i\to j}(\rho,m)\Bigr)$
    \State Update $B_t(i,j)$ and $\gamma_t(i,j)$ if lower-cost witness found
  \EndFor

  \smallskip
  \State \textit{// Route-consistent soft acceptance}
  \State $A_t^{\mathrm{meet}}\gets
         \bigl\{(i,j)\in\mathrm{Adj}(\hat{\Pi}_t)\mid
         \chi_t(i,j)=1,\;
         E_{\mathrm{soft}}^{t-1}\cup\{(i,j)\}\text{ satisfies degree constraints}\bigr\}$
  \State $E_{\mathrm{soft}}^t\gets
         \bigl(E_{\mathrm{soft}}^{t-1}\cap\mathrm{Adj}(\hat{\Pi}_t)\bigr)
         \cup A_t^{\mathrm{meet}}$

  \smallskip
  \State \textit{// --- COMPLETE ---}
  \If{$\mathrm{Adj}(\hat{\Pi}_t)\subseteq E_{\mathrm{soft}}^t$}
    \State \textbf{break}
  \EndIf

\EndFor
\State \Return assembled plan from accepted bridges in $E_{\mathrm{soft}}^T$
\end{algorithmic}
\end{algorithm}

\clearpage

\section{Supporting Experiment Diagnostics}
\label{app:experiments-support}
\subsection{Official-Stitch Route-Length Validation}
\label{app:stitch-bfs-validation}

\begin{table}[H]
  \centering
  \small
  \setlength{\tabcolsep}{5pt}
  \begin{tabular}{ll ccccc}
    \toprule
    & & Task 1 & Task 2 & Task 3 & Task 4 & Task 5 \\
    \midrule
    \multirow{3}{*}{Medium}
      & BFS shortest path (wu)                  &    40    &    40    &    24    &    40    &    32    \\
      & Plan mean (wu)                          &   37.0   &   35.5   &   21.6   &   35.5   &   29.1   \\
      & \% within $1.1\!\times\!$BFS $\uparrow$ &  100.0   &  100.0   &  100.0   &  100.0   &  100.0   \\
    \midrule
    \multirow{3}{*}{Large}
      & BFS shortest path (wu)                  &    60    &    76    &    48    &    56    &    60    \\
      & Plan mean (wu)                          &   56.9   &   71.7   &   48.1   &   53.4   &   57.5   \\
      & \% within $1.1\!\times\!$BFS $\uparrow$ &  100.0   &   97.5   &  100.0   &  100.0   &  100.0   \\
    \midrule
    \multirow{3}{*}{Giant}
      & BFS shortest path (wu)                  &   120    &   104    &   120    &   104    &    68    \\
      & Plan mean (wu)                          &  117.5   &   99.8   &  109.5   &   96.8   &   62.6   \\
      & \% within $1.1\!\times\!$BFS $\uparrow$ &  100.0   &  100.0   &  100.0   &  100.0   &  100.0   \\
    \bottomrule
  \end{tabular}
  \caption{Per-task plan length relative to the BFS grid shortest path on
  AntMaze-Stitch. The BFS reference is the 4-connected maze-grid shortest path
  with \texttt{maze\_unit}$=4$\,wu/cell. Plan mean is computed from the same 80
  rollouts per task used in the stitch benchmark (4 repeats $\times$ 20
  rollouts).}
  \label{tab:plan_length_bfs}
\end{table}

\begin{figure}[H]
  \centering
  \setlength{\tabcolsep}{1pt}
  \begin{tabular}{@{}ccccc@{}}
    \includegraphics[width=0.19\textwidth]{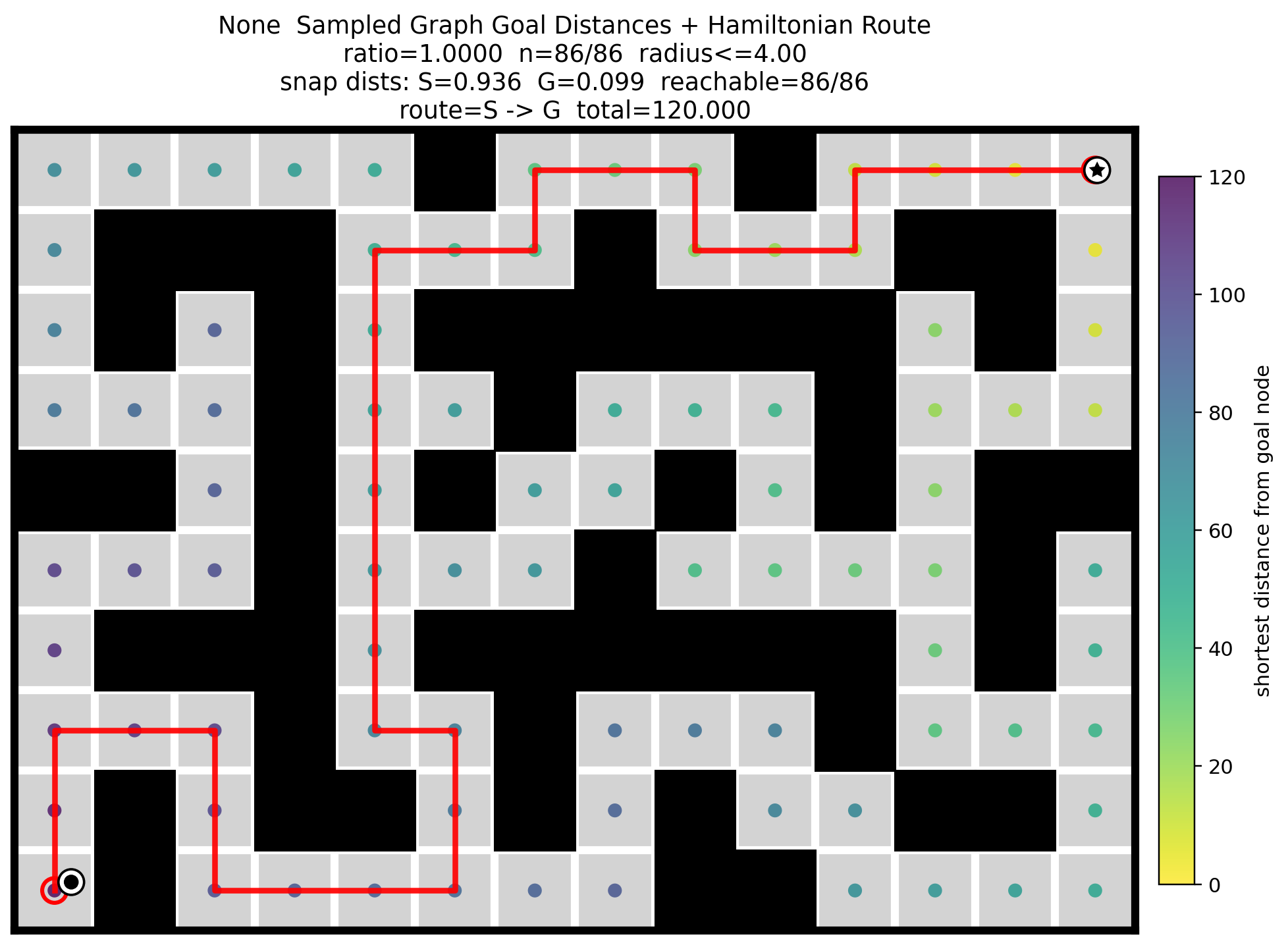} &
    \includegraphics[width=0.19\textwidth]{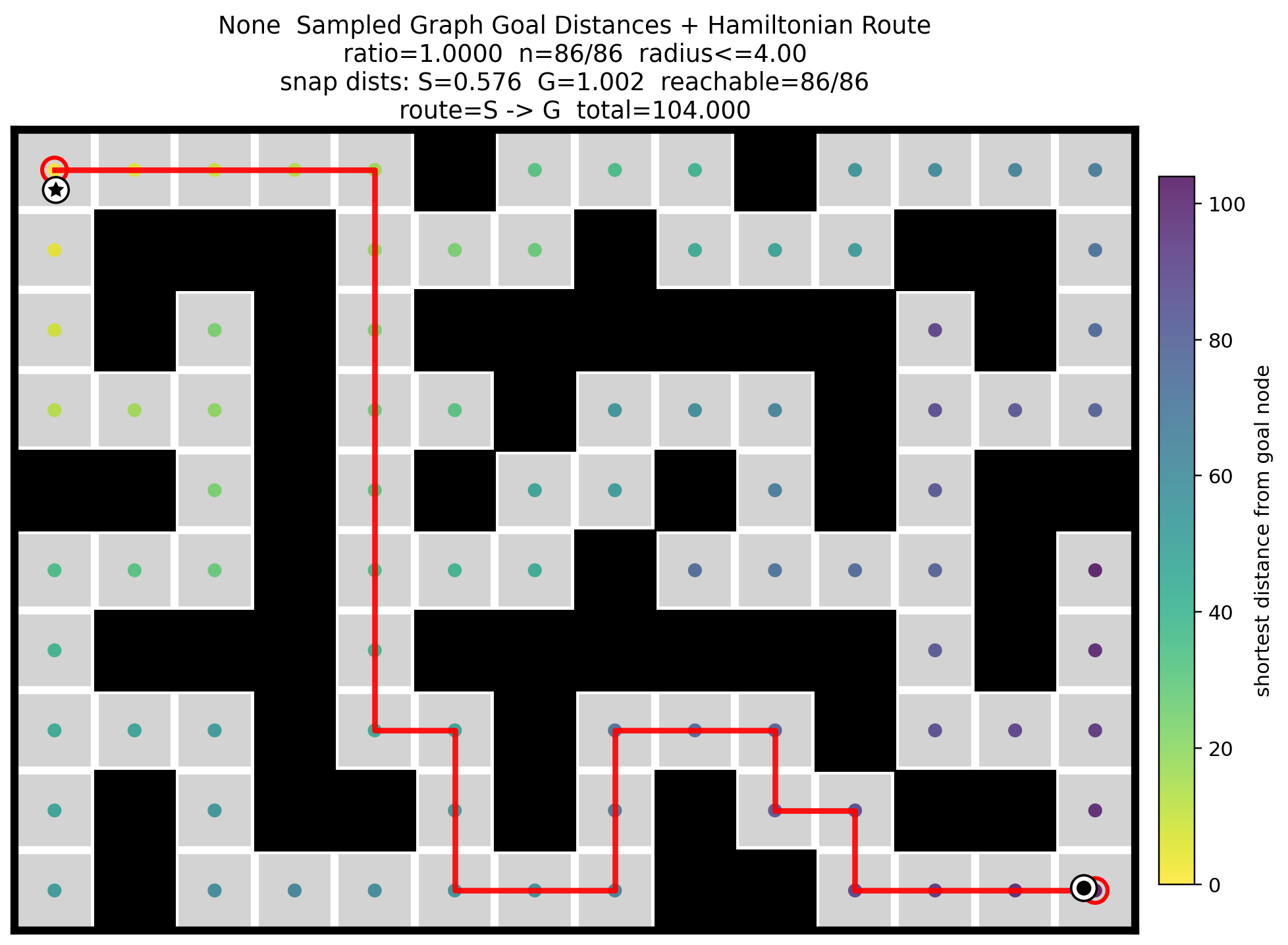} &
    \includegraphics[width=0.19\textwidth]{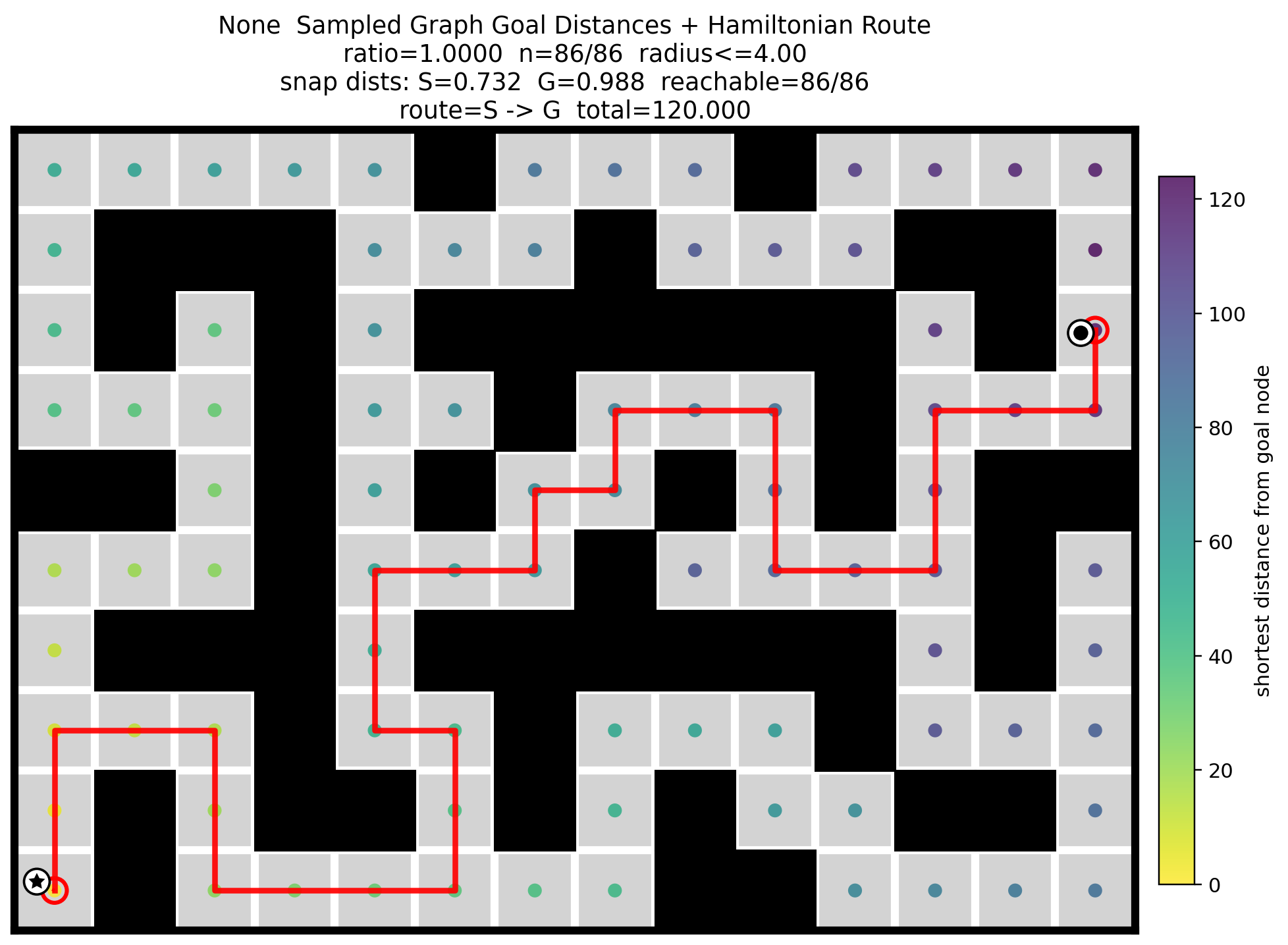} &
    \includegraphics[width=0.19\textwidth]{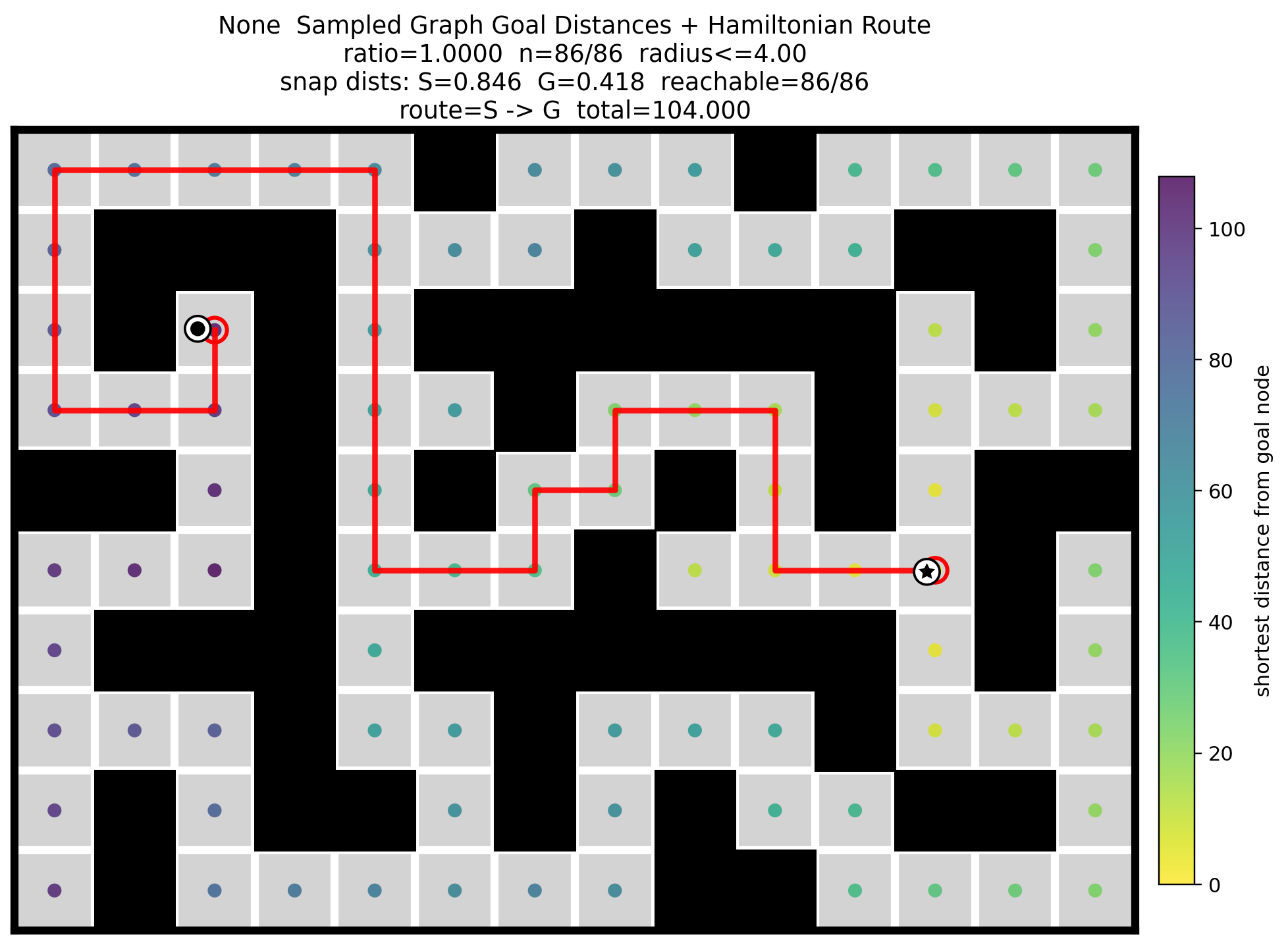} &
    \includegraphics[width=0.19\textwidth]{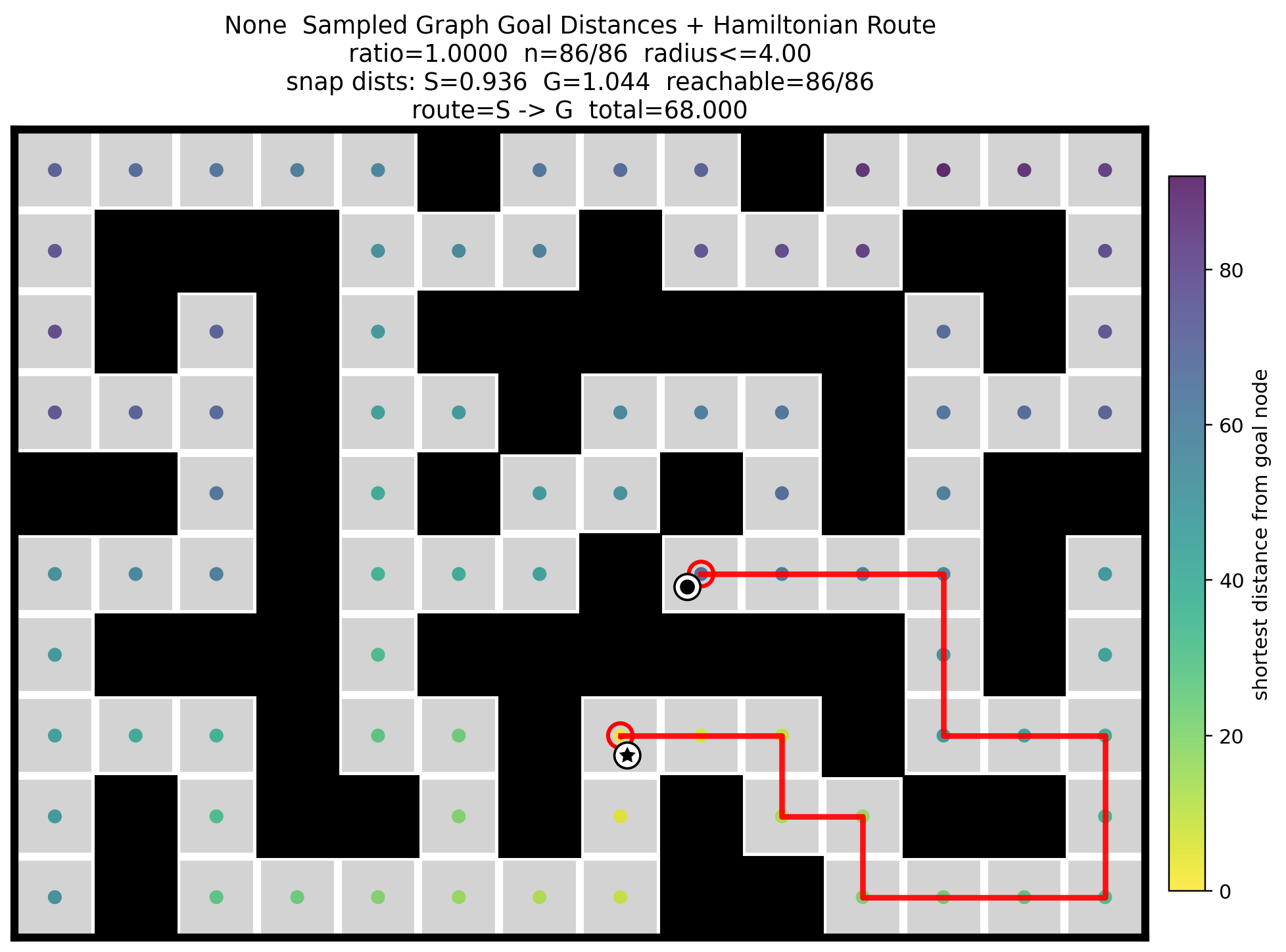} \\
    \includegraphics[width=0.19\textwidth]{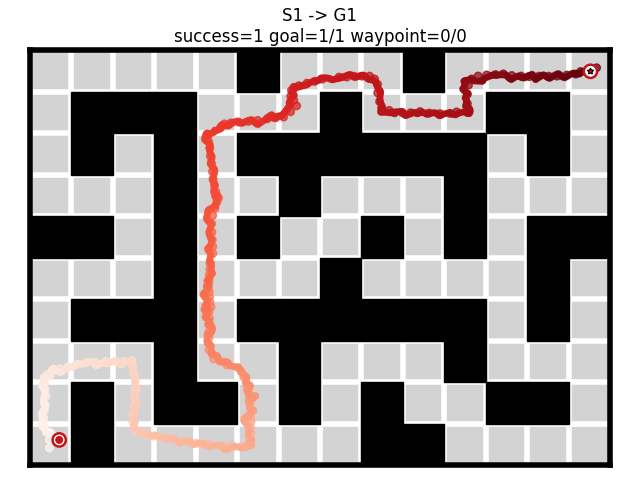} &
    \includegraphics[width=0.19\textwidth]{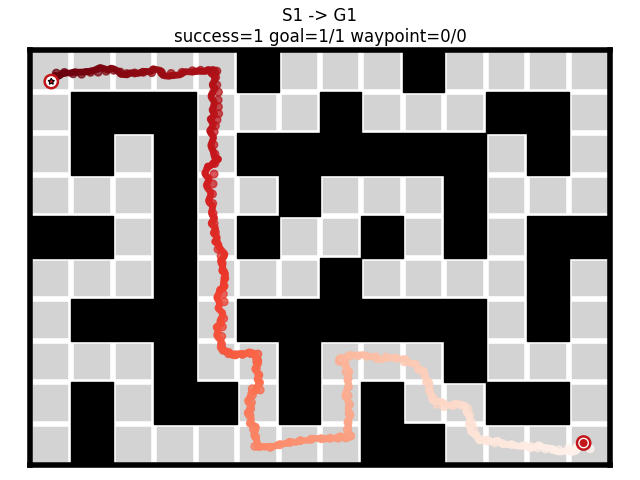} &
    \includegraphics[width=0.19\textwidth]{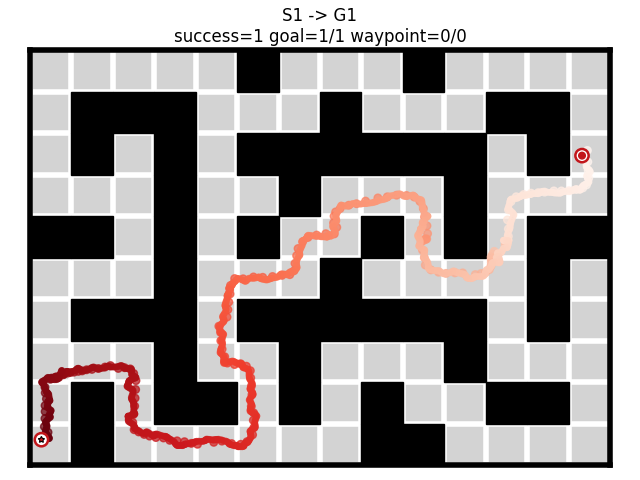} &
    \includegraphics[width=0.19\textwidth]{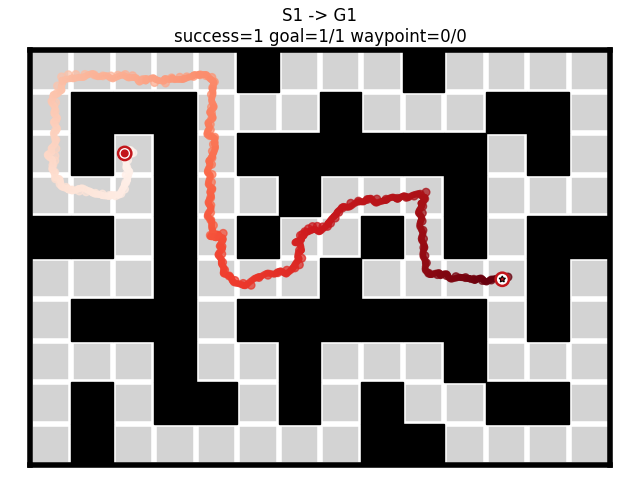} &
    \includegraphics[width=0.19\textwidth]{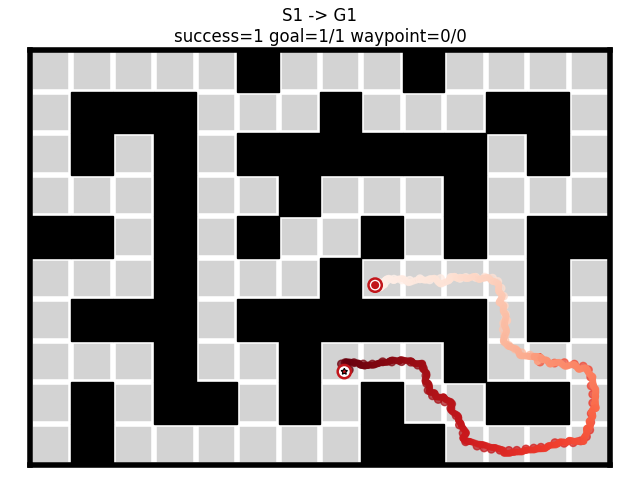}
  \end{tabular}
  \caption{Qualitative route-length validation on the official OGBench
  \texttt{antmaze-giant-stitch-v0} tasks, corresponding to the Giant row of
  Table~\ref{tab:plan_length_bfs}. Columns show Tasks~1--5. The top row shows
  the BFS grid shortest-path reference, and the bottom row shows the
  corresponding ChronoForest rollout. The rollouts follow the same maze
  corridors as the BFS references, providing a qualitative companion to the
  BFS-relative route-length measurements in Table~\ref{tab:plan_length_bfs}.}
  \label{fig:antmaze_giant_stitch_routes}
\end{figure}

Table~\ref{tab:plan_length_bfs} provides the official-stitch support for the
route-efficiency claim in the main text. Across the 15 task-split combinations,
14 achieve 100.0\% of rollouts within $1.1\times$BFS, and the only exception is
Large Task~2 at 97.5\%, where two rollouts out of 80 exceed the threshold.

\subsection{Full-Planner Efficiency Diagnostics}
\label{app:planning-cost-support}

\begin{table*}[t]
  \centering
  \footnotesize
  \setlength{\tabcolsep}{4pt}
  \begin{tabular}{l cc cc cc}
    \toprule
    & \multicolumn{2}{c}{ChronoForest} & \multicolumn{2}{c}{Start-tree-only} & \multicolumn{2}{c}{L2-root selection} \\
    \cmidrule(lr){2-3}\cmidrule(lr){4-5}\cmidrule(lr){6-7}
    Task & Plan. Lat. (s) $\downarrow$ & Nodes $\downarrow$ & Plan. Lat. (s) $\downarrow$ & Nodes $\downarrow$ & Plan. Lat. (s) $\downarrow$ & Nodes $\downarrow$ \\
    \midrule
    Task 1 & \textbf{63.0} & \textbf{9.0} & 73.8 & 27.0 & 68.2 & 17.0 \\
    Task 2 & 56.0 & \textbf{7.8} & 91.9 & 33.1 & \textbf{43.0} & 11.9 \\
    Task 3 & \textbf{66.5} & \textbf{8.9} & 92.9 & 35.2 & 72.4 & 19.1 \\
    Task 4 & \textbf{66.1} & \textbf{10.7} & 71.5 & 24.8 & 179.6 & 40.0 \\
    Task 5 & 76.5 & \textbf{10.9} & \textbf{67.8} & 21.0 & 535.5 & 31.9 \\
    Total & \textbf{65.6} & \textbf{9.5} & 79.6 & 28.2 & 179.8 & 24.0 \\
    \bottomrule
  \end{tabular}
  \caption{Task-wise planning-cost diagnostics on
  \texttt{antmaze-giant-stitch-v0} for the ChronoForest evaluation, a
  \emph{Start-tree-only} ablation that restricts expansion to the start tree,
  and an \emph{L2-root selection} ablation that replaces temporal-distance
  root ranking with geometric L2 ranking. Entries report mean planning-only
  latency in seconds and mean expanded nodes per rollout over one evaluation
  repeat with 10 rollouts per task.}
  \label{tab:planning_cost_ablation}
\end{table*}

Table~\ref{tab:planning_cost_ablation} provides the system-level support for the
search-efficiency claim in the main text. ChronoForest expands the fewest nodes
on every task and overall, reducing the total node count to 9.5 from 28.2 for
\emph{Start-tree-only}, which restricts expansion to the start tree, and 24.0
for \emph{L2-root selection}, which replaces temporal-distance root ranking
with geometric L2 ranking. This ties the first gap to the value of
bidirectional search and the second to the stability of
temporal-distance-guided prioritization.

\subsection{Broader-Regime Teleport Uncertainty Ablation}
\label{app:teleport-uncertainty}

\begin{table*}[t]
  \centering
  \scriptsize
  \setlength{\tabcolsep}{4pt}
  \resizebox{\textwidth}{!}{%
  \begin{tabular}{lcccccccccc}
    \toprule
    Method & GCBC & CRL & GC-SSCP & HIQL & SVL & GCIVL & Eik-HIQL & Eik-HiQRL & ChronoForest (w/o uncertainty) & ChronoForest \\
    \midrule
    Stitch & 31$\pm$6 & 31$\pm$4 & 33$\pm$4 & 36$\pm$2 & 38$\pm$1 & 39$\pm$3 & 47$\pm$2 & 60$\pm$3 & 49.2$\pm$2.9 & \textbf{68.8$\pm$3.6} \\
    \bottomrule
  \end{tabular}%
  }
  \caption{Quantitative stitch-only results on the OGBench AntMaze-Teleport
  benchmark. The two ChronoForest entries share the same directional
  quasimetric value model on \texttt{antmaze-teleport-stitch-v0} and differ
  only in whether uncertainty-aware ranking is disabled or enabled.}
  \label{tab:antmaze_teleport_benchmark}
\end{table*}

Table~\ref{tab:antmaze_teleport_benchmark} broadens the scope of the main-body
claims to a stochastic and quasimetric regime. Setting the uncertainty-aware
ranking coefficient to 0 instead of 10 lowers ChronoForest from $68.8\pm3.6$
to $49.2\pm2.9$, supporting the claim that uncertainty-aware node valuation
matters when route realization risk depends directly on stochastic
transitions. Our two rows report mean $\pm$ std over four repeated
evaluations, whereas the prior baselines are copied from reported OGBench
results and serve only as external scale.

\section{Pairwise Two-Agent Route-Order Robustness}
\label{app:route-order-robustness}
Table~\ref{tab:hamiltonian_online_fixed_pairs} expands the $M=2$ block of Table~\ref{tab:hamiltonian_online_fixed} into all ten task pairs. For each pair, the online, temporal-fixed, and graph-fixed variants are matched on the same 18 waypoint-group instances (three repeats times six groups), so the comparison isolates only the route-order update policy. Online re-solving improves over temporal-fixed on all ten pairs, with path savings ranging from 1.3\% to 7.9\% and averaging 5.7\%, confirming that the aggregate gain in the main text is not driven by a small subset of favorable pairs.

\begin{table*}[t]
  \centering
  \small
  \setlength{\tabcolsep}{5pt}
  \resizebox{\textwidth}{!}{%
  \begin{tabular}{lccccccccccc}
    \toprule
    Task Pair & 1--2 & 1--3 & 1--4 & 1--5 & 2--3 & 2--4 & 2--5 & 3--4 & 3--5 & 4--5 & Total \\
    \midrule
    Online Total Length $\downarrow$ & 226.3 & 234.9 & 237.5 & 204.3 & 230.9 & 224.5 & 189.8 & 228.9 & 201.2 & 184.6 & 216.3 \\
    Temporal Fixed Total Length $\downarrow$ & 229.2 & 253.2 & 256.5 & 216.2 & 246.7 & 236.7 & 197.0 & 239.1 & 218.3 & 200.4 & 229.3 \\
    Graph Fixed Total Length $\downarrow$ & 218.4 & 233.4 & 232.6 & 198.4 & 221.4 & 219.4 & 181.8 & 225.1 & 193.6 & 177.0 & 210.1 \\
    Path Saved vs. Temporal Fixed $\uparrow$ & 1.3\% & 7.2\% & 7.4\% & 5.5\% & 6.4\% & 5.2\% & 3.7\% & 4.3\% & 7.8\% & 7.9\% & 5.7\% \\
    \bottomrule
  \end{tabular}
  }
  \caption{Pairwise two-agent Hamiltonian sweep on
  \texttt{antmaze-giant-stitch-v0} over all ten $\binom{5}{2}$ task pairs from
  the five-task benchmark. Entries report the mean total postprocessed route
  length $\sum_{a=1}^{2} L_a$ for each task pair, and \textit{Path Saved vs.
  Temporal Fixed} reports the percent reduction relative to the temporal-fixed
  planner.}
  \label{tab:hamiltonian_online_fixed_pairs}
\end{table*}

\begin{figure*}[t]
  \centering
  \small
  \setlength{\tabcolsep}{3pt}
  \begin{tabular}{@{}c@{\hspace{3pt}}c@{\hspace{3pt}}c@{\hspace{3pt}}c@{}}
    & \shortstack{\textbf{Low}\\1--2 (1.3\%)} & \shortstack{\textbf{Near-Median}\\2--4 (5.2\%)} & \shortstack{\textbf{High}\\4--5 (7.9\%)} \\
    \textbf{Temporal} &
    \includegraphics[width=0.26\textwidth]{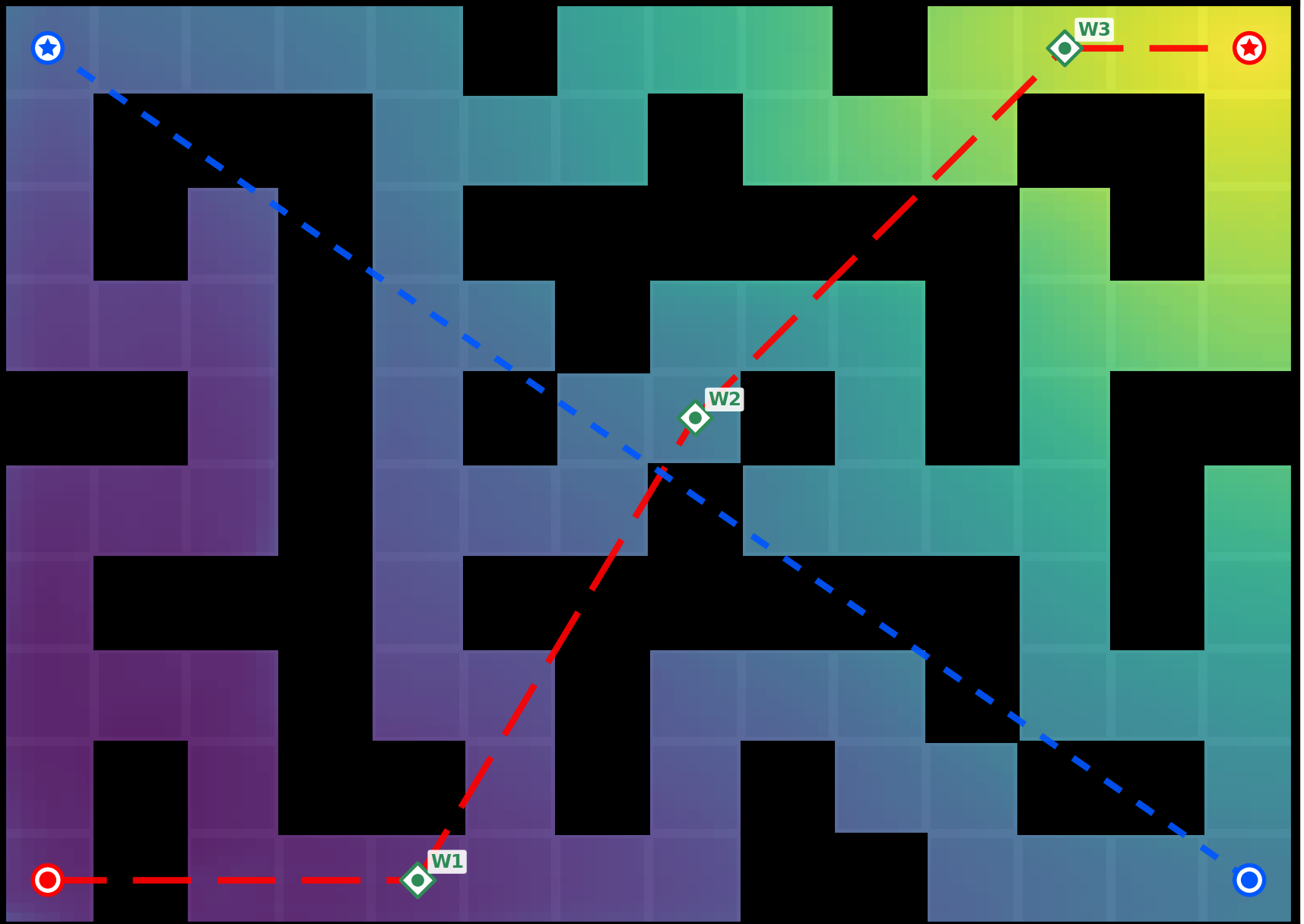} &
    \includegraphics[width=0.26\textwidth]{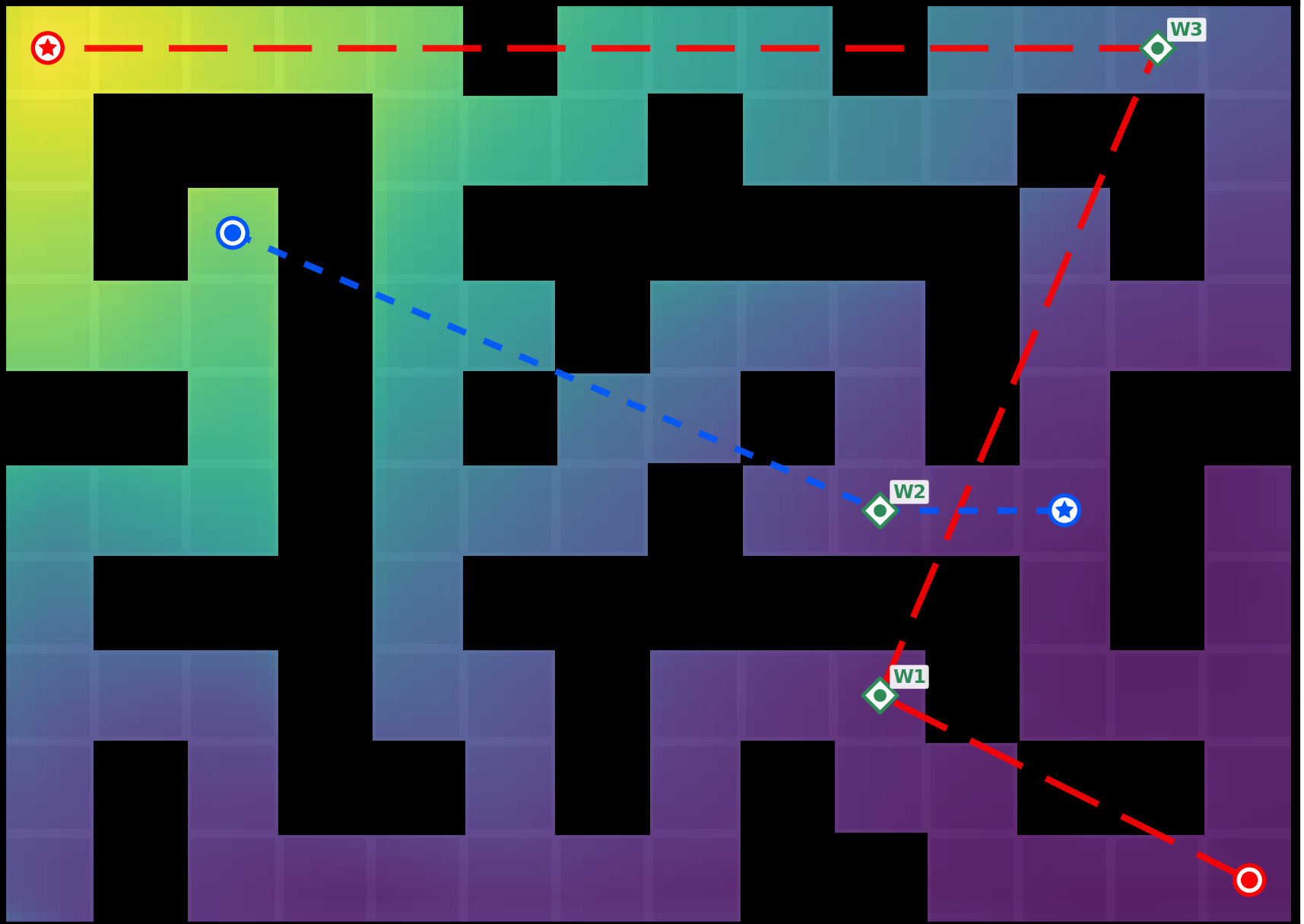} &
    \includegraphics[width=0.26\textwidth]{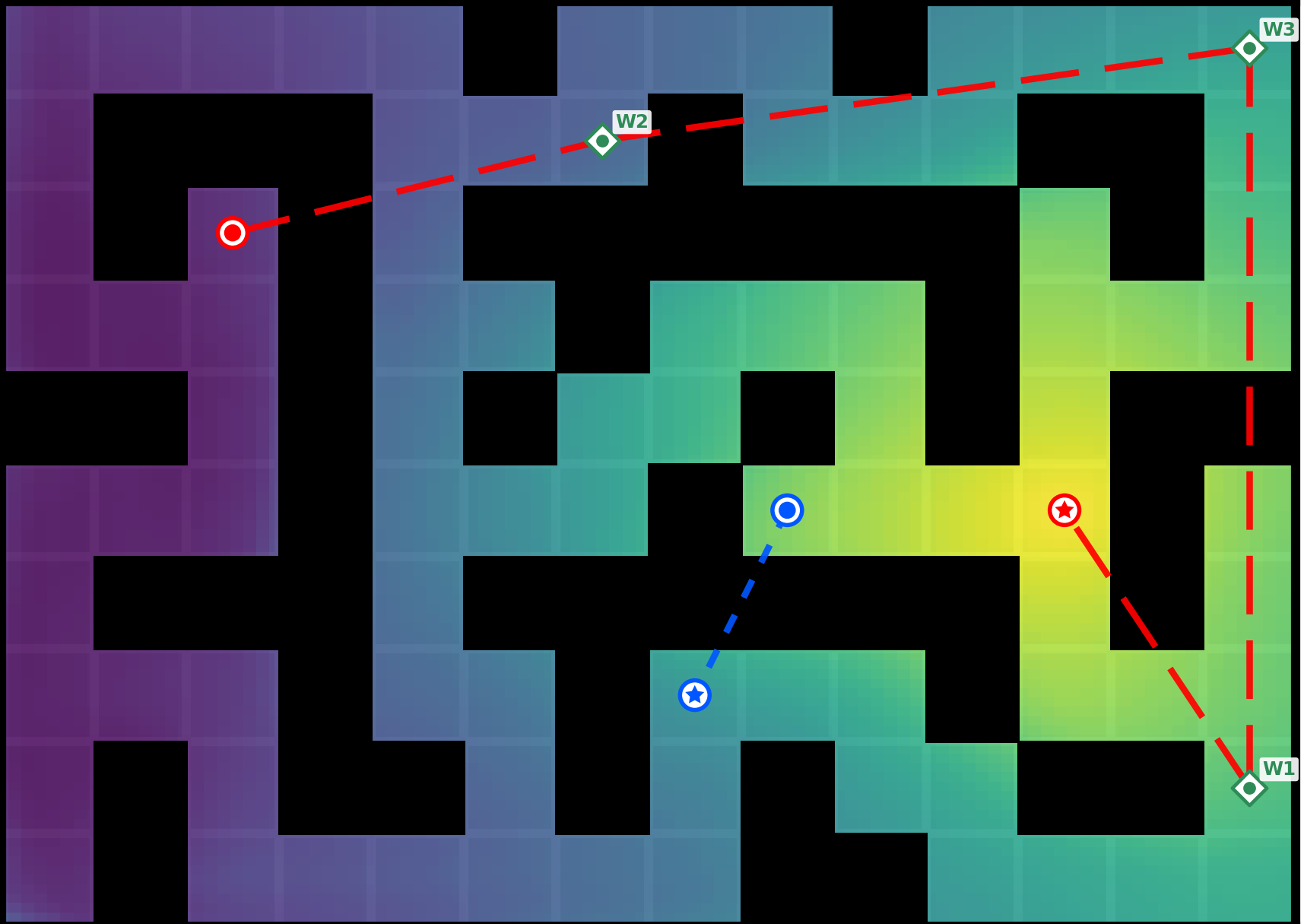} \\
    \textbf{Graph-fixed} &
    \includegraphics[width=0.26\textwidth]{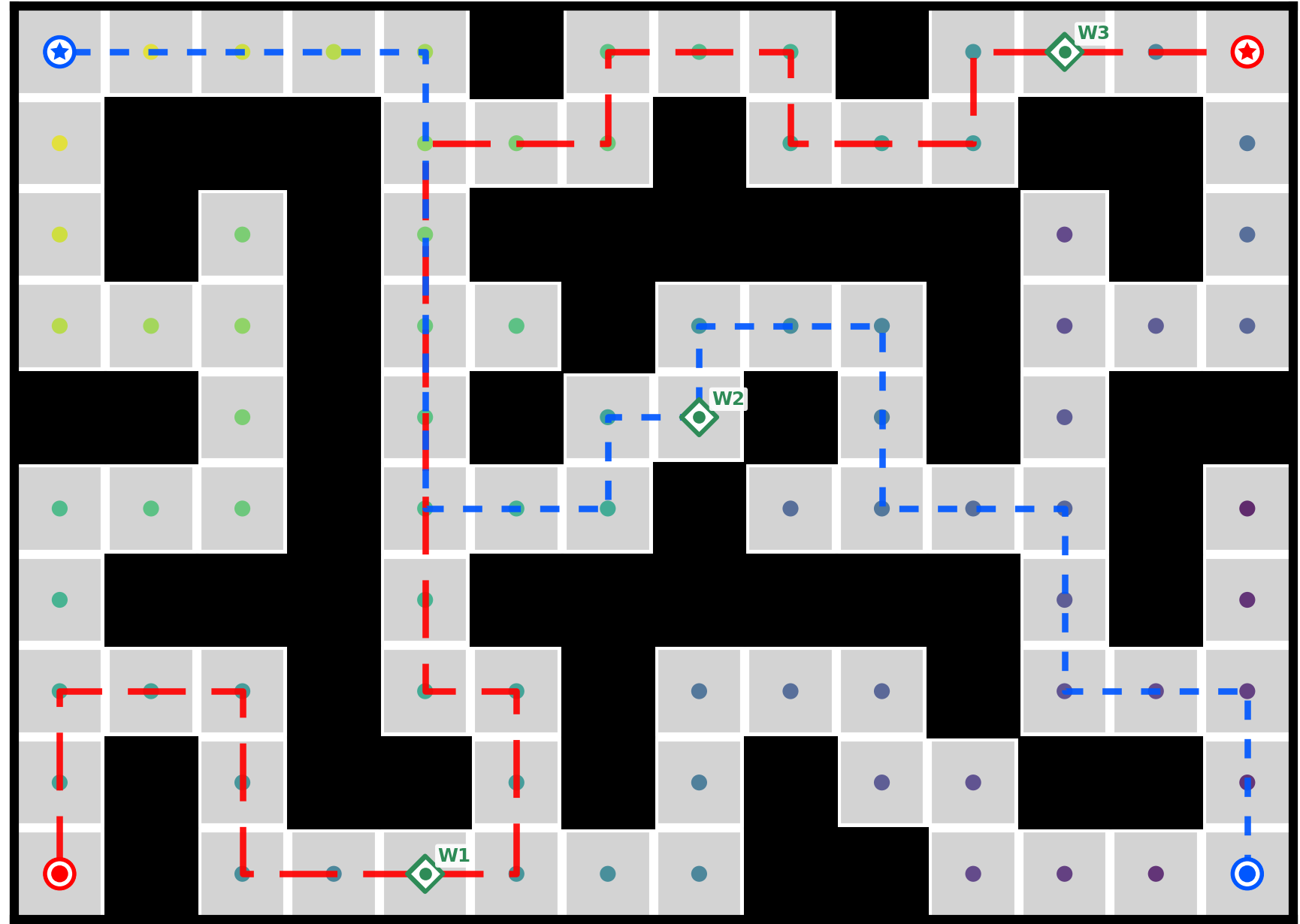} &
    \includegraphics[width=0.26\textwidth]{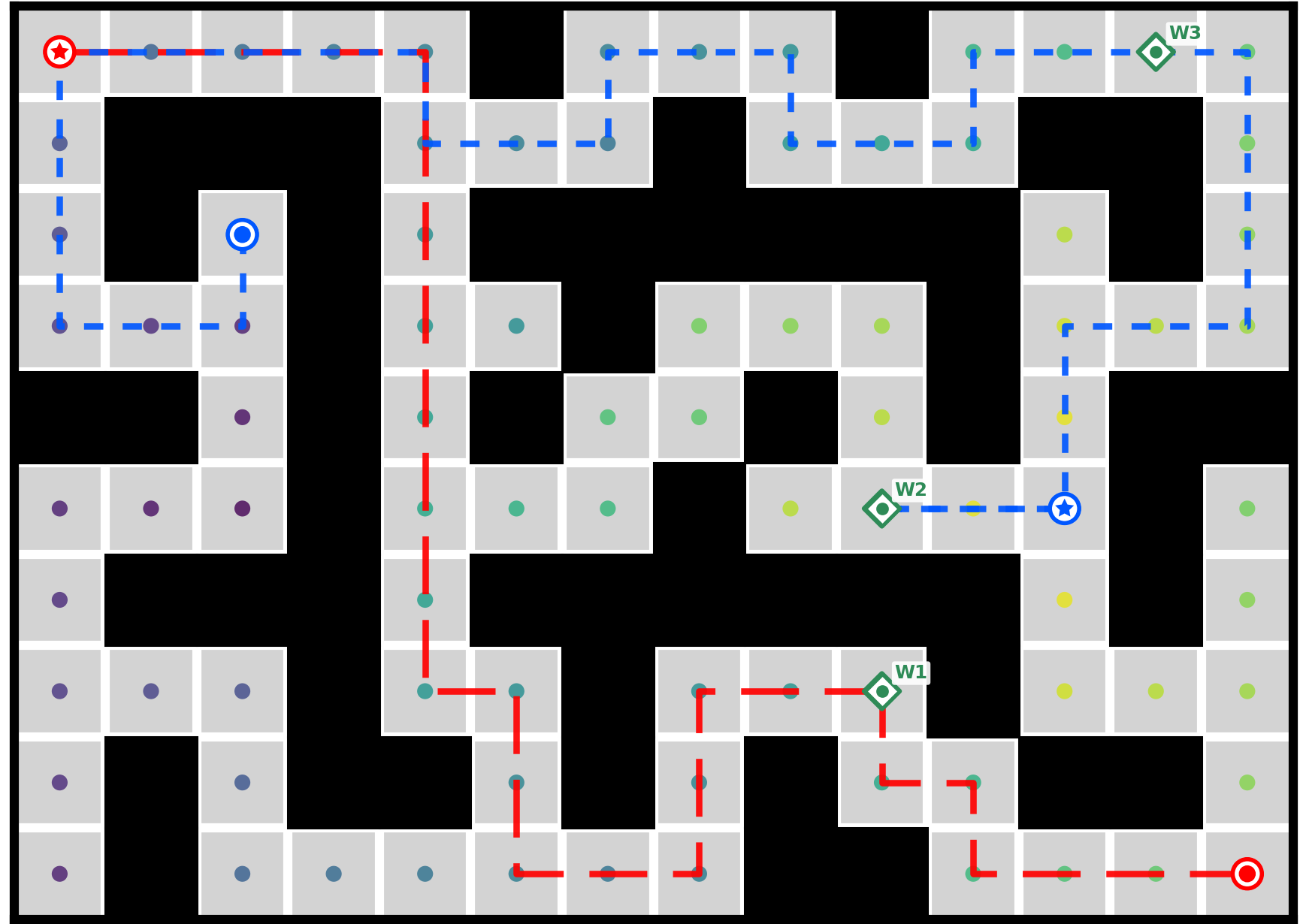} &
    \includegraphics[width=0.26\textwidth]{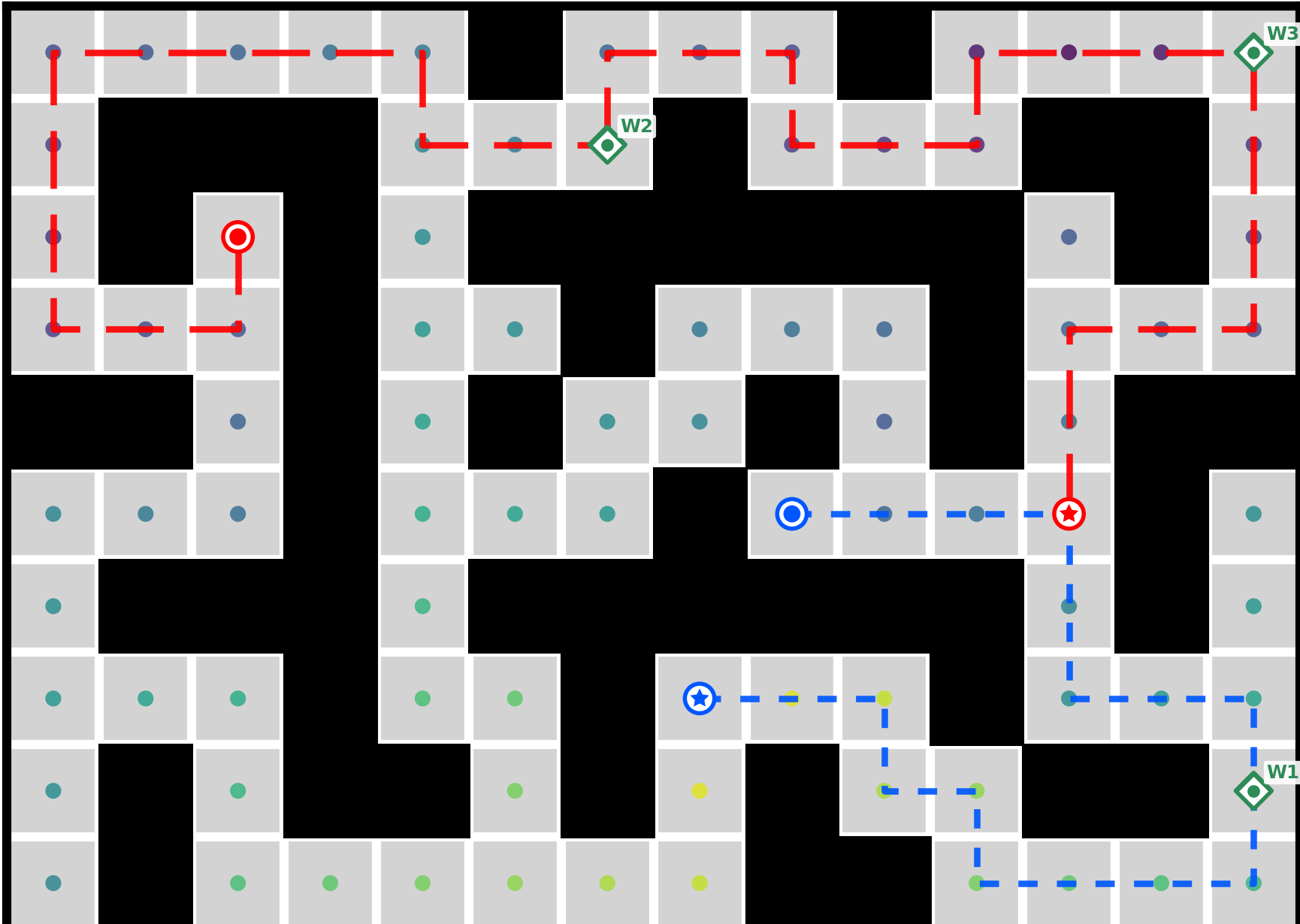} \\
    \shortstack{\textbf{Online}\\\textbf{postprocessed}} &
    \includegraphics[width=0.26\textwidth]{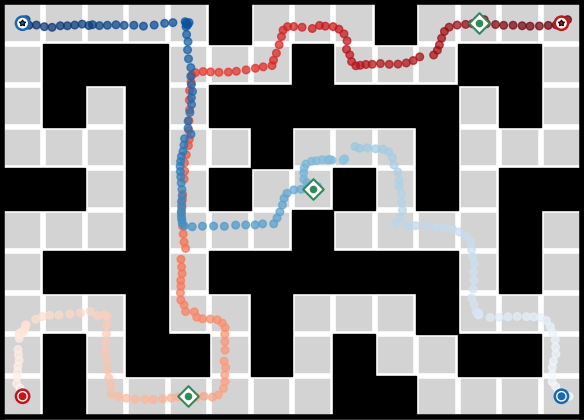} &
    \includegraphics[width=0.26\textwidth]{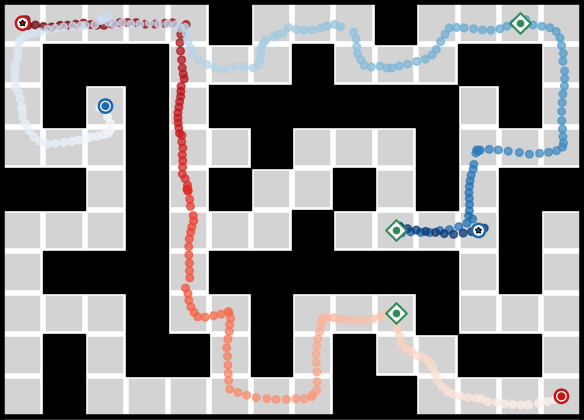} &
    \includegraphics[width=0.26\textwidth]{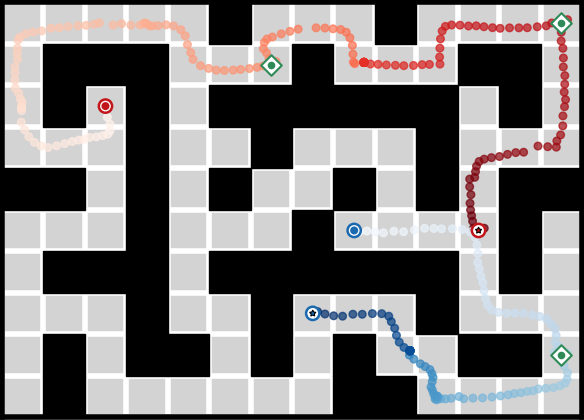} \\
  \end{tabular}
  \caption{Representative matched waypoint-group instances for the $M=2$ pairwise sweep in Table~\ref{tab:hamiltonian_online_fixed_pairs}. Columns use the first matched group from the lowest-savings pair 1--2, the near-median-savings pair 2--4, and the highest-savings pair 4--5. Rows show the temporal-distance route, the privileged graph-fixed route, and the online postprocessed route. Across all three examples, online re-solving moves the postprocessed route toward the graph-fixed ordering rather than preserving the temporal-distance order.}
  \label{fig:hamiltonian_online_fixed_pair_examples}
\end{figure*}

Figure~\ref{fig:hamiltonian_online_fixed_pair_examples} shows the first matched waypoint-group instance from three task pairs in Table~\ref{tab:hamiltonian_online_fixed_pairs}: the lowest-savings pair 1--2 (1.3\%), the near-median-savings pair 2--4 (5.2\%), and the highest-savings pair 4--5 (7.9\%). Across all three examples, the temporal-distance order disagrees with the graph-fixed route, while the online postprocessed route shifts toward the graph-fixed reference instead of preserving the temporal order.

\section{Implementation Details}
\label{app:implementation}

\subsection{Software and Hardware}
\label{app:impl:infra}

All experiments are conducted inside a Docker container based on image \texttt{mctd:0.1}.
Key dependencies include PyTorch~2.5.1, JAX~0.6.2, Flax (compatible with JAX~0.6.2), MuJoCo~2.1.0 (\texttt{mujoco-py}), OGBench~1.0.1, PyTorch Lightning~2.1.2, and Hydra~1.3.2, running on Python~3.10 with CUDA~12.4.
All training and evaluation runs on a single NVIDIA RTX 3090 GPU with 24\,GB VRAM and 32\,GB system RAM.

\subsection{Datasets}
\label{app:impl:data}

We evaluate on two OGBench~\citep{park2024ogbench} AntMaze environments:
\texttt{antmaze-giant-stitch-v0} and \texttt{antmaze-teleport-stitch-v0},
each containing five navigation tasks (\texttt{num\_tasks}$=5$).
Raw dataset episodes consist of 201 frames.
We subsample every $k{=}5$ frames (frame stride \texttt{jump}$=5$),
yielding an effective episode length of 200 tokens.
Observations are projected to 2D $(x, y)$ coordinates (\texttt{obs\_dim\_indices}$=[0,1]$)
for the diffusion model; the DQL actor uses the full 29-dimensional proprioceptive state
augmented with a 2D goal.

\subsection{Diffusion Forcing Transformer}
\label{app:impl:dft}

The trajectory diffusion model is a non-causal transformer
trained with Diffusion Forcing~\citep{chen2024diffusionforcing}.

\paragraph{Architecture.}
The model uses a pyramid scheduling matrix with frame stack $F{=}10$,
\texttt{context\_frames}$=1$, and \texttt{padding\_mode}\,$=$\,\texttt{same}.
The transformer has $d_\text{model}{=}128$, 12 layers, 8 attention heads,
and feedforward dimension 512.

\paragraph{Diffusion.}
We use a linear noise schedule with $T{=}1000$ steps,
$x_0$-prediction objective, SNR clipping at 5.0,
cumulative SNR decay $\rho{=}0.98$, and stabilization level 10.

\paragraph{Training.}
We train with Adam ($\beta_1{=}0.9$, $\beta_2{=}0.999$, weight decay $10^{-4}$),
learning rate $2{\times}10^{-4}$ with cosine decay (ratio 0.1) and 10{,}000 warmup steps,
batch size 2{,}048 for 600{,}015 gradient steps.
Training takes approximately 24 hours on a single RTX 3090.

\subsection{DQL Actor}
\label{app:impl:dql}

We train a Diffusion Q-Learning (DQL)~\citep{wang2023diffusion} actor
to execute sub-goal-conditioned actions in the environment.

\paragraph{Architecture.}
The actor is a three-layer MLP
($256{\to}256{\to}256$, Mish activation) conditioned on a sinusoidal time embedding
(dimension 16), with input dimension 31 (29D proprioceptive state $+$ 2D goal).
The critic is a 64-ensemble vectorized network
($256{\to}256{\to}256{\to}1$ per member).
The actor uses 5 diffusion denoising steps with a VP noise schedule.

\paragraph{Training.}
We train for 200 epochs with Adam ($\text{lr}{=}3{\times}10^{-4}$),
batch size 256, gradient norm clipping at 7.0,
EMA decay 0.995, behavior cloning weight $p{=}0.2$,
LCB coefficient 4.0, discount $\gamma{=}0.99$, and target EMA rate $\tau{=}0.005$.
Training takes approximately 3 hours.

\subsection{Temporal Distance Representation}
\label{app:impl:hilp}

We train a quasimetric temporal distance function following
the dual goal representation framework of \citet{park2024hilp}.
Only Phase~1 (dual representation training) is used;
Phase~2 (downstream GCVF) is not employed in this work.

\paragraph{Architecture.}
Encoder $\psi$ and $\phi$ each map states to a 32-dimensional latent space
through three-layer MLPs ($512{\to}512{\to}512$) with LayerNorm.
We use the \texttt{neg\_l2} aggregator for \texttt{antmaze-giant} and the \texttt{quasimetric} (\texttt{neg\_l2} with \texttt{ReLU}) aggregator for \texttt{antmaze-teleport}.

\paragraph{Training.}
We train for 3{,}000{,}000 steps with Adam ($\text{lr}{=}3{\times}10^{-4}$),
batch size 1{,}024, discount $\gamma{=}0.99$, EMA rate $\tau{=}0.005$,
expectile 0.95, gradient clipping at 1.0, and goal sampling ratios
$p_\text{traj}{=}0.625$, $p_\text{rand}{=}0.375$, $p_\text{curr}{=}0.0$.
Training takes approximately 2 hours.

\subsection{Planning Hyperparameters}
\label{app:impl:planning}

Table~\ref{tab:planning_hparams} lists the hyperparameters used during
MCTD evaluation on all environments.

\begin{table}[h]
\centering
\caption{MCTD evaluation hyperparameters.}
\label{tab:planning_hparams}
\small
\begin{tabular}{llr}
\toprule
\textbf{Category} & \textbf{Parameter} & \textbf{Value} \\
\midrule
\multirow{2}{*}{Diffusion / DDIM}
  & Sampling timesteps          & 100   \\
  & DDIM $\eta$                 & 0.1   \\
\midrule
\multirow{3}{*}{Tree search}
  & Max expansions (\texttt{mctd\_max\_search\_num}) & 40    \\
  & Sequence dividing factor    & 10    \\
  & Segment episode length      & 200   \\
\midrule
\multirow{6}{*}{Uncertainty estimation}
  & Uncertainty mode            & \texttt{expected\_root\_node\_dist} \\
  & Uncertainty rank $\lambda$  & 0.1   \\
  & Fast samples per candidate  & 6     \\
  & Fast sampling steps         & 50    \\
  & TD overestimation coeff.\ $\alpha$ & 0.005 \\
  & Max intra-cluster TD dist.  & 30.0  \\
\midrule
\multirow{3}{*}{KDE guidance}
  & KDE bandwidth $\sigma$      & 0.1   \\
  & KDE grad.\ threshold coeff. & 0.3   \\
  & KDE dataset sample ratio    & 0.1   \\
\midrule
\multirow{3}{*}{HILP guidance}
  & Guidance scales             & $[0.8,\;0]$ \\
  & Direct $x_0$ guidance scale $\kappa$ & 0.2 \\
  & Far-target TD threshold     & $-200.0$ \\
\midrule
\multirow{2}{*}{Bidirectional convergence}
  & Meeting $\delta$            & 1.2   \\
  & TD execution scale          & 0.2   \\
\bottomrule
\end{tabular}
\end{table}

\end{document}